\documentclass[preprint,10pt,5p,twocolumn,authoryear]{elsarticle}

\usepackage{amssymb}
\usepackage{pdflscape}
\usepackage{makecell}
\usepackage[T1]{fontenc}
\usepackage{amsmath}
\usepackage{amssymb}
\usepackage{amsmath}
\usepackage{caption} 
\usepackage{booktabs} 
\usepackage{multirow} 
\usepackage{array} 
\usepackage{graphicx}
\usepackage{subcaption}
\usepackage{float}
\usepackage{hyperref}
\usepackage{multicol}
\usepackage{lipsum} 
\usepackage{xcolor} 
\usepackage{tabularx}   
\usepackage{booktabs}   
\usepackage{fontawesome5}
\usepackage{pifont}
\usepackage{placeins}
\usepackage{afterpage}
\usepackage{longtable}
\usepackage{booktabs}
\usepackage{stfloats} 
\let\oldcite\cite
\renewcommand{\cite}[1]{\textcolor{cyan!75!black}{\oldcite{#1}}}

\journal{Medical Image Analysis}

\begin{document}

\begin{frontmatter}



\title{Deep Learning in Dental Image Analysis: A Systematic Review of Datasets, Methodologies, and Emerging Challenges} 

\cortext[cor1]{Corresponding Author, Email: litao@nankai.edu.cn}
\author[label1,label2]{Zhenhuan Zhou} 
\author[label3]{Jingbo Zhu}
\author[label5]{Yuchen Zhang}
\author[label6]{Xiaohang Guan}
\author[label1,label2]{Peng Wang}
\author[label1,label2,label4]{Tao Li\corref{cor1}}

\affiliation[label1]{organization={College of Computer Science, Nankai University},
            city={Tianjin},
            postcode={300350}, 
            country={China}}
\affiliation[label2]{organization={Key Laboratory of Data and Intelligent System Security, Ministry of Education},
            country={China}}
\affiliation[label3]{organization={College of Software, Nankai University},
            city={Tianjin},
            country={China}}
\affiliation[label4]{organization={Department of stomatology, Tianjin Union Medical Center (The First Affiliated Hospital of Nankai University)},
            city={Tianjin},
            country={China}}
\affiliation[label5]{organization={Tianjin Stomatological Hospital},
            city={Tianjin}, 
            country={China}}
\affiliation[label6]{organization={Haihe Lab of ITAI},
            city={Tianjin},
            postcode={300459}, 
            country={China}}

\begin{abstract}
Efficient analysis and processing of dental images are crucial for dentists to achieve accurate diagnosis and optimal treatment planning. However, dental imaging inherently poses several challenges, such as low contrast, metallic artifacts, and variations in projection angles. Combined with the subjectivity arising from differences in clinicians’ expertise, manual interpretation often proves time-consuming and prone to inconsistency. Artificial intelligence (AI)–based automated dental image analysis (DIA) offers a promising solution to these issues and has become an integral part of computer-aided dental diagnosis and treatment. Among various AI technologies, deep learning (DL) stands out as the most widely applied and influential approach due to its superior feature extraction and representation capabilities. To comprehensively summarize recent progress in this field, we focus on the two fundamental aspects of DL research—datasets and models. In this paper, we systematically review 260 studies on DL applications in DIA, including 49 papers on publicly available dental datasets and 211 papers on DL-based algorithms. We first introduce the basic concepts of dental imaging and summarize the characteristics and acquisition methods of existing datasets. Then, we present the foundational techniques of DL and categorize relevant models and algorithms according to different DIA tasks, analyzing their network architectures, optimization strategies, training methods, and performance. Furthermore, we summarize commonly used training and evaluation metrics in the DIA domain. Finally, we discuss the current challenges of existing research and outline potential future directions. We hope that this work provides a valuable and systematic reference for researchers in this field. All supplementary materials and detailed comparison tables will be made publicly available on GitHub.
\end{abstract}

\end{frontmatter}

\section{Introduction}
With the continuous improvement of global living standards and the steady advancement of medical science, oral health has increasingly become an important indicator of human well-being and social development, as well as an essential component of overall health for all humankind (\cite{listl2025oral}). However, over the past three decades, oral diseases have remained among the most prevalent health conditions worldwide, significantly affecting the health and quality of life of billions of people (\cite{benzian2025global}). According to statistics from the World Health Organization, as of 2019, more than 3.5 billion people—nearly half of the global population—were affected by oral diseases of varying severity (\cite{world2022global, world2024global}). By 2021, this number had risen to 3.69 billion \cite{bernabe2025trends}, exceeding the combined total cases of the world’s five major non-communicable diseases (mental disorders, cardiovascular disease, diabetes mellitus, chronic respiratory diseases, and cancers) by over 1 billion (\cite{world2022global}).

A significant factor contributing to this issue is the acute shortage of oral healthcare professionals on a global scale. Developing a competent junior dentist typically demands over ten years of training, leading to a global average of merely about 3.3 dentists per 10,000 individuals (\cite{gallagher2024health}). This shortage is especially severe in low-income and less developed regions (\cite{world2025workforce}). With the rapid developments in machine learning and artificial intelligence (AI) technologies, AI-driven computer-aided diagnosis (CAD) systems are increasingly being adopted in the dental field (\cite{cui2022fully, wang2025tooth}). These systems alleviate the scarcity of dental professionals, reduce their workload, and improve diagnostic efficiency and accuracy. By minimizing the subjectivity associated with variations in clinical experience, they provide robust support for precise diagnosis and treatment planning. The implementation of semi-automatic or fully automatic intelligent Dental Image Analysis (DIA) represents a crucial step and prerequisite for advancing such CAD systems.

\begin{figure*}[h]
	\centerline{\includegraphics[width=\linewidth]{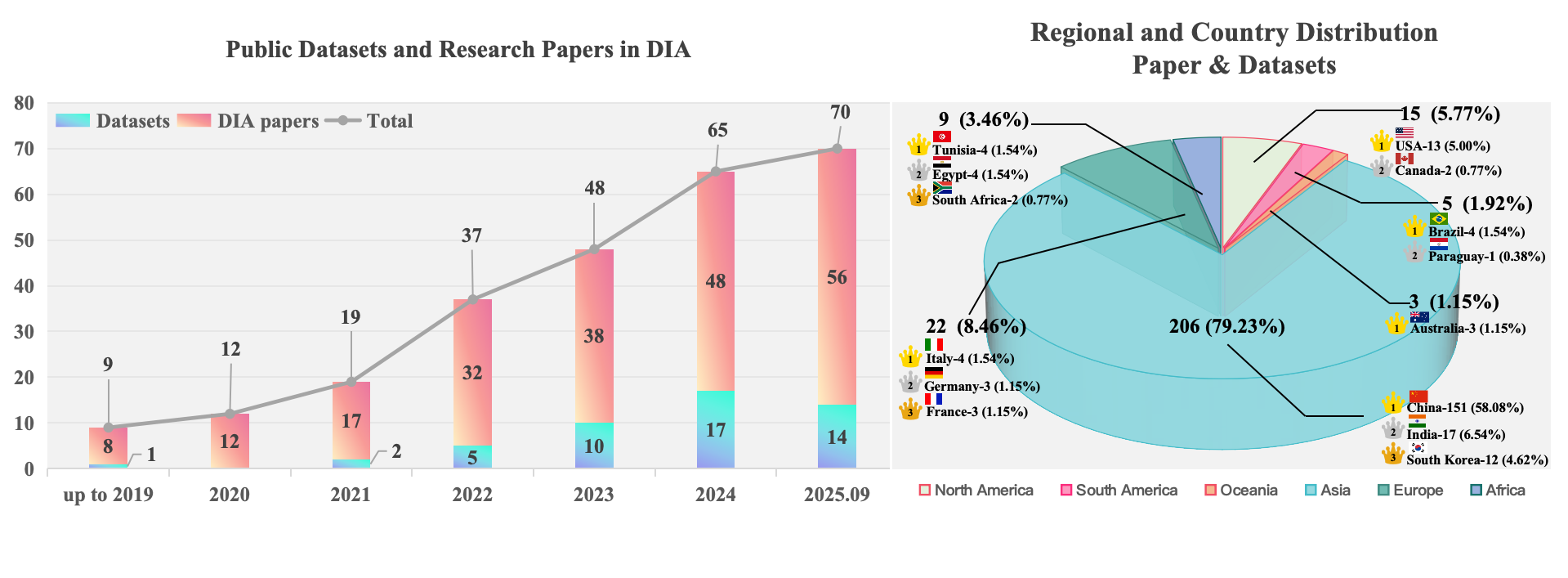}}
	\caption{Statistical results from the papers reviewed in this study are visualized in two charts. The bar chart on the left depicts the annual distribution of DL-based, DIA-related papers and public datasets, with the gray line indicating their cumulative total. The pie chart on the right illustrates the regional distribution, highlighting the top three countries with the highest number of studies in each continent.}
	\label{fig1}
\end{figure*}

As one of the most widely applied AI technologies today, deep learning (DL) (\cite{lecun2015deep}) has demonstrated remarkable potential in various domains of medical image analysis and processing, such as dermoscopy (\cite{mirikharaji2023survey, wu2025trans}), fundus imaging (\cite{he2022progressive, li2023generic}), pathology (\cite{song2024analysis, shen2025large}), ultrasound (\cite{ibtehaz2023acc, lin2023shifting}), chest X-rays (\cite{shao2022lidp, zhao2024topicwise}), and abdominal CT scans (\cite{rahman2025effidec3d, he2023h2former}). Unlike other fields of medical imaging, dental imaging possesses unique characteristics that present diverse challenges for DIA. For instance, unfavorable projection angles can lead to ambiguous tooth topology or overlapping teeth in 2D images (\cite{chaitanya2024enhancing}); noise interference and the inherently low contrast between soft tissues (\cite{venkateshConeBeamComputed2017}); metallic implants can introduce artifacts that hinder recognition and interpretation (\cite{liu2023toothsegnet}); and the presence of deciduous teeth in children may cause feature confusion (\cite{zhou2024nkut}). Naturally, as the importance of oral health continues to grow, an increasing number of AI researchers have turned their attention to DIA, leading to the emergence of numerous DL-based methods. The left panel of Figure \ref{fig1} illustrates the annual growth trend of DIA-related publications employing DL from 2016 to September 2025. It can be observed that the number of publications began to increase rapidly from 2022 and reached its most vigorous momentum in 2024. The right panel of Figure \ref{fig1} shows the geographical distribution of all included studies, with Asia—led by China—accounting for the largest share, followed by the Americas and Europe, while Africa has relatively few studies. This highlights the imbalance in the regional distribution of existing DIA research, which we will elaborate on in Section \ref{sec5}.

In summary, the application of DL in DIA has achieved substantial progress; however, a systematic review and synthesis are essential. From the perspectives of both dental practitioners and computer science researchers, a comprehensive examination of existing public dental imaging datasets and AI-based DL methodologies is crucial for advancing the field of intelligent DIA. Such a review not only aids in understanding how current approaches address inherent challenges and their limitations but also helps identify unresolved issues and anticipate emerging research trends in DIA.

\textbf{Relevant surveys.} Several existing surveys cover areas that are partially similar to this paper. For instance, Nambiar et al. (\cite{nambiar2024comprehensive}) conducted a concise review of 30 studies, categorizing them by dental specialty and illustrating the impact of DL-based models and their variants on dental practice. Alfadley et al. (\cite{alfadley2024progress}) summarized 13 studies to explore the application of UNet-based DL models in pulp segmentation. Kot et al. (\cite{kot2025evolution}) systematically reviewed 30 studies, categorizing them according to backbone network types, aiming to assess the evolution and performance of DL in tooth segmentation. Wu et al. (\cite{wu2025segmentation}) conducted a systematic review of 145 publications on dental CT image segmentation algorithms, classifying them into traditional algorithms and DL-based approaches (approximately 50 studies), and outlined the transition from classical methods to DL techniques. Compared to the aforementioned studies, our work differs in several key aspects: \textbf{(1)} Previous studies have largely overlooked a systematic investigation and meta-analysis of publicly available dental datasets. In contrast, this work offers a thorough overview of commonly used dental imaging modalities and accessible public datasets, conducting clustering analysis from multiple perspectives, including data modality, dataset scale, and acquisition methods. \textbf{(2)} Most existing research primarily focuses on panoramic (PAN) and cone-beam computed tomography (CBCT) as the main imaging modalities, emphasizing tooth segmentation tasks. Consequently, they fail to comprehensively cover the diverse tasks and modalities involved in DIA. Our review broadens this scope and provides a more holistic analysis. From a task categorization perspective, we refine dense tooth prediction into tooth-level and anatomical-level prediction, extending the review beyond tooth analysis to include lesion and maxillofacial structure prediction, along with dental classification tasks. Moreover, we incorporate previously underrepresented imaging modalities such as Intraoral Scanning (IOS), Intraoral Photography (IP), and periapical radiographs (PR). \textbf{(3)} Some prior reviews were conducted mainly from a medical perspective, with limited coverage and little attention to regional disparities. They also lacked in-depth analyses of model architectures and performance from a computer science viewpoint. In comparison, our survey systematically reviews 211 papers on DL-based DIA algorithms and 49 publicly accessible datasets, significantly surpassing previous efforts in both scope and scale. We perform clustering across all literature and datasets, providing multi-perspective analyses from a computer science standpoint. We hope that this work can offer valuable insights and guidance for researchers, fostering continuous progress and innovation within the DIA community.

\textbf{Search strategy.}
We conducted a systematic search across several widely used databases in computer science and medicine, restricting the time span to 2019–September 2025. The platforms included \href{https://scholar.google.com/}{Google Scholar}, \href{https://dblp.uni-trier.de/}{DBLP}, \href{https://ieeexplore.ieee.org/Xplore/home.jsp}{IEEE Xplore}, \href{https://link.springer.com/}{SpringerLink}, \href{https://www.elsevier.com/}{Elsevier}, and \href{https://pubmed.ncbi.nlm.nih.gov/}{PubMed}. For DBLP, the query was formulated as follows: \texttt{(trans* | deep | neural | learn*) (oral | dent* | tooth* | teeth | mandib* | maxill* | CBCT | "cone beam" | "intraoral scan*" | "panoramic X-ray") (segment* | classif* | detect*)}. Similar keyword combinations were applied to the other platforms to ensure comprehensive coverage of studies relevant to this review. For dataset collection, in addition to the above academic databases, we also searched open repositories oriented towards computer science, including \href{https://huggingface.co/papers/trending}{HuggingFace}, \href{https://www.kaggle.com/}{Kaggle}, and \href{https://www.mendeley.com/}{Mendeley}. The initial search yielded approximately 450 papers and 80 datasets. Two computer science researchers then independently performed the first round of screening to remove irrelevant items. The exclusion criteria for research papers were as follows:
\begin{itemize}
\item Studies not utilizing DL-based methods for DIA;
\item Works focusing on oral cancer, oral epithelium, or cell-level microscopic imaging that fall outside the scope of dental imaging;
\item Non-medical studies irrelevant to dental images (e.g., mechanical gear studies);
\item Preprints from arXiv and similar platforms were meticulously examined, and low-quality non-peer-reviewed papers were excluded;
\item Low-quality studies were excluded during the selection process. Specifically, datasets without public access and 2D datasets containing fewer than 50 cases or 3D datasets with fewer than 10 cases were not included in this review. For research papers, exclusion criteria encompassed studies with ambiguous methodologies, poorly designed experiments, or low reproducibility.
\end{itemize}

After applying these criteria, we identified a total of 211 research papers and 49 publicly available dental datasets. We then categorized them by publication year and geographic distribution (Fig. \ref{fig1}). The results indicate a significant increase in DIA-related DL research beginning in 2022, with both the number of publications and dataset releases steadily rising and reaching a peak in September 2025.

\begin{figure*}[h]
	\centerline{\includegraphics[width=\linewidth]{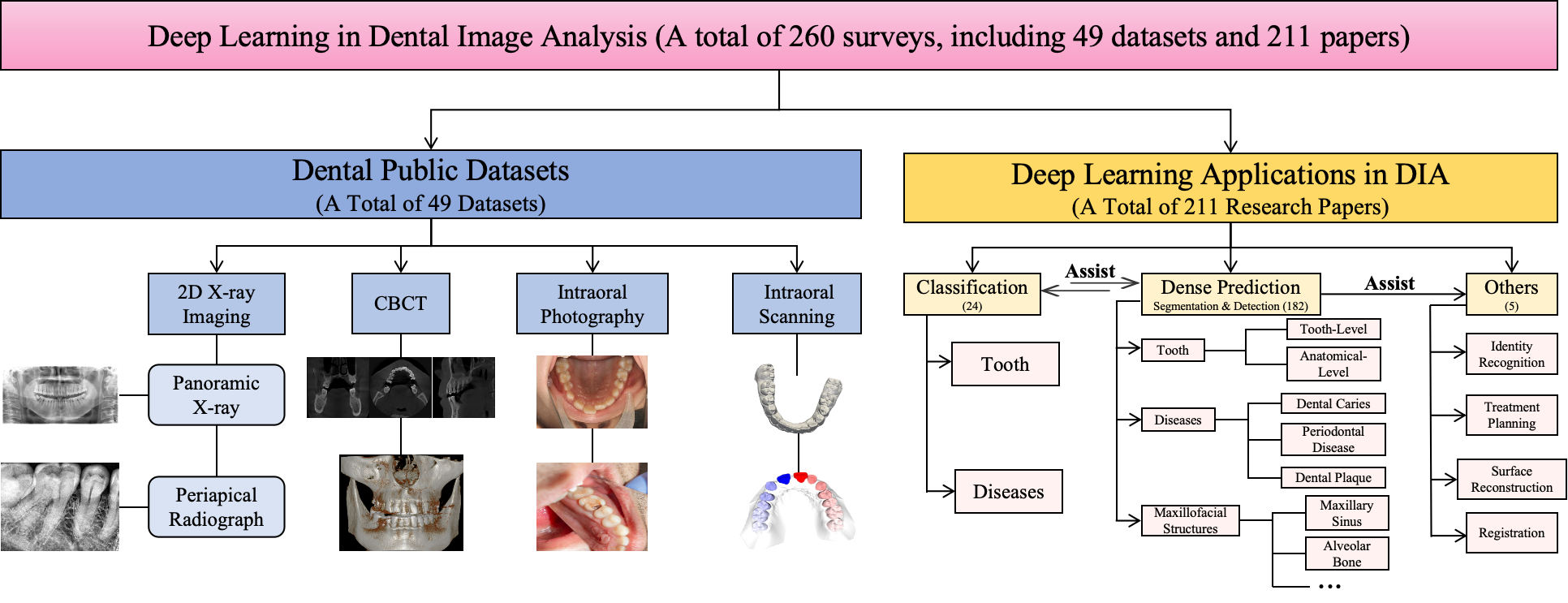}}
	\caption{Taxonomy of Deep Learning Applications in Dental Image Analysis. The diagram is divided into two main sections: Dental Public Datasets and Deep Learning Applications in Dental Image Analysis.}
	\label{fig2}
\end{figure*}

\textbf{Taxonomy.} For clarity of discussion, we categorized the publicly available dental datasets and DL-based DIA algorithms covered in this paper, as illustrated in Fig. \ref{fig2}. On the left side, from the perspective of datasets, all 49 datasets are clustered according to four commonly used dental imaging modalities, which include 2D X-ray imaging, CBCT, IP captured by standard cameras, and IOS. On the right side, from the perspective of models and applications, the 211 reviewed papers are classified into different DIA tasks, covering dense prediction, classification, and other related tasks. Examples of different data modalities and finer-grained literature categorizations are also provided in Fig. \ref{fig2}.

\textbf{Contribution.} In summary, the main contributions of this paper can be summarized as follows:
\begin{itemize}
\item We conducted a comprehensive and in-depth survey, reviewing 260 deep learning-based DIA studies—approximately an order of magnitude larger in scale compared to previous reviews. We also proposed an efficient taxonomy for all reviewed studies, designed to help readers from different backgrounds establish a clear and systematic understanding of this field.

\item We introduced the commonly used devices and imaging modalities in dental imaging, providing a systematic and comprehensive review of publicly available dental datasets, thereby addressing a gap left by previous surveys. This effort not only offers researchers a clear understanding of the current status of available DIA datasets but also provides valuable insights into the goals and directions for future dataset development.

\item We categorized all included studies based on their DIA task objectives and model architectures. Additionally, we provided a comprehensive statistical summary for each study, detailing the datasets used, network architectures, training parameters, hardware configurations, evaluation metrics, and performance outcomes, along with key innovations and highlights of each work.

\item Furthermore, we offer a detailed discussion and analysis of the current challenges and limitations of applying DL to DIA from both the dataset and model \& application perspectives. We also outline future trends, research prospects, and potential directions for exploration in this field.
\end{itemize}

\textbf{Paper Organization.} The remainder of this paper is organized as follows: Section \ref{sec2} introduces commonly used dental imaging modalities and the publicly available datasets reviewed in this study. Section \ref{sec3} discusses fundamental DL architectures and, in a task-oriented manner, examines DIA algorithms and applications, beginning with dense prediction. Section \ref{sec4} reviews commonly used loss functions, optimizers, GPUs, and performance evaluation metrics utilized in model training and testing. Section \ref{sec5} delves into the challenges and limitations of existing research, as well as potential directions for future development. Finally, Section \ref{sec6} concludes the paper.

\section{Dental Imaging and Public Datasets}
\label{sec2}
The establishment of high-quality public datasets is crucial for advancing research in DL within the field of DIA and constitutes one of the essential prerequisites for developing CAD systems in dentistry. In this section, we first introduce the commonly used dental imaging modalities, followed by a detailed overview of the publicly available datasets surveyed in this paper.

\subsection{Dental Imaging}
As an essential component of medical clinical imaging, dental imaging provides a crucial reference for disease diagnosis and treatment planning in dentistry, with well-established operational standards and clearly defined functional distinctions \cite{vandenbergheModernDentalImaging2010}. Dental imaging can be broadly categorized into four types: 2D X-ray imaging (such as PR and PAN), CBCT, IOS, and IP. In this section, we offer a brief introduction to their imaging techniques, data formats, and clinical applications.

\subsubsection{2D X-ray imaging}
\begin{figure*}[h]
	\centerline{\includegraphics[width=\linewidth]{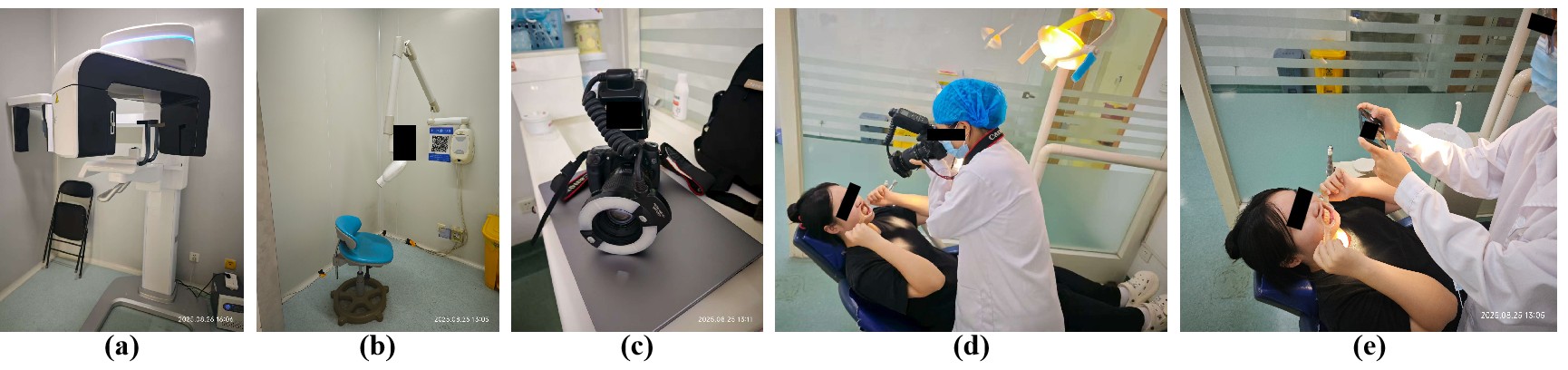}}
	\caption{Dental imaging devices utilized in actual clinical practice, featuring images captured at the Department of Stomatology, First Affiliated Hospital of Nankai University. (a) 2D panoramic radiography device; (b) periapical radiography device; (c) DSLR camera for intraoral photography; (d) radiologist taking intraoral photographs using a camera; (e) radiologist taking intraoral photographs using a mobile device.}
	\label{fig7}
\end{figure*}

2D X-ray imaging remains an essential and widely utilized modality in dental clinical diagnostics. It functions by transmitting X-rays through dental structures and adjacent tissues to produce images on a detector, which can be either film or a digital sensor \cite{rozylo2021panoramic}. Based on the acquisition method and imaging scope, 2D X-ray can be categorized into extraoral imaging (e.g., panoramic radiographs) and intraoral imaging (including periapical, bitewing, and occlusal radiographs). Fig \ref{fig7} (a) and (b) respectively display the PAN and PR imaging devices photographed in a clinical setting. As illustrated in Fig \ref{fig2}, panoramic radiographs provide a comprehensive view of the entire dentition and maxillofacial structures in one image, making them particularly suitable for orthodontic treatment planning and the extraction of impacted wisdom teeth \cite{chauhan2023overview}. Conversely, intraoral imaging, exemplified by PR, delivers detailed localized information, typically focusing on one or a few teeth or a specific region of the dentition, and is commonly employed for caries diagnosis and preoperative planning of root canal treatment \cite{vandenbergheModernDentalImaging2010, stera2024diagnostic}. For analyzing 2D X-ray imaging, images are frequently stored in JPG or PNG formats.

\subsubsection{Cone-Beam Computed Tomography (CBCT)}
\begin{figure}[h]
	\centerline{\includegraphics[width=\linewidth]{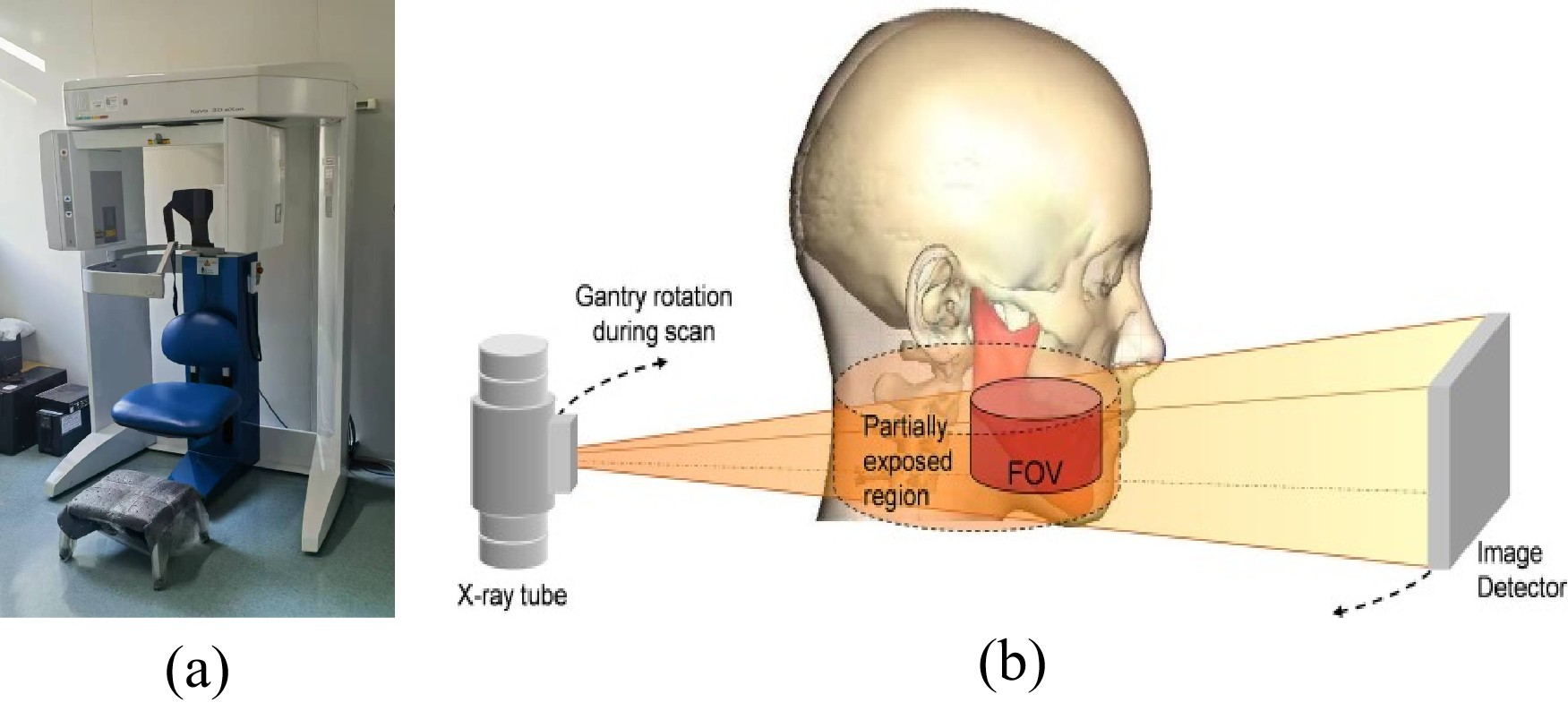}}
	\caption{(a) CBCT imaging device, captured at the Department of Stomatology, First Affiliated Hospital of Nankai University; (b) The fundamental principle of CBCT scanning involves a rotating gantry, an X-ray tube, and an image detector \cite{kaasalainen2021dental}.}
	\label{fig3}
\end{figure}
While 2D X-ray imaging is extensively utilized in dentistry, it presents certain limitations, such as distortions or tooth overlaps resulting from the projection of three-dimensional anatomical structures onto a two-dimensional plane \cite{vandenbergheModernDentalImaging2010}. In 1995, Mozzo et al. introduced the world's first CBCT prototype, the NewTom DVT 9000 \cite{mozzo1998new}, marking the official advent of CBCT imaging technology. As depicted in Fig. \ref{fig3}, with advanced hardware based on imaging physics and continually refined reconstruction algorithms, CBCT facilitates diverse three-dimensional visualizations and high-dimensional reconstructions of volumetric data, effectively addressing many limitations of 2D X-ray imaging \cite{kaasalainen2021dental}. Consequently, CBCT has seen increased application in dentistry. CBCT images can be analyzed using either 2D slices or 3D volumetric data. The raw CBCT data are usually stored in DICOM format \cite{mildenberger2002introduction}, a leading imaging standard in the medical field. Post-CBCT scan, a folder containing a series of consecutively numbered DICOM files is generated, recording not only imaging data but also basic patient information, such as sex and age. For computer science researchers, converting DICOM files into a more straightforward and efficient 3D data format, NIfTI (.nii), is preferred \cite{li2016first}. NIfTI-formatted data facilitate efficient preprocessing operations such as resampling, slicing, or patch extraction, enabling the creation of datasets tailored to various models and tasks. Open-source tools like ITK-Snap \cite{py06nimg} and 3D Slicer \cite{fedorov20123d} aid in the 3D visualization of NIfTI data and offer rich annotation interfaces to support the construction of deep learning datasets and model training. However, despite the growing sophistication of CBCT technology, it still faces certain limitations, including higher imaging costs, low contrast, and metal artifacts. Therefore, in clinical practice, physicians typically make flexible choices between CBCT and 2D X-ray imaging based on the specific conditions of the patient \cite{tiburcio2021global}.

\subsubsection{Intraoral Scanning (IOS)}
\begin{figure}[h]
	\centerline{\includegraphics[width=\linewidth]{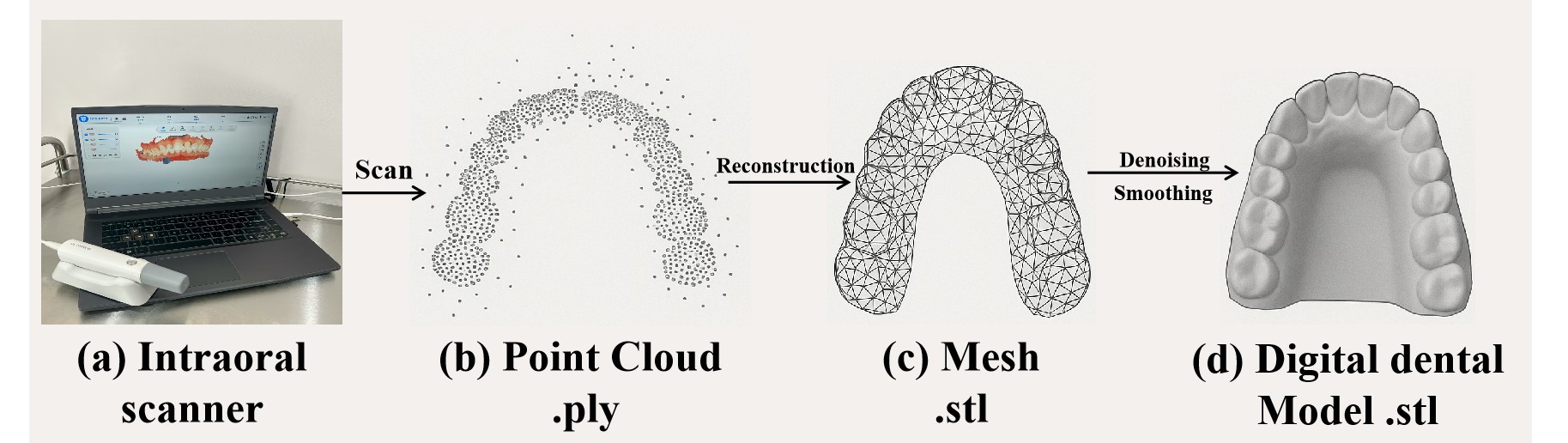}}
	\caption{The construction steps of a digital dental model: (a) Intraoral scanner; (b) The intraoral scanner captures raw data, stored in point cloud format (common file extensions such as .ply or .xyz); (c) Mesh model (typically in formats like .obj or .stl); (d) Digital dental model.}
	\label{fig4}
\end{figure}

Unlike the aforementioned X-ray–based radiographic imaging methods, IOS captures a 3D digital model of the teeth and oral tissues directly within the patient's mouth. This technique is extensively applied in various clinical scenarios, such as orthodontics and dental implantology \cite{eggmann2024recent}. IOS acts as a digital alternative to conventional impressions, providing faster and more accurate data acquisition while significantly enhancing patient comfort during the modeling process \cite{mangano2017intraoral}. The acquisition and processing workflow of IOS is illustrated in Fig. \ref{fig4}. Initially, a dentist uses a handheld scanning device to capture clinical point cloud data of the patient’s oral cavity (Fig. \ref{fig4}. a), which are then stored in point cloud format (Fig. \ref{fig4}. b). Subsequently, a mesh model is generated from the point cloud using triangulation or other reconstruction algorithms, consisting of vertices, edges, and faces (primarily triangular meshes, Fig. \ref{fig4}. c). Finally, the mesh model undergoes further processing and optimization to produce a digital dental model suitable for clinical use (Fig. \ref{fig4}. d).

\subsubsection{Intraoral Photography (IP)}
IP is conducted using imaging equipment, such as a DSLR camera equipped with a macro lens and flash, as illustrated in Fig. \ref{fig7} (c) and (d), to capture high-resolution 2D RGB images of the teeth, gingiva, and oral mucosa within the patient's mouth. Unlike other imaging modalities, IP offers a realistic representation of both soft and hard tissues, making it essential for oral health assessment \cite{guo2021feasibility}. Notably, IP is devoid of radiation exposure risks, while being cost-effective and easily performed. With minimal training, patients can utilize mobile devices to take images at home, as depicted in Fig. \ref{fig7} (e), rendering it highly suitable for remote dentistry and tele-diagnosis \cite{mao2025artificial}. Furthermore, because it visualizes lesion conditions directly without necessitating extensive radiological expertise, it facilitates smoother communication between patients and clinicians, aiding both parties in reaching a clearer consensus \cite{kulkarni2024comparative}. IP does not require specialized equipment, and its imaging process is akin to that of natural images. Consequently, the output is stored in standard image formats, most commonly with extensions such as '.jpg' or '.png'.

\begin{figure}[t]
	\centerline{\includegraphics[width=\linewidth]{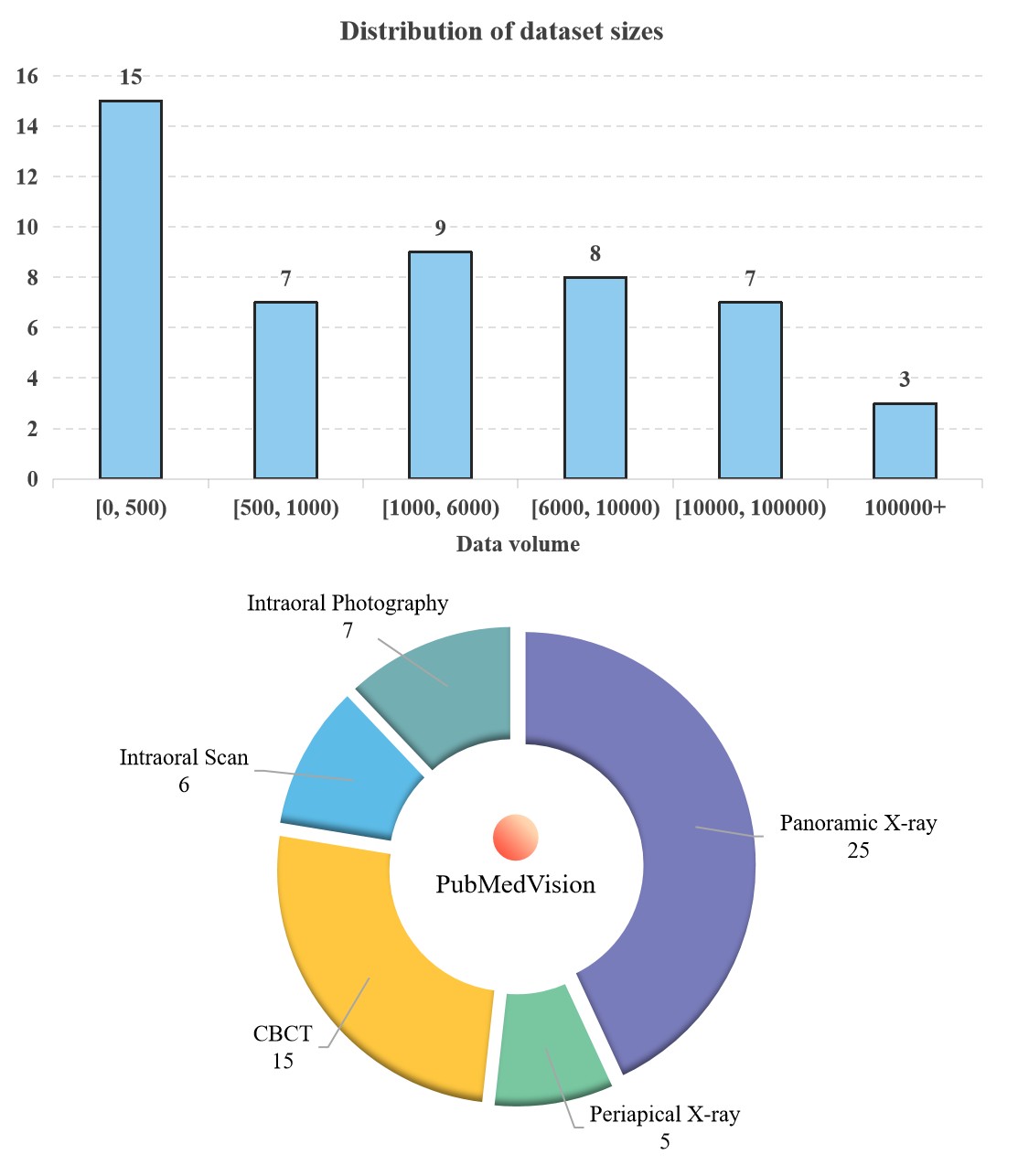}}
	\caption{The distribution of dataset information. The bar chart above illustrates the data volume distribution for each dataset, while the pie chart below depicts the distribution of dataset modalities (datasets containing multiple modalities are counted more than once). PubMedVision is a special case, as its image modalities include not only various dental imaging types but also statistical charts; hence, it is separately highlighted in the center of the pie chart.}
	\label{fig5}
\end{figure}

\subsection{Dental Public Datasets}
High-quality, large-scale public dental datasets are crucial for advancing DL research in DIA. Therefore, we have compiled 49 publicly available dental datasets in Table \ref{tab1}. These datasets originate from various regions and encompass a wide range of imaging modalities and research tasks. For each dataset, we also provide its acquisition method and link, facilitating efficient access and utilization of these resources by readers and researchers for study and research.

\begin{figure*}[t]
\centering
\includegraphics[width=\linewidth]{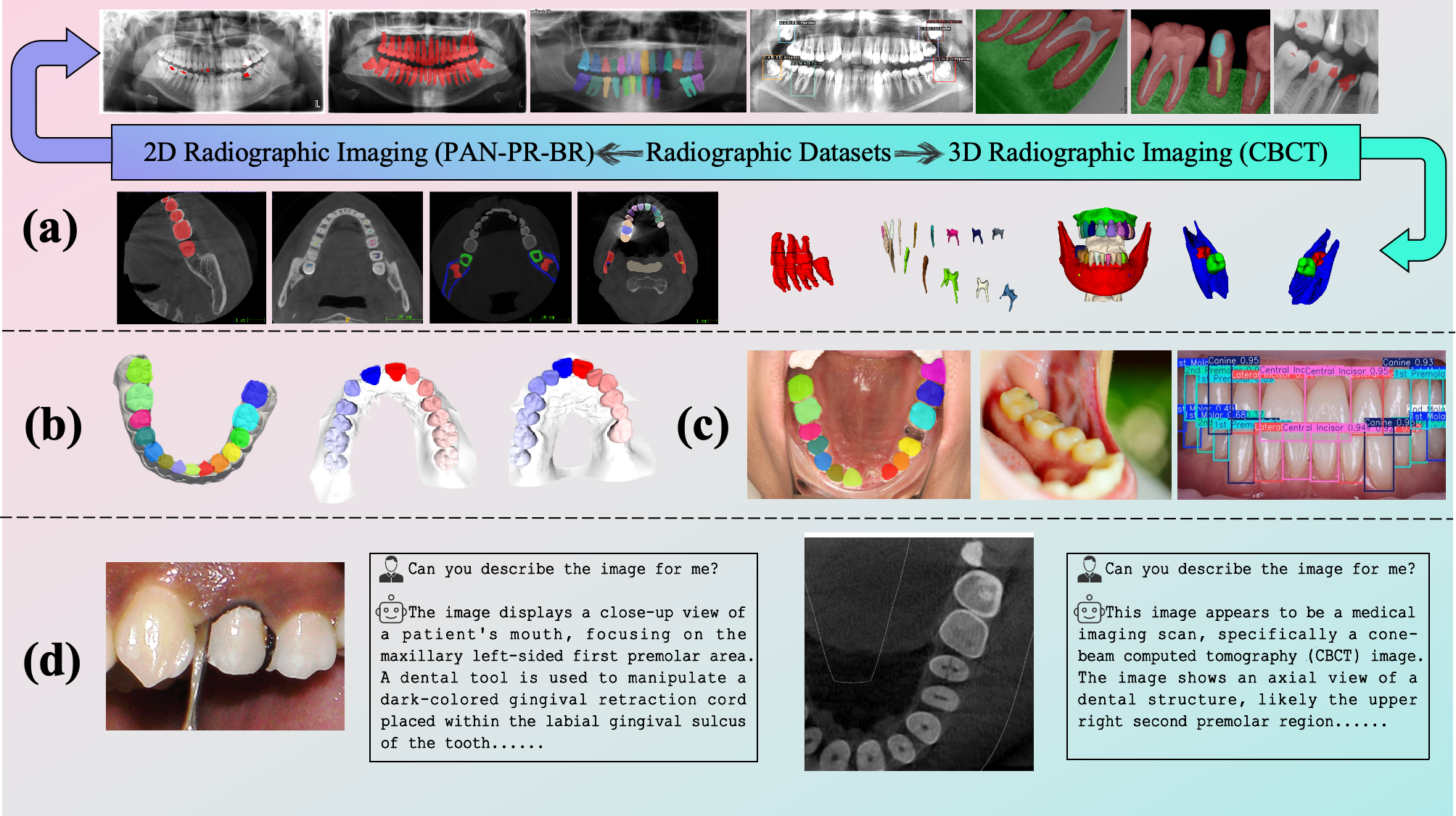}
\caption{Examples of data modalities and annotations in publicly available dental datasets, featuring sample data drawn from selected surveyed datasets: (a) dental radiographic imaging \cite{wang2023multi, HumansInTheLoop2023, panetta2021tufts, hamamci2023dentex, zhou2025prad, cui2022ctooth, gamal2024automatic, zhou2024nkut, bolelli2025segmenting}, (b) intraoral scanning (IOS) \cite{zou2024teeth, abenhamadou2023dteethseg}, (c) intraoral photography \cite{zou2024teeth, saisiddartha69_dental_anatomy_yolov8_2024}, and (d) dental image–text data represented by PubMedVision \cite{chen2024huatuogpt}.}
\label{fig6}
\end{figure*}

Fig \ref{fig5} illustrates the distribution of dataset sizes and modalities among the datasets evaluated in this study. It can be observed that the majority of datasets consist of fewer than 10,000 samples. Regarding modalities, 2D panoramic images are the most prevalent, followed by CBCT, while other modalities occur with a roughly one order of magnitude lower frequency. 2D radiographic imaging remains the most widely employed modality in dental imaging globally and is also the predominant imaging modality in this survey, accounting for more than 61.22\%. For instance, as shown in Fig. \ref{fig6} (a), Panetta et al. \cite{panetta2021tufts}, Hamamci et al. \cite{hamamci2023dentex}, Long et al. \cite{HumansInTheLoop2023}, and Wang et al. \cite{wang2025multi} have proposed clinical PAN datasets of varying sizes and sources for tooth semantic or instance-level segmentation tasks. Meanwhile, datasets by Li et al. \cite{li2024multi}, Seifossadat et al. \cite{seifossadat_dental_radiography_segmentation_2023}, Root Disease X-Ray \cite{engineeringubu_root_disease_xray_2024}, and OPG Kennedy \cite{orvile_dental_opg_kennedy_2024} focus on lesion classification or anomaly detection. In addition, Dental X-Ray \cite{rad2016digital}, Zhang et al. \cite{zhang2023children}, DC 1000 \cite{wang2023multi}, and Vipularya et al. \cite{kvipularya_dental_xray_analysis_2024} have introduced datasets dedicated to caries segmentation or lesion recognition in PAN images. Compared with PAN, publicly available PR datasets are considerably more limited, constituting only about 20\% of the volume of PAN data, indicating a potential shortage of datasets. For instance, the Dental X-Ray \cite{rad2016digital} offers a small-scale PR dataset for caries detection tasks; Root Disease X-Ray \cite{engineeringubu_root_disease_xray_2024} includes 6,578 PR images for root lesion classification; and PRAD \cite{zhou2025prad} assembles a relatively large dataset containing 10K PR images with multi-structure pixel-level annotations, of which 5K are publicly accessible upon request.

With the rising popularity of CBCT, more researchers have focused on constructing 3D radiographic databases, as shown in Fig. \ref{fig6} (a). CTooth \cite{cui2022ctooth} and CTooth+ \cite{cui2022ctooth+} are exemplary CBCT datasets developed for semi-supervised tooth semantic segmentation. Subsequently, Cui et al. \cite{cui2022fully} released a comprehensive multi-center CBCT dataset with instance-level annotations, establishing a robust benchmark for CBCT tooth segmentation. NKUT \cite{zhou2024nkut} specifically tackled pediatric tooth germ extraction by introducing a pediatric CBCT dataset for mandibular tooth germ and associated tissue segmentation. As the DIA field continues to evolve, segmenting individual teeth is increasingly insufficient for advanced dental CAD systems. Radiographic images inherently capture both teeth and maxillofacial anatomical structures, prompting researchers to develop datasets with finer-grained annotations and multi-task capabilities. These datasets not only enhance segmentation from the tooth level to the anatomical level but also broaden the scope of DIA applications. For example, ToothFairy \cite{bolelli2025segmenting} goes beyond tooth instance segmentation to include maxillofacial structures and mandibular nerve canal segmentation. Pulpy3D \cite{gamal2024automatic} and Liang et al. \cite{liang2024dual} introduced CBCT datasets for pulp segmentation. PRAD \cite{zhou2025prad} presented a large-scale PR dataset with segmentation annotations for nine anatomical structures or appliances. Dong et al. \cite{dong2023automatic} and DentalDS proposed PAN datasets for pediatric tooth maturity staging. These radiographic datasets have significantly advanced research in DIA.

In contrast to radiographic imaging, IP excels in directly capturing occlusal relationships and dental pathologies, enabling a more intuitive assessment of surface-level dental features that may not be clearly visible in radiographic modalities. Consequently, as shown in Fig. \ref{fig6} (c), public IP datasets primarily focus on tooth detection, segmentation, and disease classification tasks. For example, the Dental Anatomy \cite{saisiddartha69_dental_anatomy_yolov8_2024} dataset was introduced for tooth detection; Liu et al. \cite{liu2025fdtooth} and Huang et al. \cite{huang2025automatic} developed IP datasets targeting various types of dental cracks and fractures. The Tooth Decay dataset \cite{aminhj_tooth_decay_2023} offers a large collection of IP images for caries classification. Furthermore, Zou et al. \cite{zou2024teeth} proposed an IP dataset annotated with instance-level tooth segmentation. On the other hand, IOS supports more precise applications by generating high-resolution 3D digital dental models that accurately capture surface morphology. As shown in Fig. \ref{fig6} (b), these datasets are particularly valuable for orthodontic treatment planning and the fabrication of dental prostheses or aligners. Consequently, public IOS datasets are predominantly dedicated to precise tooth segmentation. For instance, IOS150K \cite{zou2024teeth} introduced a large-scale 2D IOS screenshot dataset for instance-level segmentation. In addition, 3D-IOSSeg \cite{li2024fine}, Teeth3DS+ \cite{abenhamadou2023dteethseg}, 

\begin{landscape}
\begin{table}[h]
\centering
\scriptsize
\setlength{\tabcolsep}{3pt}
\caption{Summary of Publicly Available Dental Datasets up to and including September 2025. \faDownload denotes datasets that can be directly downloaded via the provided link or upon registration, whereas \faPhone signifies datasets for which the download link must be acquired by contacting the paper's authors.}
\label{tab1}

\begin{tabular}{%
m{2.5cm} 
c 
c 
c 
c 
m{2.5cm} 
m{7cm} 
c 
c 
c 
}

\toprule
Dataset & Country & Year & Modality & Quantity & \multicolumn{1}{c}{Task} & \multicolumn{1}{c}{Overview} & Acquisition & License & URL \\
\midrule

Dental X-Ray \cite{rad2016digital} & Iran & 2016 & PR & 120 & Caries Detection
& PAN with annotations of 152 individual teeth affected by caries. & \faDownload & / & \href{https://mynotebook.labarchives.com/share/Vahab/MjAuOHw4NTc2Mi8xNi9UcmVlTm9kZS83NzM5OTk2MDZ8NTIuOA}{Link} \\
\midrule

TDD \cite{panetta2021tufts} & USA & 2021 & PAN & 1000 & Tooth Segmentation
& PAN with corresponding tooth masks, eye-tracking maps, audio transcription texts, and ROIs for the mandibular and maxillary areas. & \faPhone & CC BY 4.0 & \href{https://tdd.ece.tufts.edu/}{Link} \\
\midrule

Dental X-Ray Images \cite{dentalxray} & \makecell{South \\ Korea} & 2021 & PAN & 64 & \makecell[l]{Detection \\ Tooth Segmentation}
& PAN in PNG format, each accompanied by tooth detection and semantic segmentation annotations. & \faDownload & CC BY-NC-SA 4.0 & \href{https://www.kaggle.com/datasets/killa92/dental-x-ray-images-dataset}{Link} \\
\midrule

CTooth \cite{cui2022ctooth} & China & 2022 & CBCT & 22 & Tooth Segmentation
& CBCT in NIfTI format, pixel spacing: 0.25 × 0.25 $mm^{2}$, resolution: $256 \times 256$. & \faPhone & CC BY 4.0 & \href{https://github.com/liangjiubujiu/CTooth?tab=readme-ov-file}{Link} \\
\midrule

CTooth+ \cite{cui2022ctooth+} & China & 2022 & CBCT & 168 & Tooth Segmentation
& In addition to the CTooth dataset, 146 unlabeled CBCT scans were included. & \faPhone &  CC BY-NC 4.0 & \href{https://www.kaggle.com/datasets/weiweicui/ctooth-dataset}{Link} \\
\midrule

\cite{cui2022fully} & China & 2022 & CBCT & 150 & Tooth Segmentation
& The CBCT data are provided with instance-level tooth segmentation masks. & \faPhone & CC BY 4.0 & \href{https://github.com/ErdanC/Tooth-and-alveolar-bone-segmentation-from-CBCT}{Link} \\
\midrule

Vident-lab \cite{katsaros2022multi} & Poland & 2022 & \makecell{Dental- \\ Videos} & 1000 & Tooth Segmentation
& An asymmetrically annotated video dataset of natural teeth in phantom scenes is introduced for multi-task video processing, including tooth restoration, tooth segmentation, and inter-frame homography estimation. & \faDownload & CC BY-NC 4.0 & \href{https://mostwiedzy.pl/en/open-research-data/vident-lab-a-dataset-for-multi-task-video-processing-of-phantom-dental-scenes,624013248648956-0}{Link} \\
\midrule

ToothFairy \cite{cipriano2022deep} & Italy & 2022 & CBCT & 347 & IAN Segmentation
& Collected from Affidea Medical Center, the dataset provides both sparse and dense annotations, including IAN segmentation masks. & \faDownload & / & \href{https://ditto.ing.unimore.it/maxillo/}{Link} \\
\midrule

Panoramic Dental Xray \cite{brahmi2023panoramic} & Tunisia & 2023 & PAN & 107 & /
& 107 PAN images with a resolution of $2964 \times 1464$, without annotations. & \faDownload & CC BY 4.0 & \href{https://data.mendeley.com/datasets/73n3kz2k4k/2}{Link} \\
\midrule

\cite{zhang2023children} & China & 2023 & PAN & 100 & \makecell[l]{Caries Segmentation \\ Disease Detection} 
& pediatric PAN includes annotations for overall tooth structure segmentation and detection annotations based on dental disease classification. & \faDownload & CC BY 4.0 & \href{https://springernature.figshare.com/articles/dataset/Children_s_Dental_Panoramic_Radiographs_Dataset/21621705}{Link} \\
\midrule

Dentex \cite{hamamci2023dentex} & Switzerland & 2023 & PAN & 3,903 & \makecell[l]{Abnormal Teeth- \\ Detection and \\ Classification}.
& PAN from three different institutions, annotated with quadrants, tooth numbers, and types of dental lesions. & \faDownload & CC-BY 4.0 & \href{https://github.com/ibrahimethemhamamci/DENTEX}{Link} \\
\midrule

\cite{dong2023automatic} & China & 2023 & PAN & 673 & Maturity Staging
& Patients aged 3 to 14 years and include orthodontic brackets as well as slight blurring or deformation. Permanent tooth maturity was annotated using FDI tooth numbering in combination with the modified Demirjian method. & \faPhone & / & / \\
\midrule

Teeth3DS+ \cite{abenhamadou2023dteethseg} & Tunisia & 2023 & IOS & 1800 & Tooth Segmentation
& The dataset includes over 900 patients, featuring instance-level annotations of teeth along with key landmarks. & \faDownload & CC BY-NC-ND 4.0 & \href{https://github.com/abenhamadou/3DTeethSeg_MICCAI_Challenges}{Link} \\
\midrule

Dental Radiography \cite{imtkaggleteam_dental-radiography} & Iran & 2023 & PAN & 1272 & /
& Original PAN images without annotations and provide training, validation and test set grouping & \faDownload & CC BY-SA 4.0 & \href{https://www.kaggle.com/datasets/imtkaggleteam/dental-radiography?select=train}{Link} \\
\midrule

DC 1000 \cite{wang2023multi} & China & 2023 & PAN \& BR & 2986 & Caries Segmentation
& The dataset consists of 2,389 BR and 597 PAN, each accompanied by corresponding caries segmentation labels. & \faDownload & CC BY-SA 4.0 & \href{https://github.com/Zzz512/MLUA?tab=readme-ov-file}{Link} \\
\midrule

\cite{HumansInTheLoop2023} & Bulgaria & 2023 & PAN & 598 & Tooth Segmentation
& The dataset comprises 598 PAN with instance-level tooth segmentation labels. & \faDownload & CC0 1.0 license & \href{https://www.kaggle.com/datasets/humansintheloop/teeth-segmentation-on-dental-x-ray-images}{Link} \\
\midrule

\cite{pivoshenko_panoramic_mandibles_2023} & U.K. & 2023 & PAN & 116 & \makecell[l]{Mandible \\ Segmentation}
& This dataset includes 116 panoramic radiographs (PAN) with corresponding mandibular segmentation masks, covering a wide range of dental conditions from complete dentition to partial edentulism and full edentulism. & \faDownload & CC BY-SA 3.0 & \href{https://www.kaggle.com/datasets/volodymyrpivoshenko/panoramic-dental-x-rays-with-segmented-mandibles}{Link} \\
\midrule

MICCAI 2023 STS Challenge \cite{ricoleehduu_STS-Challenge_2023} & China & 2023 & CBCT \& PAN & 580+6500 & Tooth Segmentation
& This dataset contains images along with corresponding semantic segmentation masks of the tooth regions. & \faDownload & CC BY-NC-ND 4.0 & \href{https://zenodo.org/records/10597292}{Link} \\
\midrule
&  &  &  & & & & & \multicolumn{2}{c}{\textbf{\small(continued on next page)}}\\
\end{tabular}
\end{table}
\end{landscape} 

\begin{landscape}
\begin{table}[h]
\centering
\scriptsize
\setlength{\tabcolsep}{3pt}
\captionsetup{list=no} 
\caption*{Table \ref{tab1} (Continued from the last page. IAN: Inferior Alveolar Nerve)} 

\begin{tabular}{%
m{2.5cm} 
c 
c 
c 
c 
m{2.5cm} 
m{7.5cm} 
c 
c 
c 
}

\toprule
Dataset & Country & Year & Modality & Quantity & \multicolumn{1}{c}{Task} & \multicolumn{1}{c}{Overview} & Acquisition & License & URL \\
\midrule

Pulpy3D \cite{gamal2024automatic} & Egypt & 2024 & CBCT & 548 & \makecell[l]{Pulp Segmentation \\ IAN Segmentation}
& An extension of the ToothFairy, with annotations including both semantic and instance-level labels for the pulp and the IAN. 150 cases were manually annotated, while the remaining annotations were generated automatically by DL model. & \faDownload &CC BY-SA 4.0 &\href{https://github.com/mahmoudgamal0/Pulpy3D}{Link} \\
\midrule

IO150K \cite{zou2024teeth} & China & 2024 & IOS \& IP & 150K & Tooth Segmentation
& The dataset comprises 2D screenshots derived from IOS along with corresponding instance-level annotations of teeth, all of which were generated using a human-in-the-loop annotation approach. & \faPhone & CC BY-SA 4.0 & \href{https://zoubo9034.github.io/TeethSEG/}{Link} \\
\midrule

Multimodal Dental Dataset \cite{huang2024multimodal} & China & 2024 & \makecell{CBCT \\ PAN} & \makecell{8 \\ 329} & Multimodal learning & The dataset contains multi-modal dental images from 169 patients, where 16,203 PR were not directly captured but generated from CBCT scans. & \faDownload &  CC BY 4.0 & \href{https://physionet.org/content/multimodal-dental-dataset/1.1.0/}{Link} \\
\midrule

Orthodontic Dental Dataset \cite{wang20243d} & China & 2024 & IOS & 1060 & \makecell[l]{Tooth Segmentation \\ Landmarks Detection} & The dataset includes 1,060 pre- and post-treatment dental arch models from 435 patients, along with tooth instance segmentation annotations and crown landmark annotations. & \faPhone & CC 0 & \href{https://zenodo.org/records/11392406}{Link} \\
\midrule

NKUT \cite{zhou2024nkut} & China & 2024 & CBCT & 133 & \makecell[l]{Tooth Segmentation \\ Bone Segmentation}
& This dataset contains pediatric CBCT images with segmentation annotations for mandibular third molar germs, second molars, and the alveolar bone. & \faDownload &  CC BY-NC 4.0 & \href{https://data.mendeley.com/datasets/c4hhrkxytw/4}{Link} \\
\midrule

Dental OPG X-RAY \cite{rahman2024comprehensive} & Bangladesh & 2024 & PAN & 232  & Lesion detection \& classification.
& This dataset includes PAN and classifies teeth into six categories: healthy teeth, dental caries, impacted teeth, infections, tooth fractures, and broken crowns or roots.  & \faDownload & CC BY-NC 4.0 & \href{https://data.mendeley.com/datasets/c4hhrkxytw/4}{Link} \\
\midrule

\cite{li2024multi} & China & 2024 & PAN & 6536 & Lesion detection \& classification
& This dataset consists of three sets of data collected from different hospitals, covering dental conditions such as impacted teeth, periodontitis, and dental caries, with annotations including segmentation masks and classification labels. & \faDownload & CC BY-NC 4.0 & \href{https://github.com/qinxin99/qinxini}{Link} \\
\midrule

Tooth decay \cite{aminhj_tooth_decay_2023} & Iran & 2024 & PR \& IP & 8579+ & Lesion classification & This dataset covers the detection and analysis of dental caries, including images of decayed and damaged teeth. & \faDownload & MIT & \href{https://www.kaggle.com/datasets/aminhj/tooth-decay}{Link} \\
\midrule

\cite{seifossadat_dental_radiography_segmentation_2023} & Iran & 2024 & PAN & 29591 & Lesion classification & This dataset covers the detection and analysis of dental caries, including images of decayed and damaged teeth. & \faDownload & CC BY-SA 4.0 & \href{https://www.kaggle.com/datasets/abbasseifossadat/dental-radiography-segmentation}{Link} \\
\midrule

Dental Anatomy \cite{saisiddartha69_dental_anatomy_yolov8_2024} & India & 2024 & IP & 724 & Tooth detection & used for tooth detection and is divided into training, validation, and test sets, containing 505, 112, and 107 images with corresponding annotations in each folder, respectively. & \faDownload & CC BY-SA 4.0 & \href{https://www.kaggle.com/datasets/saisiddartha69/dental-anatomy-dataset-yolov8}{Link} \\
\midrule

\cite{lokisilvres_dental_disease_panoramic_detection_2024} & India & 2024 & PAN & 13932 & Lesion Detection & This dataset contains a large number of panoramic X-ray images (PAN) along with detection labels for various dental diseases, annotated in both YOLO and COCO formats. & \faDownload & Apache 2.0 & \href{https://www.kaggle.com/datasets/lokisilvres/dental-disease-panoramic-detection-dataset}{Link} \\
\midrule

\cite{kvipularya_dental_xray_analysis_2024} & India & 2024 & PAN & 70 & Tooth Segmentation
& The data are high-quality samples obtained from local hospitals and trusted acquaintances, accompanied by semantic segmentation labels for the teeth. & \faPhone & CC BY-SA 3.0 & \href{https://www.kaggle.com/datasets/kvipularya/a-collection-of-dental-x-ray-images-for-analysis}{Link} \\
\midrule

Root Disease X-Ray \cite{engineeringubu_root_disease_xray_2024} & Thailand & 2024 & PR & 6578 & \makecell[l]{Disease \\ Classification}
& PR used for dental disease classification.  & \faDownload & CC BY-NC-SA 4.0 & \href{https://github.com/abenhamadou/3DTeethSeg_MICCAI_Challenges}{Link} \\
\midrule

PubMedVision \cite{chen2024huatuogpt} & China & 2024 & Various & 14212 & \makecell[l]{Image-Text \\ Pretraining} & A medical image-text paired dataset containing 646,759 image-text pairs, of which 14,212 are related to dentistry, featuring diverse formats and including both Chinese and English bilingual data. & \faDownload & / & \href{https://huggingface.co/datasets/FreedomIntelligence/PubMedVision}{Link} \\
\midrule

3D-IOSSeg \cite{li2024fine} & China & 2024 & IOS & 440 & Tooth Segmentation
& The dataset includes IOS from 220 patients, covering various dental diseases with instance-level segmentation annotations, and incorporates extreme dental conditions to enhance model robustness. & \faDownload & / & \href{https://reurl.cc/0vjLXY}{Link} \\
\midrule

MICCAI 2024 STS Challenge \cite{sts2024_organizers} & China & 2024 & \makecell{CBCT \\ PAN} & \makecell{350 \\ 2400} & Tooth Segmentation & The 2nd STS Challenge dataset includes paired panoramic radiographs (PAN) and CBCT scans, with instance-level tooth segmentation annotations. & \faPhone & / & \href{https://STS-challenge.github.io}{Link} \\
\midrule

\cite{liang2024dual} & China & 2024 & CBCT & 300 & \makecell[l]{Tooth Segmentation \\ Pulp Segmentation }
& CBCT scans from 300 anonymous patients, with semantic-level segmentation masks for teeth and root canals. & \faPhone & / & \href{https://github.com/WANG-BIN-LAB/D3UNet}{Link} \\
\midrule

&  &  &  & & & & & \multicolumn{2}{c}{\textbf{\small(continued on next page)}}\\
\end{tabular}
\end{table}
\end{landscape} 

\begin{landscape}
\begin{table}[h]
\centering
\scriptsize
\setlength{\tabcolsep}{3pt}
\captionsetup{list=no} 
\caption*{Table \ref{tab1} (Continued from the last page.)} 

\begin{tabular}{%
m{2.5cm} 
c 
c 
c 
c 
m{2.5cm} 
m{7.5cm} 
c 
c 
c 
}

\toprule
Dataset & Country & Year & Modality & Quantity & \multicolumn{1}{c}{Task} & \multicolumn{1}{c}{Overview} & Acquisition & License & URL \\
\midrule

TRPR \cite{silva2025holistic} * & Brazil & 2025 & PAN & 8029 & \makecell{Image–Text \\ Pretraining}
& The TRPR dataset consists of PAN and their corresponding textual reports, without tooth segmentation annotations. The reports were written in Portuguese by two dental experts. & \faPhone & / &\href{https://github.com/IvisionLab/MEDIA-datasets}{Link} \\
\midrule

MMDental \cite{wang2025mmdental} & China & 2025 & CBCT & 660 & \makecell{Image–Text \\ Pretraining}
& MMDental dataset includes CBCT and corresponding detailed expert medical records, along with both initial and follow-up records. & \faDownload & CC BY-NC-ND 4.0 & \href{https://doi.org/10.6084/m9.figshare.28505276}{Link} \\
\midrule

FDTooth \cite{liu2025fdtooth} & HK, China & 2025 & CBCT \& IP & 241 & \makecell{Fenestration and \\ Dehiscence \\ Detection} & FDTooth dataset consists of paired IP and CBCT scans of both children and adults, including 2,892 annotated bounding boxes of anterior teeth with fenestration or Dehiscence. & \faPhone &  \href{https://physionet.org/content/fdtooth/view-license/1.0.0/}{PhysioNet 1.5} & \href{https://physionet.org/content/fdtooth/1.0.0/}{Link} \\
\midrule

PRAD \cite{zhou2025prad} & China & 2025 & PR & 10K & Anatomical-level Tooth Segmentation 
& The PRAD dataset includes real clinical PR along with corresponding anatomical-level segmentation masks, covering nine types of tooth structures, lesions, and instruments. & \faPhone & CC BY-NC 4.0 & \href{https://github.com/nkicsl/PRAD}{Link} \\
\midrule

DentalDS \cite{wang2025towards}  & China & 2025 & PAN & 2583 & Tooth development grading & DentalDS dataset contains images of pediatric orthodontic patients, with annotations including bounding boxes for the mandibular teeth and developmental stage labels based on the Demirjian method, totaling 18,081 annotated teeth. & \faDownload &  CC CC BY-NC-ND 4.0 & \href{ https://github.com/ybupengwang/DDSNet}{Link} \\
\midrule

STS-Tooth \cite{wang2025multi} & China & 2025 & \makecell{PAN \\ CBCT} & \makecell{4K \\ 148K}  & \makecell{Tooth \\ Segmentation}. & This dataset includes multimodal PAN and CBCT data from both pediatric and adult patients, providing 900 PAN images and 8,800 CBCT scans with corresponding tooth semantic segmentation masks.  & \faDownload & CC BY-NC-ND 4.0 & \href{https://zenodo.org/records/14827784}{Link} \\
\midrule

\cite{faizan2025annotated} & Pakistan & 2025 & IP & 6313 & Caries Detection & This dataset includes high-resolution IP of children and adolescents, accompanied by dental caries detection annotations in multiple formats. & \faDownload & CC BY-NC-ND 4.0 & \href{https://zenodo.org/records/14827784}{Link} \\
\midrule

ToothFairy2 \cite{bolelli2025segmenting} & Italy & 2025 & CBCT & 530 & Maxillofacial Structure Segmentation & The dataset includes patients' CBCT scans along with voxel-level 3D instance annotations corresponding to 42 different maxillofacial structure categories. & \faDownload & CC BY-SA & \href{https://ditto.ing.unimore.it/toothfairy2/}{Link} \\
\midrule

\makecell[l]{OPG Kennedy \\ \cite{orvile_dental_opg_kennedy_2024}} & \makecell{Saudi \\ Arabia} & 2025 & PAN & 622 & Lesion Classification & The dataset includes panoramic X-ray images (PAN) with corresponding bounding box annotations, covering categories such as Broken Roots, Periodontally Compromised Teeth, and Kennedy Classification of Partially Edentulous Arches (Class I-IV). & \faDownload & CC BY 4.0 & \href{https://www.kaggle.com/datasets/orvile/dental-opg-kennedy-dataset}{Link} \\
\midrule

\cite{ali_dental_xray_panorama_segmentation_2024} & Egypt & 2025 & PAN & 8188 & Lesion Segmentation & The dataset includes panoramic X-ray images (PAN) with semantic-level segmentation masks for various lesions, without distinguishing between specific disease types. & \faDownload & / & \href{https://www.kaggle.com/datasets/mohamedali020/dental-x-raypanoramic-semanticsegmentation-task}{Link} \\
\midrule

\cite{kvipularya_dental_xray_analysis_2024} & Egypt & 2025 & PAN & 335 & Caries Detection & This dataset is adapted from other public datasets for detecting dental caries in children and includes bounding box annotations. & \faDownload  & MIT&\href{https://www.kaggle.com/datasets/mariamosamakhalifa/children-caries-detection-dataset}{Link} \\
\midrule

MICCAI 2025 STS \cite{sts2025_organizers} & China & 2025 & \makecell{CBCT \\ IOS} & 780 & Tooth \& Pulp segmentation, CBCT-IOS Registration.
& The segmentation subset provides semantic masks of teeth and pulp in CBCT images; the registration subset includes 380 CBCT cases paired with corresponding IOS, along with affine transformation matrices. & \faPhone & / & \href{https://songhen15.github.io/STSdevelop.github.io/miccai2025/index.html}{Link} \\
\midrule

\cite{huang2025automatic} & China & 2025 & IP & 2210 & \makecell[l]{Tooth Crack \\ Detection} & Artificial cracks were created on 432 teeth, and high-quality images were selected after photographing with a standard camera, covering multiple types of cracks. & \faDownload & / & \href{https://github.com/YCHuang18/ToothCrack}{Link} \\
\midrule

\cite{sehar2025automatic} & China & 2025 & CBCT & 105 & Tooth Segmentation
& The dataset includes CBCT scans from 105 patients, totaling approximately 14,000 tooth samples, annotated according to the FDI tooth numbering system. & \faDownload & LLC protocol & \href{https://www.embodi3d.com/files/category/41-dental-orthodontic-maxillofacial-cts/}{Link} \\
\midrule
&  &  &  & & & & & \multicolumn{2}{c}{\textbf{\small(End of Table \ref{tab1})}}\\
\bottomrule
\end{tabular}
\end{table}
\end{landscape} 
\noindent and Wang et al. \cite{wang20243d} released datasets of varying scales in 3D point cloud format for instance-level tooth segmentation on IOS data.

With the recent rise of vision-language models (VLM) in medical image analysis, several researchers have proposed image-text datasets tailored for dental applications. As illustrated in Fig. \ref{fig6} (d), PubMedVision \cite{chen2024huatuogpt} is constructed from carefully curated images and figure captions extracted from PubMed articles and medical textbooks. Among its more than 640,000 image-text pairs, approximately 14,212 pairs pertain to dentistry. The dataset includes diverse modalities, such as radiographs, IOS, IP, intraoperative color images, and statistical figures. In 2025, TRPR \cite{silva2025holistic} released a PAN dataset with images and corresponding Portuguese reports, intended for lesion classification and VLM pretraining. MMDental \cite{wang2025mmdental} introduced a multimodal dataset consisting of 403 CBCT volumes and 660 clinical cases, encompassing both initial and follow-up records, designed for multimodal learning.

Notably, since 2023, MICCAI has organized the Semi-supervised Teeth Segmentation (STS) Challenge for three consecutive years, underscoring the community’s growing focus on DIA. The STS tasks and data modalities include semantic- and instance-level segmentation of teeth on PAN and CBCT images \cite{ricoleehduu_STS-Challenge_2023, sts2024_organizers}, as well as semantic segmentation of teeth and pulp on CBCT, and CBCT-IOS registration \cite{sts2025_organizers}. These diverse publicly available dental datasets and community challenges, spanning multiple modalities and tasks, have established a robust foundation for applying deep learning in DIA.

\begin{figure*}[h]
	\centerline{\includegraphics[width=\linewidth]{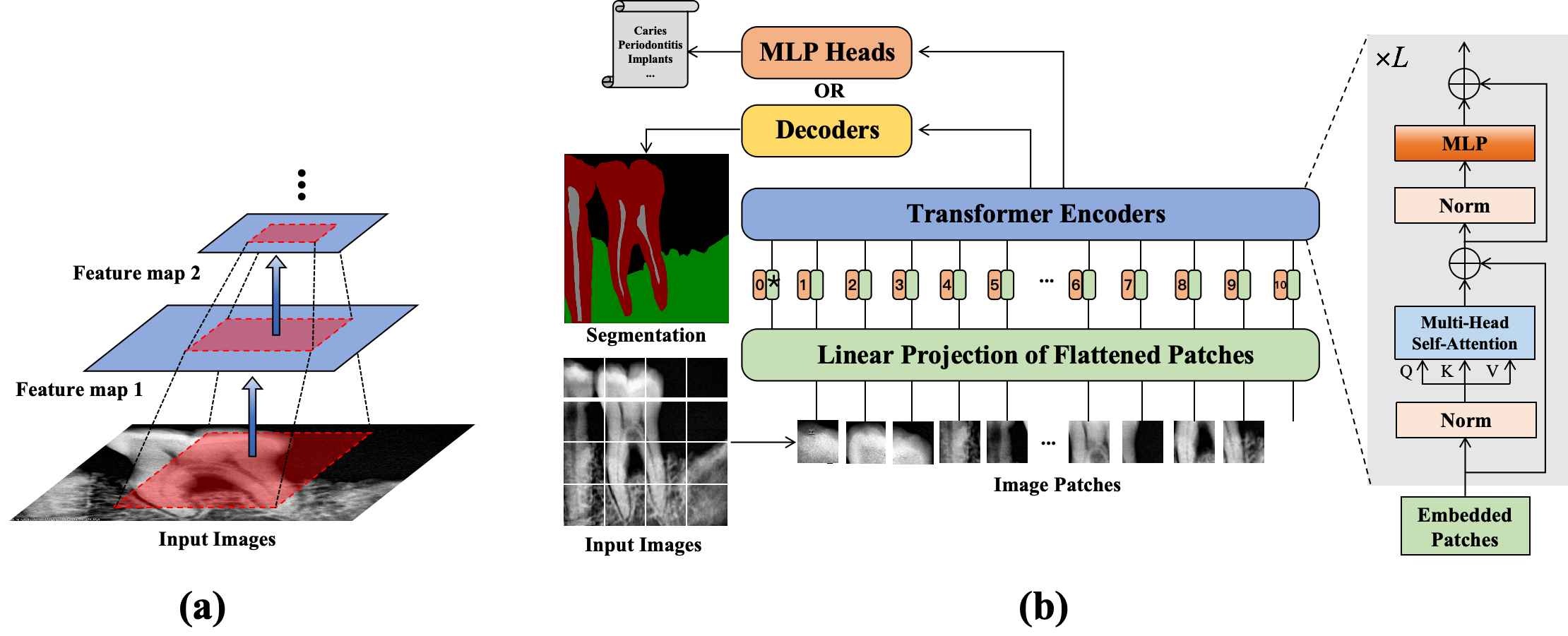}}
	\caption{Illustration of fundamental model architectures in the DIA field: (a) depicts the schematic of a CNN, where convolutional kernels extract local features and progressively reduce the resolution of feature maps through hierarchical processing; (b) shows a ViT-based network structure, where the input image is divided into patches and fed into a ViT encoder to encode global features via self-attention mechanisms, followed by a decoder or classification head to generate task-specific outputs. The PR image shown in the figure is sourced from \cite{zhou2025prad}.}
	\label{fig8}
\end{figure*}

\section{Basic technologies}
\label{sec3}
\subsection{Overview}
With the continuous accumulation of data resources and growing clinical demands, the application of DL in DIA has become increasingly widespread, fostering the development of deep models based on diverse network architectures. We categorize DIA tasks into three major groups: dense prediction tasks, classification tasks, and other tasks. Among these, dense prediction tasks, primarily represented by segmentation, constitute the dominant direction of image processing in the dental field, making up the vast majority of the surveyed literature (182 papers, $86.26\%$). Studies indicate that precise segmentation of teeth, anatomical structures, and lesions in dental images serves as the essential foundation and prerequisite for all other tasks. In this section, we first introduce several fundamental DL network models, upon which most of the surveyed models are developed as variants or through fusion. Subsequently, we provide a comprehensive review and discussion of 211 related papers, organized by task categories.

\subsection{Fundamental Techniques}
DL-based approaches for DIA can generally be categorized into three types: CNN-based methods, Transformer-based methods, and hybrid methods that integrate the strengths of both. The following subsections provide a detailed introduction to each.

\subsubsection{CNN}
The basic structure of a CNN is illustrated in Fig. \ref{fig8}. In the early 21st century, as modern CNNs achieved breakthrough success in handwritten digit recognition tasks \cite{lecun2002gradient}, CNNs gradually evolved into a cornerstone technology in the field of computer vision. Since then, CNNs have advanced rapidly, significantly surpassing traditional approaches in the ImageNet classification task \cite{krizhevsky2012imagenet}. The network depth has increased \cite{simonyan2014very}, while multi-scale feature extraction capabilities and scalability have been greatly enhanced \cite{szegedy2015going}. The residual learning framework introduced by ResNet \cite{he2016deep} established a paradigm for deep models, effectively addressing the issues of vanishing and exploding gradients in very deep networks. Meanwhile, the advent of Fully Convolutional Networks (FCN) \cite{long2015fully} enabled end-to-end semantic segmentation for the first time, equipping CNNs with pixel-level classification capability. In the field of medical image analysis, UNet \cite{ronneberger2015u} stands as the most influential deep learning model, originally proposed for cell segmentation in microscopic images. Its symmetric encoder–decoder structure and skip connections are crucial for capturing details and boundaries in medical images, which has led to its widespread adoption across diverse medical domains \cite{chen2025medical, mirikharaji2023survey}. Numerous variants have subsequently been proposed, such as UNet++ \cite{zhou2019unet++}, Attention-UNet \cite{oktay2018attention}, 3D-UNet \cite{cicek3DUNetLearning2016}, and VNet \cite{milletariVNetFullyConvolutional2016}. In the dental domain, CNNs are likewise the most widely applied core technology, with a large body of research in various DIA tasks developed on the foundation of UNet and its variants.

\subsubsection{Transformer}
 Transformer \cite{vaswani2017attention} was initially introduced in 2017 for machine translation and has since become a cornerstone of natural language processing. The key component of the Transformer is the self-attention mechanism, which enables the model to learn the relevance of each token to all other tokens and perform feature weighting accordingly. This mechanism endows the model with the ability to capture global features while effectively modeling long-range dependencies. In 2020, with the introduction of Vision Transformer (ViT) \cite{dosovitskiy2020image}, the Transformer architecture officially entered the field of computer vision, achieving remarkable success. A significant improvement over ViT, the Swin Transformer \cite{liuSwinTransformerHierarchical2021} introduces a hierarchical structure to mimic the pyramidal data flow of CNNs, making it more suitable for downstream dense prediction tasks, and employs the Shifted Window Attention mechanism to reduce computational overhead. In the domain of medical image analysis, ViT-based models quickly emerged and demonstrated outstanding performance across a wide range of tasks in different medical specialties \cite{shamshad2023transformers}. TransUnet \cite{chen2021transunet} is one of the earliest models to incorporate ViT into medical image analysis, enabling efficient segmentation of abdominal organs by integrating a ViT encoder into its architecture. Swin-UNet \cite{cao2022swin} is a purely ViT-based model built upon the Swin Transformer's pyramid-shaped hierarchical structure, demonstrating strong competitiveness in medical image segmentation tasks. MissFormer \cite{huang2021missformer} further improves upon this by incorporating multi-scale feature extraction and attention mechanisms, exhibiting higher robustness and stronger generalization in medical image segmentation. In dental image analysis, Transformers have gradually reached parity with CNNs and even surpassed them in certain tasks.
 
\subsubsection{Hybrid Model}
Transformer \cite{vaswani2017attention} was initially introduced in 2017 for machine translation and has since become a cornerstone of natural language processing. The key component of the Transformer is the self-attention mechanism, which enables the model to learn the relevance of each token to all other tokens and perform feature weighting accordingly. This mechanism endows the model with the ability to capture global features while effectively modeling long-range dependencies. In 2020, with the introduction of Vision Transformer (ViT) \cite{dosovitskiy2020image}, the Transformer architecture officially entered the field of computer vision, achieving remarkable success. A significant improvement over ViT, the Swin Transformer \cite{liuSwinTransformerHierarchical2021} introduces a hierarchical structure to mimic the pyramidal data flow of CNNs, making it more suitable for downstream dense prediction tasks, and employs the Shifted Window Attention mechanism to reduce computational overhead. In the domain of medical image analysis, ViT-based models quickly emerged and demonstrated outstanding performance across a wide range of tasks in different medical specialties \cite{shamshad2023transformers}. TransUnet \cite{chen2021transunet} is one of the earliest models to incorporate ViT into medical image analysis, enabling efficient segmentation of abdominal organs by integrating a ViT encoder into its architecture. Swin-UNet \cite{cao2022swin} is a purely ViT-based model built upon the Swin Transformer's pyramid-shaped hierarchical structure, demonstrating strong competitiveness in medical image segmentation tasks. MissFormer \cite{huang2021missformer} further improves upon this by incorporating multi-scale feature extraction and attention mechanisms, exhibiting higher robustness and stronger generalization in medical image segmentation. In dental image analysis, Transformers have gradually reached parity with CNNs and even surpassed them in certain tasks.

\section{Dental Tasks and Models}
\label{sec4}
\subsection{Dense prediction}
With the rapid development and increasing maturity of digital dentistry, dense prediction tasks, particularly segmentation, have emerged as the most widely studied, longest-standing, and most extensively explored direction in the DIA field. Efficient segmentation and detection of dental and maxillofacial structures across various imaging modalities not only significantly reduce manual labor and provide more objective diagnostic support but also serve as a fundamental basis for numerous preoperative dental applications \cite{dutra2016diagnostic}, such as orthodontic surgery planning \cite{zou2024teeth}, endodontic treatment planning \cite{wangRootCanalTreatment2023}, and tooth extraction and implant planning \cite{ma2023accurate}. In this section, we provide a systematic review of the various attempts and advancements in applying ensemble DL models to dense prediction tasks in DIA. The overall categorization framework is illustrated in Fig. \ref{fig2}.

\subsubsection{Tooth}
Tooth dense prediction is one of the core tasks in DIA and is often employed as a crucial preprocessing step for subsequent tasks. The application of CNN-based algorithms to DIA can be traced back to around 2016 \cite{eun2016oriented, choi2018boosting}, since which time DL has seen increasingly widespread use in DIA.
\begin{figure}[t]
	\centerline{\includegraphics[width=\linewidth]{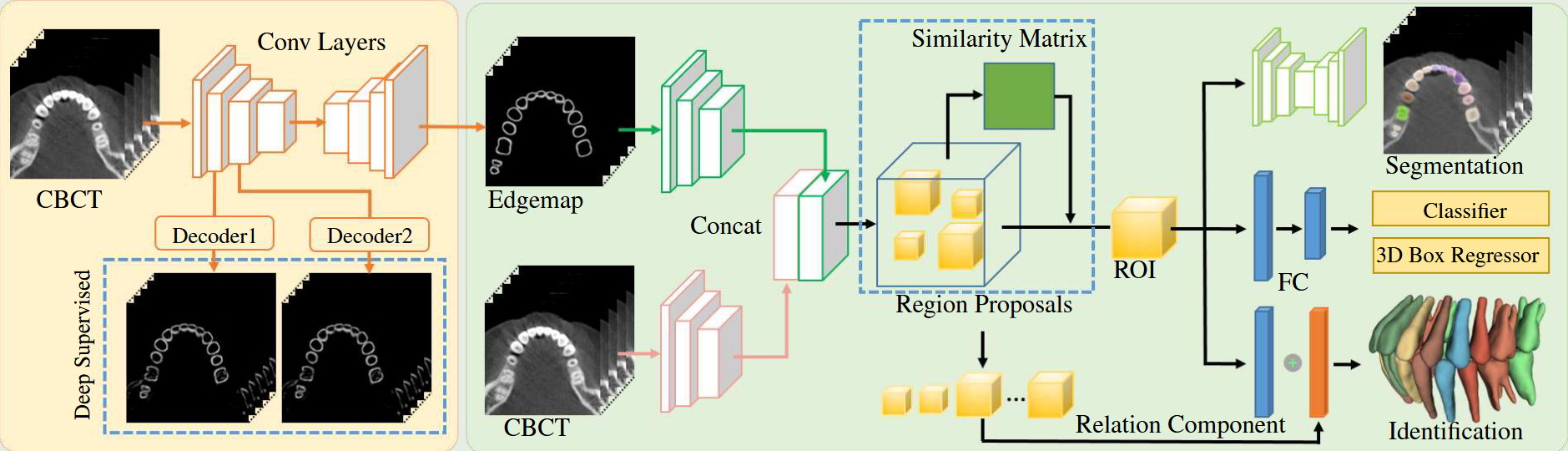}}
	\caption{Overview of ToothNet. Built upon the 3D Mask R-CNN backbone, it incorporates boundary maps and similarity matrices to assist in instance-level tooth segmentation on CBCT images. ToothNet represents the first pioneering work employing DL techniques for CBCT-based tooth segmentation. Image sourced from \cite{cui2019toothnet}.}
	\label{fig10}
\end{figure}

\paragraph{\textbf{Tooth Level}}
We define the prediction of teeth as complete entities, without distinguishing their internal anatomical structures, as tooth-level tasks, as shown in Fig. \ref{14} (a)-(f). Tooth-level tasks form the foundation of all DIA research and account for the vast majority of existing studies, as summarized in Tables \ref{tab2} and \ref{tab3}. Based on the granularity of representation, these tasks can be further categorized into two types: semantic-level \cite{majanga2019deep, sivagami2020unet, feng2023automated, zhao2021two} and instance-level \cite{zhao20213d, lin2023dbganet, zhao2024few, xu2025tooth}. The former treats all teeth as a single unified region, whereas the latter distinguishes each tooth as an independent instance. In clinical practice, the FDI tooth numbering system, as illustrated in Fig. |\ref{fig9} (a), is commonly used to facilitate more fine-grained analysis and diagnosis.

\begin{figure}[t]
	\centerline{\includegraphics[width=\linewidth]{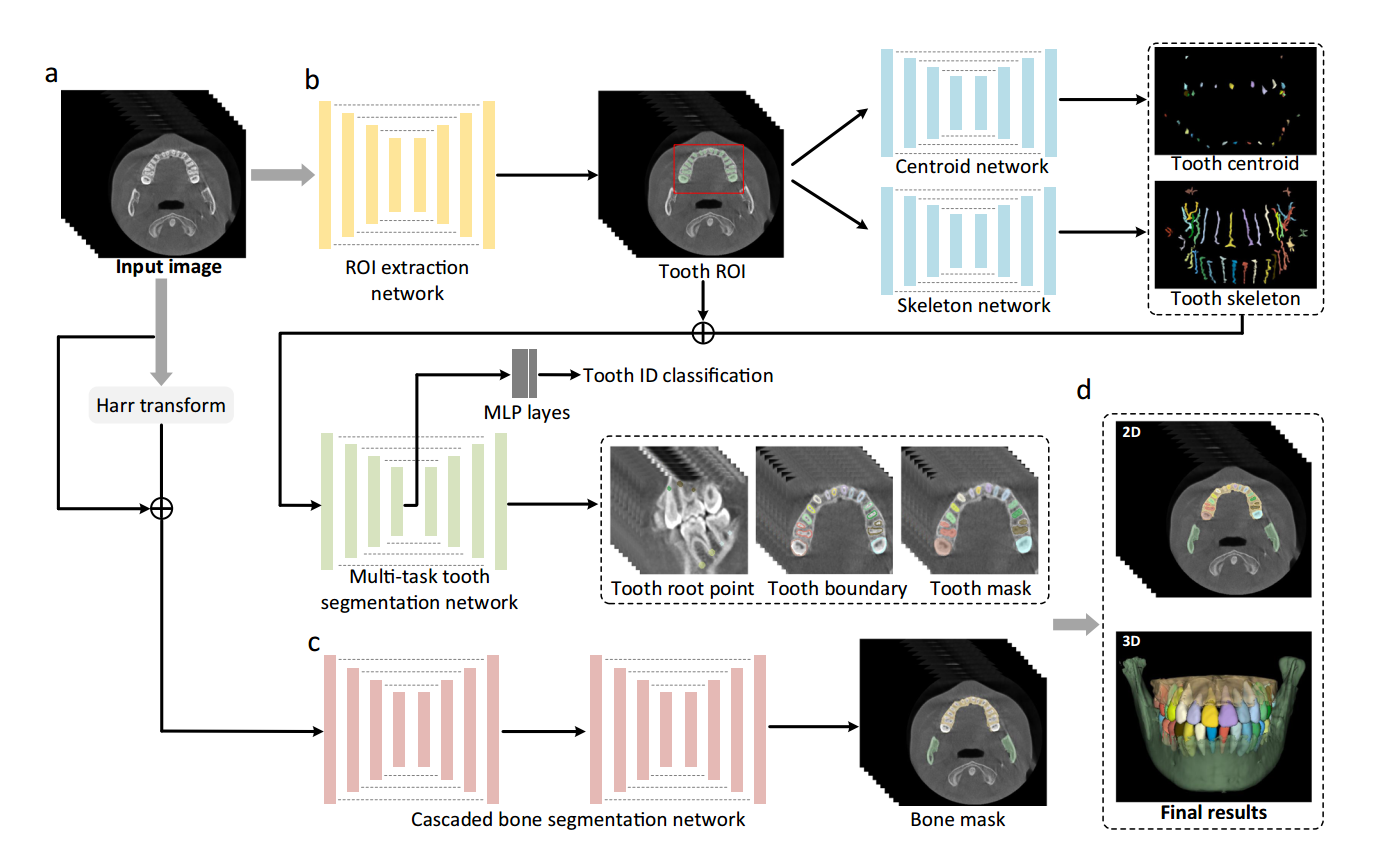}}
	\caption{Overview of a multi-task CBCT segmentation framework. It introduces tooth centroid prediction and tooth skeleton prediction as auxiliary tasks to enhance the accuracy of tooth numbering and localization, ultimately achieving precise segmentation of teeth and jaw structures in CBCT images. Image sourced from \cite{cui2022fully}.}
	\label{fig11}
\end{figure}

Prior to 2023, CNNs were the predominant architectures employed for tooth-level dense prediction tasks. As summarized in Tables \ref{tab2}, during this stage, many researchers also initiated various explorations of DL-based DIA, leading to a wide range of research efforts. In 2019, Cui et al. proposed ToothNet \cite{cui2019toothnet}, the first approach to apply DL to CBCT tooth instance segmentation. As shown in Fig. \ref{fig10}, ToothNet employs a two-stage architecture: in the first stage, a boundary extraction network is used to capture boundary information from the input; in the second stage, the extracted boundary features are fused with the original input through a similarity matrix and subsequently fed into a 3D region proposal network to perform segmentation. This multi-stage concept has also provided valuable insights for subsequent studies. For instance, Yang et al. \cite{yang2021accurate} combined traditional algorithms with DL by first using U-Net to localize tooth centroids, followed by level-set algorithms to achieve precise CBCT tooth segmentation. Wu et al. \cite{wu2022two} proposed a two-stage approach integrating iMeshSegNet and PointNet to accomplish tooth segmentation and landmark detection on IOS data. Jing et al. \cite{jing2024usct} adopted a similar two-stage strategy, where a coarse segmentation network was applied to CBCT images to obtain tooth ROIs, which were then refined using an improved V-Net-based segmentation network. Moreover, as shown in Fig. \ref{fig11}, Cui et al. further developed an automated CBCT tooth segmentation system \cite{cui2022fully} that incorporates auxiliary predictions of tooth centroids and skeletons, achieving highly accurate tooth instance segmentation on a large-scale CBCT dataset.

In addition, certain research approaches have integrated CNNs with attention mechanisms or employed cross-domain analysis to enhance the performance of tooth-level tasks across different modalities. For example, Zhong et al. \cite{zhong2025automatic} proposed a UNet variant incorporating a Grouped Global Attention module and cross-layer fusion, which effectively captured texture and contour information from low-level features in PAN images, thereby enabling efficient tooth-level semantic segmentation. Long et al. \cite{long2024dftnet} introduced a Differential Feature Fusion module into the skip connections of UNet, integrating both spatial and channel attention mechanisms. This design enhanced feature extraction while alleviating information loss during the downsampling and upsampling processes, resulting in accurate tooth segmentation in CBCT images. Zhou et al. \cite{zhou2025enhancing} developed a tooth segmentation network for IOS images based on multi-scale graph convolution and cross-domain feature fusion, leveraging the integration of global and local fine-grained information to overcome variations in tooth morphology and employing attention mechanisms for cross-domain guidance. Pan et al. \cite{pan2025mdgformer} proposed a multi-domain omni-perception collaborative guidance network that decomposed high- and low-frequency features and incorporated a Cross-Attention mechanism, thus achieving deep frequency information alignment and robust edge feature learning, ultimately surpassing previous methods in IP image tooth-level segmentation tasks.

\begin{figure}[t]
	\centerline{\includegraphics[width=\linewidth]{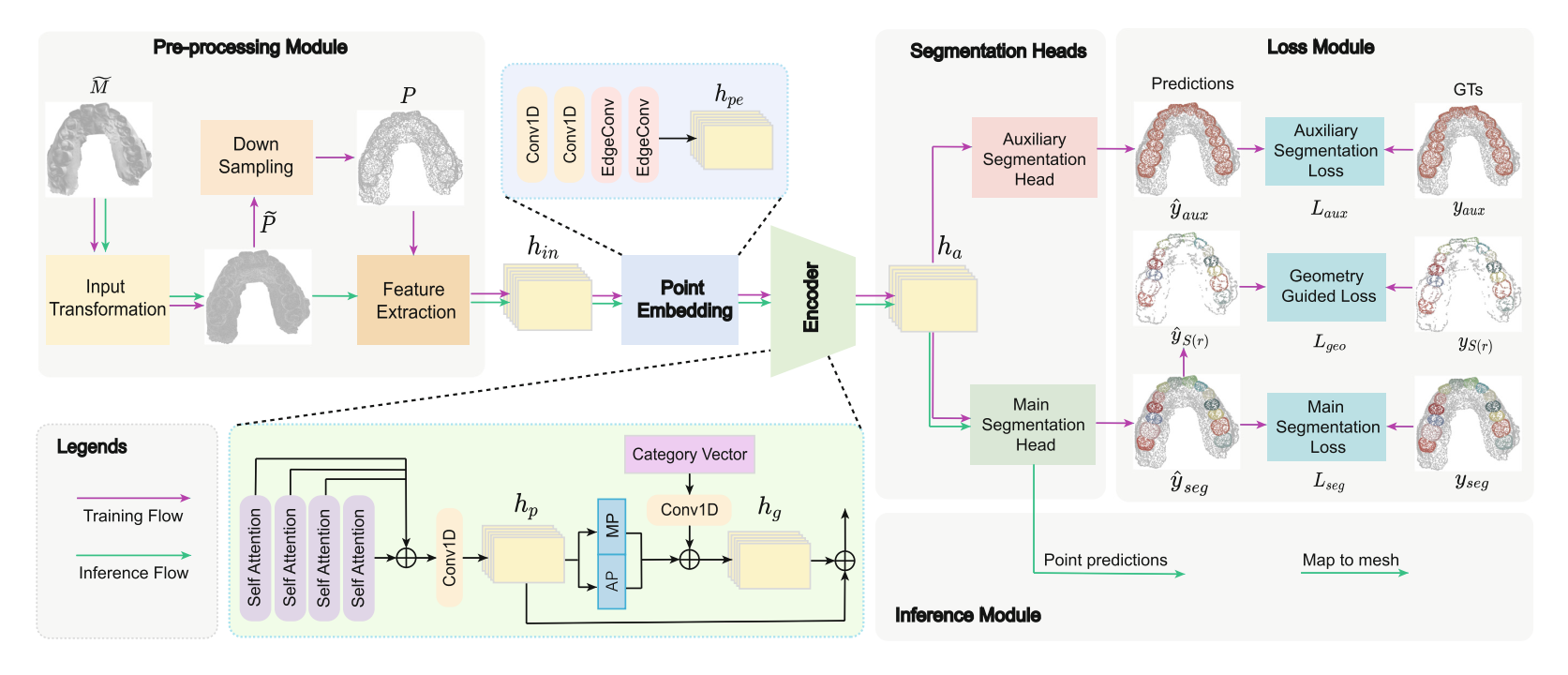}}
	\caption{The overall architecture of TsegFormer, a 3D ViT-based framework designed to address the complex boundary challenges in IOS tooth segmentation. By integrating a multi-task learning paradigm with a self-attention mechanism, TsegFormer effectively captures both local and global dependencies between teeth and gingiva, enabling more precise segmentation outcomes. Image sourced from \cite{xiong2023tsegformer}.}
	\label{fig12}
\end{figure}

\begin{figure}[t]
	\centerline{\includegraphics[width=\linewidth]{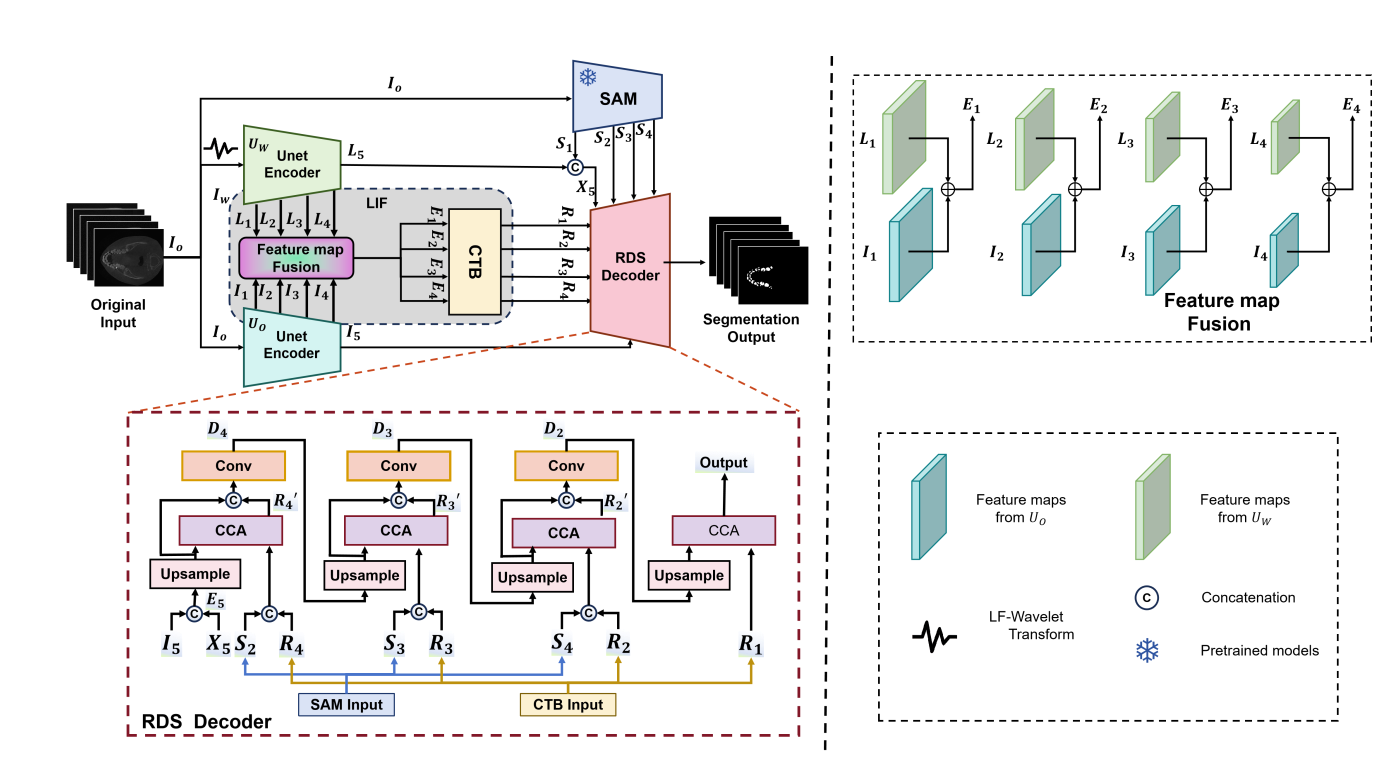}}
	\caption{The overall architecture of FDNet. In the Low-Frequency Image Fusion module, LF-Wavelet is introduced to enhance semantic representation and global structural integrity. Additionally, a large-scale ViT-based pretrained SAM encoder is employed to refine boundary features. FDNet effectively mitigates artifact interference and semantic gaps in CBCT dental images. Image sourced from \cite{feng2024fdnet}.}
	\label{fig13}
\end{figure}

As summarized in Tables \ref{tab3}, with the advent of the ViT and its growing application in medical image analysis, many researchers have focused on exploring its potential in DIA tasks. A variety of models based on both pure ViT and hybrid CNN–ViT architectures have been investigated. For pure ViT models, Xiong et al. introduced TSegFormer \cite{xiong2023tsegformer}, as depicted in Fig. \ref{fig12}, a 3D Transformer-based framework designed to address complex boundaries in tooth segmentation. By leveraging a multi-task learning paradigm combined with a self-attention mechanism, the model captures both local and global dependencies between teeth and gingiva, enabling accurate instance segmentation on IOS data. Sheng et al. \cite{sheng2023transformer} collected a dataset of 100 adult PAN images and evaluated multiple networks, demonstrating that the pure ViT-based Swin-Unet \cite{cao2022swin} outperformed other methods, thereby validating the reliability of ViT in DIA tasks. Dai et al. \cite{dai2024sparse} further proposed a pure ViT-based self-supervised framework that integrates a Masked Autoencoder with a sparse boundary prompting mechanism to achieve tooth instance segmentation on CBCT data. Compared with pure ViTs, hybrid models are more widely applied in DIA tasks. This is primarily because DIA tasks typically require models to capture global structures while simultaneously attending to local details, enabling more comprehensive and accurate feature representation. For instance, some researchers have combined CNNs and ViTs in encoder architectures to capture both rich multi-scale local details and global long-range dependencies in dental images \cite{manzari2024denunet, xi20253d, ahn2024weighted, huang2025precise}, thereby leveraging the advantages of hybrid models. Other studies have designed efficient cross-attention mechanisms tailored for dental images to achieve highly accurate tooth-level predictions \cite{wang2024trans, li2024spatial, li2024novel}. More recently, large-scale foundation models, such as SAM \cite{kirillov2023segment} and Med-SAM \cite{ma2024segment}, have been introduced into the DIA domain to harness their rich knowledge and enable efficient transfer to dental applications \cite{zhang2025easam, wang2025tooth, feng2024fdnet} (Fig. \ref{fig13}).

\begin{figure}[t]
	\centerline{\includegraphics[width=\linewidth]{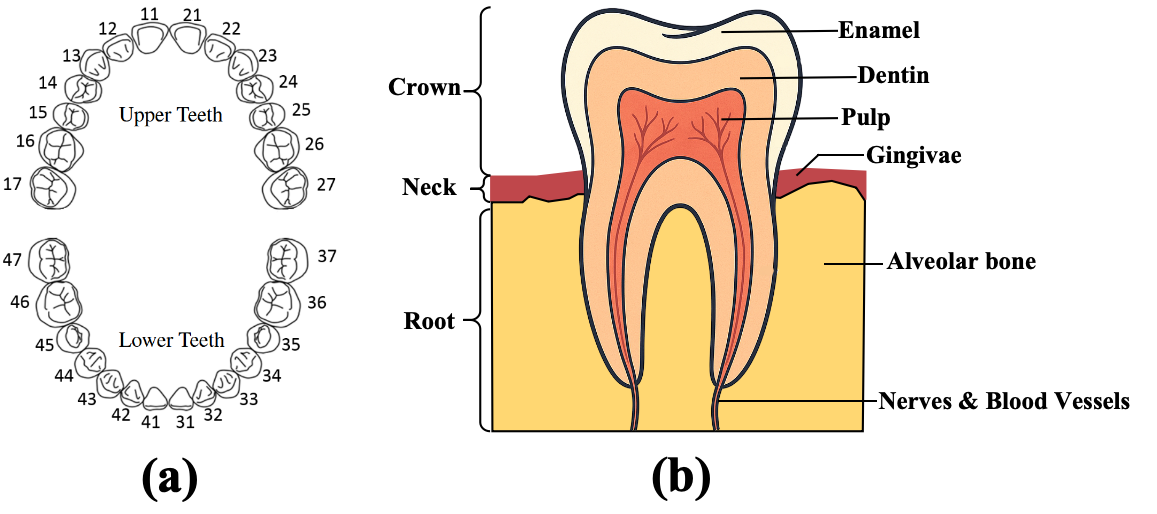}}
	\caption{Fig (a) illustrates the FDI tooth numbering system, commonly utilized for tooth-level instance segmentation and labeling, while Fig (b) depicts the anatomical structure of a tooth, serving as the target for anatomy-level prediction tasks.}
	\label{fig9}
\end{figure}

\begin{figure*}[h]
	\centerline{\includegraphics[width=\linewidth]{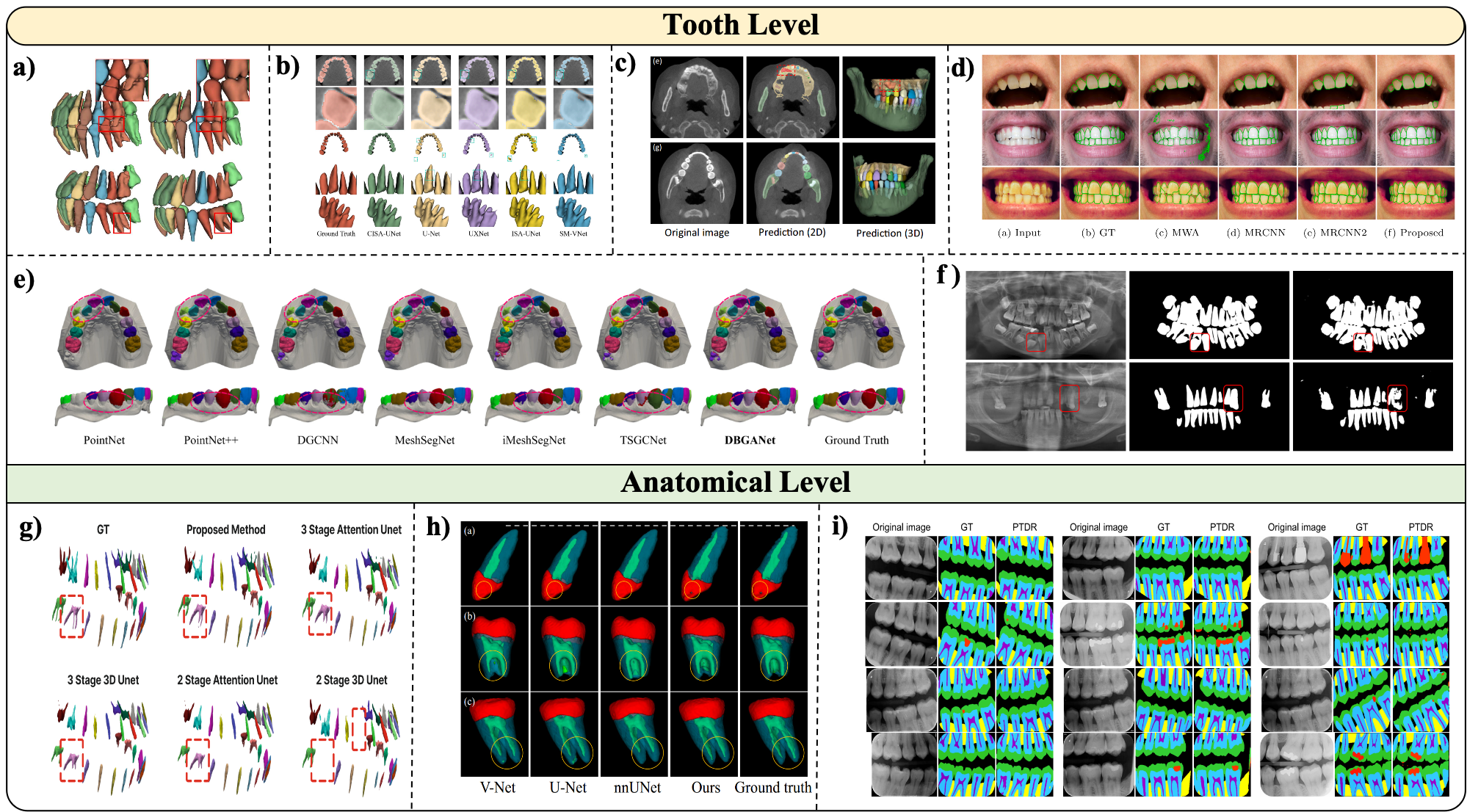}}
	\caption{Visualization of dense prediction results for dental analysis. Panels (a–f) display tooth-level dense detection results: (a) CBCT tooth instance segmentation \cite{cui2019toothnet}; (b) CBCT tooth semantic segmentation \cite{lu2025cisa}; (c) segmentation of teeth and alveolar bone in CBCT \cite{cui2022fully}; (d) tooth semantic segmentation in intraoral photographs \cite{kim2024individual}; (e) tooth instance segmentation in IOS \cite{lin2023dbganet}; and (f) tooth semantic segmentation in PAN images \cite{chen2021mslpnet}. Panels (g–i) depict anatomical-level dense detection results, as provided by \cite{kim2023automatic,tan2023dental,rousseau2023pre}, respectively.}
	\label{fig14}
\end{figure*}

\paragraph{\textbf{Anatomical Level}}
With the continuous advancement of research, the application of DL in DIA has become increasingly widespread. Tooth-level prediction has achieved remarkable results, providing insights into the arrangement of teeth and occlusal relationships. However, such predictions are limited in that they cannot capture the internal structures of teeth. Precise analysis of internal anatomical structures is critical for departments such as endodontics and serves as an essential preoperative reference for dental procedures, particularly root canal treatment. To address this limitation, researchers have extended prediction objectives from tooth-level to anatomical-level analysis, as shown in Fig. \ref{fig14} (g)-(i). As shown in Table \ref{tab4}, leveraging the powerful feature learning capabilities of DL, a variety of algorithms and models have been proposed for anatomical-level prediction. Some studies have focused on the prediction and analysis of pulp and tooth root structures. Shi et al. \cite{shi2021panoramic} investigated the performance of CNN based on LeNet-5 \cite{lecun2002gradient} in tooth root detection from PAN images, exploring different image partitioning strategies combined with edge detection operators as inputs. Li et al. developed GTUNet \cite{li2021gt}, which introduces grouped ViT into the encoder and decoder, enabling the model to preserve multi-scale information while capturing long-range dependencies for accurate root edge detection in PR. Zhang et al. proposed RCS-Net \cite{zhang2021root}, which employs a global encoder and a local decoder built upon 3D U-Net to learn fine details of the pulp region, and integrates the adaptive image enhancement algorithm CLAHE to improve robustness across various CBCT imaging conditions. Wang et al. \cite{wang2023root} adopted a two-stage framework that first uses a coarse-grained DentalNet to extract tooth-level ROIs from CBCT inputs, followed by 3D U-Net segmentation of the pulp within each ROI.

\begin{figure}[t]
	\centerline{\includegraphics[width=\linewidth]{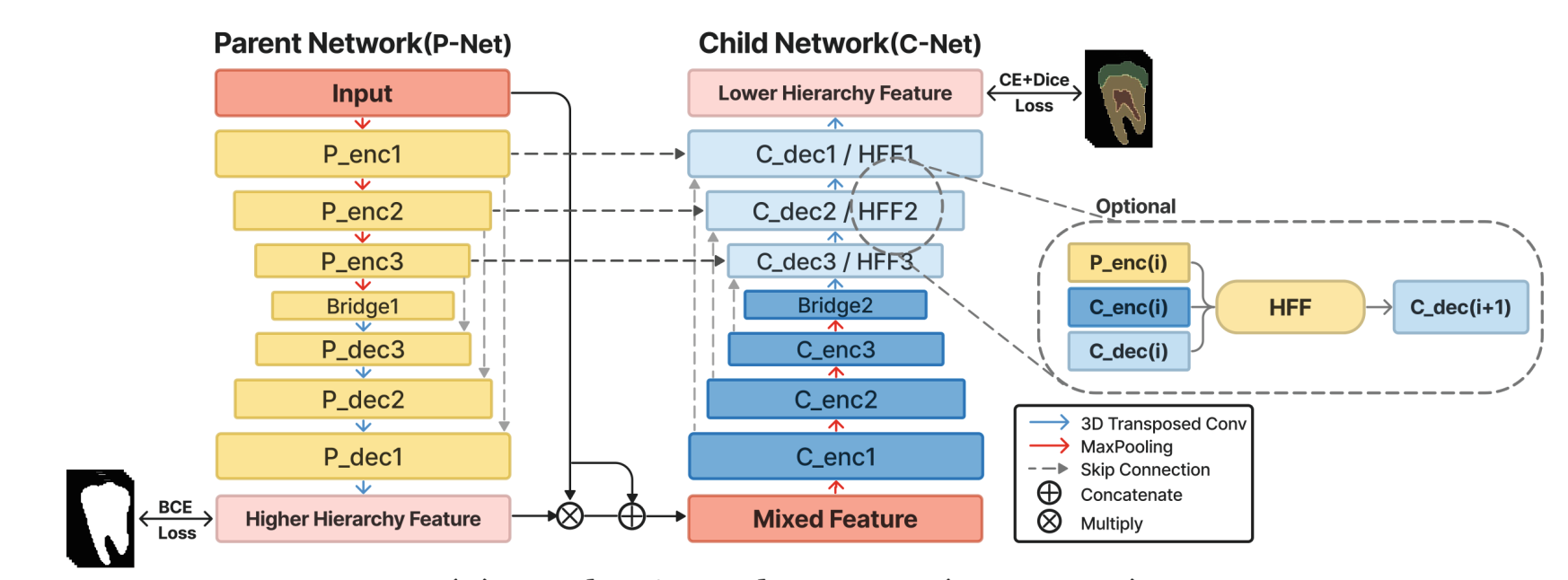}}
	\caption{DHU-Net: a three-stage CBCT tooth segmentation method comprising tooth region extraction, tooth classification and patch generation, and fine segmentation of enamel, dentin, and pulp. The HFF module fuses multi-scale features via channel–spatial attention. Image sourced from \cite{kim2023automatic}.}
	\label{fig15}
\end{figure}

Beyond the pulp and tooth root, other components such as dentin and enamel have become important targets in anatomical-level dental prediction. Rousseau et al. 

\clearpage
\begin{table*}[h]
\centering
\tiny
\setlength{\tabcolsep}{6.5pt}
\caption{A summary of tooth-level dense prediction studies based on CNN architectures, arranged in ascending order of publication year, asterisks * indicate that the study includes a detection task, symbol ※ indicate that the loss function was proposed by the authors. In the table, \ding{55} indicates that a private dataset was used, while \ding{51} indicates that a public dataset was used. S/I denotes the type of segmentation performed in the study: semantic segmentation (S) or instance segmentation (I). / indicates information not reported in the study, and in the S/I column, it indicates that the study did not include a segmentation task; for studies employing more than three evaluation metrics, we choose three most representative metrics. Tversky loss \cite{salehi2017tversky}; Smo (Smoothness) \cite{xu20183d}; AW (Adaptive Wing) \cite{wang2019adaptive}; GD (Generalized Dice) \cite{hong2019longitudinal}; CD (Chamfer Distance); SSIM (Structural Similarity); WPW-CE (Weighted Point-wise Cross Entropy); MS-SSIM (Multi-scale Structural Similarity); RC (Region Consistency); BC (Boundary Consistency); GM (GeodesicMap); DA (Detection Accuracy); FA (Identification Accuracy); ASSD (Average Symmetric Surface Distance); MSSD (Maximum Symmetric Surface Distance); DCD (Directional Cut Discrepancy); B-F1 (Boundary F1 score); RMSE (Root Mean Squared Error); HD (Hausdorff Distance).}
\label{tab2}
\renewcommand{\arraystretch}{0.6}
\begin{tabular}{%
m{1.5cm} 
c 
c 
c 
c 
c 
c 
m{5.5cm} 
}

\toprule
Paper  & Dataset & S/I & Loss & Optimizer & GPU & Evaluation &\multicolumn{1}{c}{Highlights}\\
\midrule

\cite{xu20183d} & 1200 IOS \ding{55} & I & /& / & GTX 1080 & \makecell{Mean errors:\\ 0.0848mm\\} & 3D tooth segmentation based on geometric features and boundary optimization. \\
\midrule

\makecell[l]{ToothNet \\ \cite{cui2019toothnet} }& 20 CBCT \ding{55}   & I & \makecell{CE,BCE\\SL1,MSE}& Adam & GTX 1080 Ti  & \makecell{DSC: 91.98\%\\DA: 97.75\%\\FA: 92.79\%} & Extended 3D Mask R-CNN with edge enhancement and similarity matrix, leveraging spatial relations for tooth segmentation and recognition. \\
\midrule

\cite{gou2019automatic} & 401 CBCT \ding{55}  & S & \makecell{CE} & Adam & GTX 1070 & \makecell{Accuracy: 66.7\%} & Tooth segmentation via level-set guided automatic annotation and improved U-Net. \\
\midrule

\cite{koch2019accurate} & 1500 PAN \ding{51}   & S & \makecell{CE\\BCE\\Tversky} & NAdam & / & \makecell{DSC: 93.6\%} & Patch-based U-Net with batch normalization and test-time augmentation for high-accuracy large-image tooth segmentation. \\
\midrule

\cite{majanga2019deep} & 11,148 BR \ding{55} & S & / & / & / & \makecell{Accuracy: 89\%} & Canny edge-assisted features with data augmentation enhance boundary accuracy and generalization. \\
\midrule

\cite{zanjani2019deep} &120 IOS \ding{55} & I & \makecell{BCE\\※WPW-CE} & \makecell{SGD\\Adam} & / & \makecell{IoU: 94\% \\ Precision: 93\%\\Recall: 90\%} & End-to-end 3D point cloud segmentation with PointCNN, incorporating non-uniform sampling and adversarial discrimination. \\
\midrule

\cite{rao2020symmetric} & 86 CBCT \ding{55}  & S & \makecell{BCE} & Adam & GTX 1070 & \makecell{DSC: 91.66\% \\ ASSD: 0.25mm\\MSSD: 1.18mm} & U-Net with deep bottleneck architecture and DCRF post-processing achieves high-accuracy segmentation. \\
\midrule

UDS-NET \cite{lee2020automated}& 102 CBCT \ding{55} & S & \makecell{BCE} & Adam & Titan X & \makecell{DSC: 91.8\% \\ Precision: 90.4\%\\ Recall: 93.2\%} & UDS-Net leverages multi-phase training, Dense Blocks, and Spatial Dropout to enhance feature representation and generalization. \\
\midrule

TSASNet \cite{zhao2020tsasnet} &1500 PAN \ding{51}  & S & \makecell{BCE\\SSIM} & Adam & GTX 1080 Ti & \makecell{DSC: 92.72\% \\Accuracy: 96.94\%\\Specificity: 97.81\%} & TSASNet integrates global and local attention mechanisms for precise boundary extraction in low-contrast panoramic dental images. \\
\midrule

\cite{silva2020study} * & 1,321 PAN \ding{51}  & I & / & SGD & RTX Titan& \makecell{PANet: \\Accuracy: 96.7±0.1\%\\Specificity: 98.7±0.2\%} & Two-stage instance segmentation networks enable accurate tooth detection, segmentation, and numbering in panoramic X-rays with blurred boundaries. \\
\midrule

CRIS  \cite{van2020object}& 170600 IP \ding{55}  & I & / & / & / & \makecell{F-measure: 81.9\%} & CRIS refines individual tooth contours by integrating instance masks with RCF edge features. \\
\midrule

\cite{sivagami2020unet} & 1171 PAN \ding{51}  & S & \makecell{BCE} & Adam &/ & \makecell{DSC: 94\% \\Specificity: 95\%\\F1-Score: 93\%} & UNet-based end-to-end FCN efficiently extracts features from PAN using encoder-decoder structure. \\
\midrule

\cite{zhang2020automatic} & 120 IOS \ding{55} & I & \makecell{Weighted-\\BCE} & /& GTX 1080 Ti & \makecell{DCD: 0.046mm\\Accuracy: 98.87\%} & Harmonic mapping with improved FCC algorithm efficiently transforms 3D tooth meshes for 2D processing and boundary refinement. \\
\midrule

\cite{sun2020tooth} & 100 IOS \ding{55} & I & \makecell{CE\\Smo\\※RC,※BC} & / & / & \makecell{Accuracy: 97\%} & FeaStNet-based graph convolutional network performs multi-label tooth segmentation by integrating comprehensive vertex features with explicit anatomical priors to ensure precise vertex-wise correspondence. \\
\midrule

c-SCN \cite{sun2020automatic}* & 100 IOS \ding{55}  & I & \makecell{CE,CD\\Smo\\※GM} & Adam & GTX 1080 Ti & \makecell{DSC(Upper): \\Molar: 98\%\\Premolar: 98\%\\Canine: 98\%} & Use FeaStNet to produce geodesic maps, multi-label classification, and dense correspondence for vertex attribute transfer. \\
\midrule

\cite{yang2021accurate} & 5244 CBCT \ding{55}& S & \makecell{CE} & Adam & / & \makecell{DSC: 97.91\% \\ DA: 97.33\%\\B-F1: 98.24\%} & Combining deep CNN with ellipse-shaped priors in a level set framework enables precise tooth boundary refinement. \\
\midrule

MSLPNet  \cite{chen2021mslpnet}& 1500 PAN \ding{51} & S & \makecell{Dice\\BCE\\MS-SSIM} & Adam & GTX 1080 Ti & \makecell{DSC: 93.01\% \\ Accuracy: 97.30\%\\  Specificity: 98.45\%} & Employ multi-scale location perception modules to enhance fuzzy root boundary and global feature representation. \\
\midrule

ToothPix \cite{cui2021toothpix}& 1500 PAN \ding{55}& S & \makecell{L1\\BCE} & Adam & RTX 2070 & \makecell{IoU: 90.42\% \\Specificity: 98.52\%\\F1-score: 94.86\%} & Use a GAN with encoder-decoder residual generator and FCN discriminator to enhance mask realism. \\
\midrule

\cite{harsh2021attention} & 1500 PAN \ding{51}& S & \makecell{Dice} & Adam & RTX 2060 & \makecell{DSC: 94.7\% \\Accuracy: 92.8\%\\F1-score: 88.09\%} & Attention gates in U-Net skip connections retain critical features while suppressing redundant information. \\
\midrule

\cite{zhao20213d} & 80 IOS \ding{55} & I & \makecell{CE} & Adam & GTX 1080 & \makecell{Accuracy: 94.84\% \\mIoU: 87.09\%} & End-to-end GNN integrates local augmentation, graph attention, and global features for multi-scale 3D tooth mesh representation. \\
\midrule

DLLNet \cite{lang2021dllnet}* & 77 IOS \ding{55}  & / & \makecell{GD\\AW\\MSE} & Adam & GTX 1080 Ti & RMSE: 0.386mm & Design two branches: landmark localization and landmark area segmentation, and integrate the attention mechanism to achieve detection. \\
\midrule

TSegNet \cite{cui2021tsegnet}& 2000 IOS \ding{55}& I & \makecell{L1\\CD\\CE,BCE} & Adam & GTX 1080 Ti & \makecell{DSC(point): 98.0\% \\DSC(surface): 98.6\%\\F1-score: 94.2\%} & Two-stage framework with centroid regression and confidence-aware attention enables instance-level tooth feature extraction and boundary refinement. \\
\midrule

\cite{jang2021fully} * & \makecell{194 PAN\ding{55} \\97 CBCT\ding{55}} & I & / & Adam& RTX 2070 & \makecell{DSC: 94.79\%\\HD: 1.66mm\\ASSD: 0.14 mm} & Four-step CBCT method using multi-threshold Otsu and U-Net enables 3D tooth detection, numbering, and ROI extraction. \\
\midrule

TSGCNet \cite{zhang2021tsgcnet}& 80 IOS \ding{55} & S & \makecell{CE} & Adam & GTX 1080 & \makecell{mIoU: 88.99\%\\Accuracy: 95.25\%} & TSGCNet integrates coordinate topology and normal boundaries via dual-stream attention for discriminative geometry. \\
\midrule

TSGCN  \cite{zhao2021two}& 80 IOS \ding{55}  & I & / & Adam & GTX 1080 & \makecell{mIoU: 91.69\% \\ Accuracy: 96.96\%} & An expanded version of the above work. C/N streams employ Input-Transformer and attention fusion to improve geometric alignment and robustness.  \\
\midrule

&  &  &  &  &   &  \multicolumn{2}{c}{\textbf{\small(continued on next page)}}\\
\end{tabular}
\renewcommand{\arraystretch}{1}
\end{table*} 
\clearpage
\clearpage
\begin{table*}[h]
\centering
\tiny
\captionsetup{list=no} 
\caption*{Table \ref{tab2} (Continued from last page.) Huber \cite{ren2015faster}; SSE (Sum of Squared Errors); ASMS (Automatically Searched Metric-Sensitive); IV (Intracluster Variance); ID (Intracluster Distance); ML-InfoNCE (Multi-level InfoNCE); PE (Potential Energy); KC (Knowledge Consistency); KL-Divergence (Kullback–Leibler Divergence); CBC (Contrastive Boundary-Constrained); SL1 (Smooth L1); ASD (Average Surface Distance); PPV (Positive Predictive Value); HD95 (95th Percentile Hausdorff Distance); VOE (Volumetric Overlap Error).} 
\setlength{\tabcolsep}{3.5pt}
\renewcommand{\arraystretch}{0.9}
\begin{tabular}{%
m{1.5cm} 
c 
c 
c 
c 
c 
c 
m{5.5cm} 
}

\toprule
Paper  & Dataset & S/I & Loss & Optimizer & GPU & Evaluation &\multicolumn{1}{c}{Highlights}\\
\midrule

LTPEDN \cite{salih2022local}& 1500 PAN \ding{55} & S & \makecell{Dice} & / & / & \makecell{DSC: 92.42\% \\Accuracy: 94.32\%} & LTPEDN achieves efficient DIA via multi-stage feature learning and edge enhancement while reducing computational complexity. \\
\midrule

BDU-net  \cite{zhang2022bdu}& 916 PAN \ding{55} & I & \makecell{BCE} & SGD & RTX 2080 Ti & \makecell{mDSC: 92\%} & BDU-net enhances dental recognition and segmentation by multi-scale DFE fusion and dual boundary-region subnetworks. \\
\midrule

\cite{li2022semantic} & 350 CBCT \ding{55}  & I & \makecell{CE\\Dice} & Adam &  Tesla V100 & \makecell{DSC: 91.13±0.45\% \\Precision: 92.13±0.29\%\\HD: 1.00±0.27mm} &GCN-based attention mechanism with global context improves spatial modeling and boundary recognition. \\
\midrule

LeFUNet \cite{rao2022lefunet}*& 1616 PAN \ding{51}  & S & / & Adam & RTX 2080 & \makecell{DSC: 96.94±0.74\% \\Precision: 96.81±0.23\%\\Specificity: 99.61±.83\%} & LeFUNet leverages multi-stage preprocessing and SE modules to enhance feature reuse for panoramic tooth detection. \\
\midrule

\cite{cui2022fully} & 4938 CBCT \ding{55}  & I & \makecell{BCE\\ L2 \\ CE} & Adam & Tesla V100 & \makecell{DSC: 94.1\% \\ Sensitivity: 93.9\%\\ASD: 0.17mm} & Three-stage V-Net framework: ROI localization, centroid and skeleton-guided instance refinement, and Haar-filtered cascaded V-Net for alveolar bone. \\
\midrule

\cite{chandrashekar2022collaborative} *  & 1500 PAN \ding{51}  & I & \makecell{CE\\BCE,SL1\\IOU,SSE} & SGD & / & \makecell{Accuracy: 98.44\% \\ F1-Score: 98.75\% \\ mAP: 97.78\%} & A collaborative deep learning–based tooth detection and segmentation model, enabling precise segmentation and recognition on PAN. \\
\midrule

\cite{xie2022automatic} & 78 CBCT \ding{55}  & I & \makecell{CE} & / & RTX 2080 Ti & \makecell{DSC: 88±3\% \\ Precision: 98±3\%\\ASSD: 0.53±0.34mm} & Multi-task U-Net integrates CSA modules to enhance multi-scale features for pixel-wise foreground and landmark prediction. \\
\midrule

\cite{aseretto2022contrast} & 1500 PAN \ding{51}  & S & \makecell{CE} & Adam & GTX 1070 & \makecell{F1-score: 91\% \\ Precision: 92\%} & U-Net-based FCN uses a ResNet18 encoder with transfer learning to improve performance. \\
\midrule

\cite{rundo2022hierarchical} & 52 CBCT \ding{55} & S & \makecell{Dice} & Adam & RTX 3090 & \makecell{DSC: 66.2\% \\ Specificity: 89.9\%\\F1-score: 65.6\%} & A 3D convolutional encoder–decoder with deep supervision and skip connections enables efficient dental arch segmentation. \\
\midrule

\cite{shibuya2022semantic} & 119 PAN \ding{55} & S & \makecell{CE} & Adam & / & \makecell{mIoU: 67.1\%} & Based on DeepLab v3+, combining ASPP and decoder for segmentation, with Alpha Blending enhancing training diversity. \\
\midrule

IGNet  \cite{liu2022heterogeneous} & 210 IOS \ding{55}  & S & \makecell{CE\\※ASMS} & SGD & Tesla V100 & \makecell{mIoU: 81.25\% \\mAcc: 96.56\%} & IGNet fuses heterogeneous coordinate and normal graphs via cross-attention and uses metric-driven automated loss search. \\
\midrule

\cite{caylak2022automated} & 131 PAN \ding{55}  & S & / & / & MX150  & \makecell{DSC(best): 90.26\%} & Exploring the performance of different UNet architectures in DIA after transfer learning with various pretrained weights. \\
\midrule

\cite{tian20223d} * & 210 IOS \ding{55} & I & \makecell{CE\\L2\\IV,ID} & / & Titan X & \makecell{mAP50\%: 55.0\% \\mAP25\%: 72.4\%} & A 3D U-Net backbone with SSC jointly performs point-wise semantic segmentation and centroid prediction for improved tooth modeling. \\
\midrule

\cite{boudraa2022segmentation} & 85 IOS \ding{55}  & I & / & / & Tesla K80& \makecell{DSC: 98.04\% \\ Sensitivity: 97.99\%\\PPV: 98.19\%} & An improved MeshSegNet employs mesh-based multi-scale feature learning with 15D inputs for efficient 3D tooth segmentation. \\
\midrule

\cite{wu2022two} * & 136 IOS \ding{55} & I & \makecell{Dice\\MSE} & / & RTX 2080 Ti & \makecell{Sensitivity: 94.6±5.4\%\\DSC: 94.3±4.4\%\\HD: 2.420±3.140 mm} & A two-stage approach that leverages segmentation as the foundation to automatically identify anatomical landmarks in IOS. \\
\midrule

DArch  \cite{qiu2022darch}& 4773 IOS \ding{55}  & I & \makecell{CE\\Huber} & / & RTX 3090 & \makecell{DSC: 97.7\% \\ Accuracy: 99.68\%\\Recall: 85.39\%} & A coarse-to-fine dental arch prediction using Bézier curves and GCN is proposed, with APS sampling improving detection accuracy. \\
\midrule

\cite{li2022multi} & 180 IOS \ding{55} & I & \makecell{L2\\Dice} & / & / & \makecell{DSC: 95.68\% \\ Sensitivity: 95.73\%\\PPV: 95.83\%} & A BEM block is proposed to resolve local context ambiguity by offset learning, integrating multi-scale contexts for accurate tooth segmentation. \\
\midrule

STSNet  \cite{liu2022hierarchical}& 13000 IOS \ding{55}  & I & \makecell{CE\\※ML-InfoNCE} & SGD & / & \makecell{DSC: 94.85\% \\mIoU: 93.12\% \\Accuracy: 97.22\%} & STSNet leverages self-supervised contrastive pretraining and fine-tuning with limited labeled data to improve 3D tooth segmentation. \\
\midrule

\cite{cao2023stagewise} & 148 CBCT \ding{55}  & I & \makecell{CE\\※PE\\Dice} & Adam & RTX 3090 & \makecell{DSC: 96.89\% \\ HD95: 0.54mm\\ASD: 0.11mm} & Three-stage instance segmentation with spectral filtering and dilated convolutions improves tooth boundary accuracy. \\
\midrule

\cite{jana2023automatic} & 50 IOS \ding{55}  & I & / & / & RTX 8000 & \makecell{DSC(1st.): 92.88\% \\Accuracy(1st.): 94.12\%\\Sensitivity(1st.): 93.89\%} & The study evaluates general and dental-specific point cloud segmentation methods on mesh datasets. \\
\midrule

KCNet  \cite{lin2023lightweight} & 1321 PAN \ding{51}  & S & \makecell{CE\\KC\\KL-Divergence} & / & / & \makecell{DSC: 89±3.8\% \\HD: 5.617±0.756mm\\VOE: 0.22±0.153mm} & Proposed lightweight KCNet uses knowledge distillation to maintain high accuracy while reducing computational cost. \\
\midrule

\cite{gan2023concrete} & 750 CBCT \ding{55}  & S & \makecell{CE\\Dice} & / & / & / & Combining U-Net segmentation with Noise2Sim denoising enables region-adaptive image enhancement. \\
\midrule

\cite{tan2023boundary} & 103 IOS \ding{55} & I & \makecell{BCE\\CBC\\DICE} & / & / & \makecell{mIoU: 83.77\%\\Accuracy: 90.98\%} & Graph neural network with boundary-constrained loss improves tooth and gingiva segmentation accuracy. \\
\midrule

BFFNet \cite{zhang2023boundary} & 106 PAN \ding{51}  & S & \makecell{IOU\\BCE} & Adam & \makecell{RTX 2080 Super} & \makecell{DSC: 79.11\% \\HD: 0.0174mm} & BFFNet enhances tooth segmentation by fusing boundary features with global contextual information. \\
\midrule

WMC-Net  \cite{li2023semi}& 312 CBCT \ding{51}  & S & \makecell{CE\\MSE\\Dice} & SGD &RTX 3090 Ti & \makecell{DSC: 76.98±5.59\%} & The weakly-consistent network integrates CBAM and EG-MT to improve tooth boundary recognition. \\
\midrule

\cite{leclercq2023dentalmodelseg} & 78 IOS \ding{55} & I & \makecell{CE\\Dice} & Adam & Titan V & \makecell{DSC: 96±3\% \\Sensitivity: 98±11\%\\PPV: 98±3\%} & PyTorch3D-based 2D rendering generates multi-view RGB normal maps and ground-truth labels. \\
\midrule

\cite{feng2023automated} & 312 CBCT \ding{51}  & S & \makecell{BCE\\Dice} & / & \makecell{RTX 4090\\RTX 4070 Ti} & \makecell{Dice: 77.5\% \\ HD: 0.269mm} & A 3D U-Net with group normalization was employed to optimize small-batch training. \\
\midrule

&  &  &  &  &   &  \multicolumn{2}{c}{\textbf{\small(continued on next page)}}\\
\end{tabular}
\renewcommand{\arraystretch}{1}
\end{table*} 
\clearpage
\clearpage
\begin{table*}[h]
\centering
\tiny
\captionsetup{list=no} 
\caption*{Table \ref{tab2} (Continued from last page). FPR (False Positive Rate); NSD (Normalized Surface Dice); Lbl-T (Teeth Labeling Accuracy)} 
\setlength{\tabcolsep}{3.5pt}
\renewcommand{\arraystretch}{0.8}
\begin{tabular}{%
m{1.5cm} 
c 
c 
c 
c 
c 
c 
m{5.5cm} 
}

\toprule
Paper  & Dataset & S/I & Loss & Optimizer & GPU & Evaluation &\multicolumn{1}{c}{Highlights}\\
\midrule

\cite{zhang2023high} & 3000 CBCT \ding{51}  & S & \makecell{CE\\Dice\\KL-Divergence} & SGD & RTX 3090 & \makecell{DSC: 72.61\% \\HD: 0.2595mm} & Morphology-driven preprocessing combined with dynamic convolution enables multi-head deep supervision. \\
\midrule

\cite{wang2023multi} & 362 CBCT \ding{51} & S & / & AdamW & Tesla V100 & \makecell{DSC: 80.58\% \\ HD: 0.1599mm} & A semi-supervised domain-adaptive framework with FTA and adaptive thresholds enhances performance. \\
\midrule

\cite{zhao2023recognition} & 280 PAN \ding{55} & I & \makecell{CE\\BCE\\SL1} & / & Tesla K80 & \makecell{mPrecision: 86.32\% \\ mRecall: 90.05\%\\ mF1-score: 87.30\%} & Implemented tooth instance segmentation using Mask R-CNN on a self-constructed dataset. \\
\midrule

Teeth U-Net \cite{hou2023teeth} & 1500 PAN \ding{55}  & S & \makecell{CE} & RMSprop & Titan V & \makecell{DSC: 94.28\% \\Accuracy: 98.53\%\\Precision: 95.62\%} & The model integrates multiscale and self-attention modules with dense skip connections for encoder-decoder tooth feature fusion. \\
\midrule

\cite{jana2023critical} & 589 IOS \ding{51}  & I & / & / & RTX 8000 & \makecell{DSC(Whole Jaw):\\1st.:91.05\%} & Conducted a comparative analysis of several point cloud segmentation methods on a self-constructed dataset. \\
\midrule

\cite{duan20233d} & 200 IOS \ding{55}  & I & \makecell{CE} & / &RTX 3090 & \makecell{mIoU: 92.1\% \\ Accuracy: 96.97\%} & Decouple coordinate and normal features, leveraging semantic-guided graph transformation and cross-domain attention for segmentation. \\
\midrule

DBGANet  \cite{lin2023dbganet}& 637 IOS \ding{51} & I & \makecell{CE\\BCE\\SL1} & Adam &  RTX 3090 & \makecell{Teeth3DS:\\ DSC: 94.48±0.83\% \\ Sensitivity: 96.71±0.66\%} & DBGANet employs voxel and surface branches with geometric attention for boundary-enhanced bi-directional feature fusion in 3D segmentation. \\
\midrule

MRCM-UCTransNet \cite{wen2024mrcm} & 30 CBCT \ding{55}  & S &\makecell{Focal} & /& Tesla V100 & \makecell{DSC: 90.612\% \\ mIoU: 82.836\%\\Sensitivity: 93.855\%} & Enhance the U-Net backbone by integrating multi-scale residual CNN and a CTrans block for effective feature extraction and channel-wise attention fusion. \\
\midrule

\cite{beser2024yolo} * & 3854 PAN \ding{55} & I & IOU & SGD &  / & \makecell{Precision: 99 \% \\Recall: 99\%\\mAP: 98\%} & Based on YOLOv5, a deep learning model was proposed for tooth detection in children during the mixed dentition period. \\
\midrule

\cite{zannah2024semantic} & 389 PAN  \ding{55} & S & \makecell{Dice} & Adam & RTX 3080 Ti & \makecell{DSC(L3): \\1st.: 90.35\%\\2nd.: 90.33\%} & Compared U-Net and its variants (Attention, Residual, U-Net++) to assess module contributions to dental segmentation performance. \\
\midrule

USCT  \cite{jing2024usct}& \makecell{66 CBCT\ding{51} \\ 19 CBCT \ding{55}}  & S & \makecell{MSE\\Dice} & SGD & Titan XP & \makecell{DSC: 90.76\% \\HD95: 6.29mm \\ASD: 1.76mm} & USCT integrates symmetric consistency learning with uncertainty regularization, employing ACIM and cross-task constraints for robust semi-supervised segmentation. \\
\midrule

CRML-Net \cite{lyu2024crml} & 129 CBCT \ding{51}  & S & \makecell{Focal\\BCE, Dice\\Huber} & Adam & RTX 4090 & \makecell{DSC: 92.43\% \\ HD: 1.21mm\\ASD: 0.31mm} & CRML-Net skeleton fused with CBCT for features, multi-task branches jointly predict tooth masks and boundaries. \\
\midrule

\cite{lv2024tooth} & 20 CBCT \ding{55} & I & \makecell{L1\\BCE\\Dice} & / & / & \makecell{DSC: 83.26\% \\ Precision: 93.82\%\\HD: 2.44mm} & Three-stage network: ROI detection, skeleton extraction, and skeleton-guided V-Net for precise single-tooth segmentation with multi-scale fusion. \\
\midrule

\cite{brahmi2024automatic} * & 180 PAN \ding{55}  & I & / & SGD & Tesla T4 & \makecell{mAP: 90\% \\ Precision: 96\%\\F1-score: 63\%} & Mask R-CNN framework with ResNet101-FPN backbone for simultaneous tooth instance segmentation and identification. \\
\midrule

\cite{kim2024individual} & 46 IP \ding{51} & I & \makecell{BCE} & Adam & / & \makecell{mF1-score: \\T01:96.49\%\\T02: 97.30\%} & ResNeSt-50 generates pseudo-edges; dual-level set active contours with tooth enhancement map refine accurate tooth segmentation. \\
\midrule

DFTNet \cite{long2024dftnet} & / CBCT \ding{51} & I & \makecell{CE\\Dice} & Adam & / & \makecell{DSC: 97.15\% \\Recall: 96.95\%\\Precision: 97.36\%} & V-Net with DFF performs voxel segmentation; Mask R-CNN detects instances; fused results yield precise tooth instance masks. \\
\midrule

DAE-Net \cite{chen2024dae} * & 600 PAN \ding{51}  & I & \makecell{IOU\\Dice} & Adam & RTX 4060 Ti & \makecell{IoU: 27.0\% \\ NSD: 59.5\%\\IA: 26.7\%} & DAE-Net uses FPN and RPN with RFEM and CBAM to generate tooth instance masks and refine segmentation. \\
\midrule

\cite{ji2024two} & 2400 PAN \ding{51} & I & / & / & RTX 4090 & \makecell{DSC(instance): 79.82\% \\ NSD(instance): 84.14\%\\mIoU(instance): 75.66\%} & A two-stage nnU-Net framework with semi-supervised self-training performs quadrant-wise fine tooth segmentation numbering. \\
\midrule

\cite{isensee2024scaling} & / CBCT \ding{51}  & I & / & / & \makecell{GH200\\Tesla A100}  & \makecell{DSC: 92.53\% \\ HD95: 18.472mm} & Deepened nnU-Net ResEnc L with seven residual encoder stages, optimized training and post-processing for accurate tooth segmentation. \\
\midrule

UNet 3+ \cite{csahin2024automated}& 2793 PAN \ding{51}  & S & \makecell{Dice} & / & RTX 3090 Ti & \makecell{DSC: 93.31\% \\ Accuracy: 97.36\%\\Precision: 93.25\%} & U-Net 3+ adds multi-scale skip connections and feature fusion to improve tooth structure segmentation. \\
\midrule

\cite{qin2024inter} & \makecell{175 CBCT\ding{55} \\ 148 CBCT \ding{51}} & S & \makecell{Dice} & / & / & \makecell{DSC: \\Pub: 92\%\\Cli: 92.05\%} & 3D U-Net adds inter-layer attention, residual blocks, and attention gates to enhance boundary segmentation and multi-scale feature fusion. \\
\midrule

CAMS  \cite{osama2024cue} & 693 PAN \ding{51}  & S & / & / & / & \makecell{mDSC: 90.41\% \\ mIoU: 83.18\%\\F1-score: 90.41\%} & Three U-Net models: Specialized per quadrant, All-In-One multi-class, CAMS uses cue mask for conditional segmentation. \\
\midrule

\cite{chen2024novel} & 350 CBCT \ding{51}  & I & \makecell{CE\\Dice} & AdamW & Tesla V100 & \makecell{$\text{DSC}_{A}$: 87.7\% \\$\text{NSD}_{A}$: 88.7\%} & The method employs a two-stage semi-supervised framework: ROI extraction via 3D-VNet followed by mask-assisted instance segmentation. \\
\midrule

RAGCNet \cite{zhao2024few} & 180 IOS \ding{51}  & I & \makecell{L1\\CE} & Adam & RTX 3090 & \makecell{Accuracy: 95.30±0.11 \%\\mIoU: 89.20±0.25\%\\mAccuracy: 90.02±0.17\%} & RAGCNet leverages a Group module and a Region-Aware Module for feature learning, achieving accurate tooth segmentation through dual-branch feature fusion and a centroid-based loss function. \\
\midrule

\makecell[l]{THISNet \\ \cite{li2023thisnet} }& 1200 IOS \ding{51} & I & \makecell{CE\\L2\\Dice\\Weighted Focal} & Adam & Tesla V100& \makecell{Accuracy: 94.69±0.53\%\\mIoU: 87.05±1.68\%} & THISNet leverages region highlighting and affinity modules for tooth instance segmentation, with multi-task learning improving classification and segmentation. \\
\midrule

CurSegNet \cite{mao2024cursegnet} & 30 IOS \ding{55}  & I & \makecell{CE} & / & GTX 2080 & \makecell{DSC: 95.73\% \\Sensitivity: 95.20\%\\PPV: 96.39\%} & Dual-branch CurSegNet integrates GAM, CurveNet aggregation, and cross-layer fusion for 3D tooth mesh segmentation. \\
\midrule

WS-TIS  \cite{wang2024weakly} & 1200 IOS \ding{51}  & I & \makecell{CE\\Focal} & / & RTX 4090 & \makecell{DSC: 97.37±0.48\% \\mAP: 96.37±0.74\%\\FPR: 17.50±1.19\%} & Disentangled re-sampling and gated-attention highlight classes, Grad-CAM localizes teeth, DGCNN models fine contours. \\
\midrule

&  &  &  &  &   &  \multicolumn{2}{c}{\textbf{\small(continued on next page)}}\\
\end{tabular}
\renewcommand{\arraystretch}{1}
\end{table*} 
\clearpage
\begin{table*}[h]
\centering
\tiny
\captionsetup{list=no} 
\caption*{Table \ref{tab2} (Continued from last page). MCC (Matthew’s Correlation Coefficient); MAE (Mean Absolute Error).} 
\setlength{\tabcolsep}{3.5pt}
\renewcommand{\arraystretch}{0.9}
\begin{tabular}{%
m{1.5cm} 
c 
c 
c 
c 
c 
c 
m{5.5cm} 
}

\toprule
Paper  & Dataset & S/I & Loss & Optimizer & GPU & Evaluation &\multicolumn{1}{c}{Highlights}\\
\midrule

\makecell[l]{HiCA \\ \cite{li2024novel} }& \makecell{1200 IOS \ding{51}\\76 IOS \ding{55} } & I & \makecell{CE} & Adam &  GTX 3090 & \makecell{Teeth3DS: \\DSC: 93.44±8.60\% \\ Accuracy: 94.92±5.47\%} & HiCA integrates cross-stream contextual and discriminative modules for multistream feature fusion and robust missing-tooth segmentation. \\
\midrule

\makecell[l]{TSG-GCN \\ \cite{liu2024individual}} & 710 CBCT \ding{55}  & I & \makecell{CE\\MSE\\Dice} & Adam & RTX 3090 & \makecell{DSC: \\Child: 94.50±4.16\% \\ Adult: 94.10±1.64\%} & TSG-GCN integrates 2D projection and 3D GCN, dynamically modeling teeth spatial relationships.
\\
\midrule

\cite{DBLP:journals/tmm/ZhuangWCZ25} & 1200 IOS \ding{51} & I & \makecell{CE\\CD\\SL1,$L_{sep}$\\※Diff}& / &/ & \makecell{DSC: 97.63\%\\Lbl-T: 95.62\%} & Robust tooth segmentation by fusing semantic segmentation and instance segmentation , followed by a global optimization and post-processing step for accurate final labeling. \\
\midrule

CMAT \cite{chen2025cross} & 5138 IOS \ding{55} & I & \makecell{CE\\Entropy\\KL-Divergence} & AdamW & RTX 3090 & \makecell{mIoU: \\CrossTooth : 82.19\%\\AbnTooth: 87.32\%} & Source prototype alignment with progressive pseudo-labels and tooth-ratio priors enables cross-domain tooth segmentation adaptation. \\
\midrule

\cite{arian2025unsupervised} & 40 IOS \ding{55}  & S & / & / & / & \makecell{mIoU: 95\%} & Unsupervised 3D dental arch segmentation with PointNet++, using GRL and Siamese for domain adaptation. \\
\midrule

\cite{ghafoor2025enhancing} & 540 PAN \ding{55}  & I & \makecell{Dice} & / & RTX 3090 & \makecell{DSC: 91.72\% \\ Precision: 80.44\%\\Specificity: 97.31\%} & U-Net with global and teeth attention enhances boundary and salient feature representation. \\
\midrule

\cite{yaswanth2025automated} & 1000 PAN \ding{51} & S & \makecell{BCE} & Adam & / & \makecell{DSC: 90.7\% \\Accuracy: 97.8\%\\mAP: 96.4\%} & U-Net++ with ECA attention enhances multi-scale feature fusion and key channel representation. \\
\midrule

MSCA-UNet \cite{pan2025multi} & 2500 PAN \ding{51}  & S & \makecell{CE\\Dice} & Adam & RTX 4090 & \makecell{DSC: 92.37±0.01\%} & U-Net integrates adaptive convolution and multi-scale learning for global context and high-resolution detail. \\
\midrule

CISA-UNet \cite{lu2025cisa} & 27474 CBCT \ding{51}  & S & \makecell{DICE} & Adam & RTX 3090 Ti & \makecell{DSC: 94.10±1.93\% \\ MCC: 94.13±1.90\%\\Accuracy: 99.94±0.020\%} & CISA-UNet, built upon the 3D UNet backbone, integrates edge maps and CLAHE-enhanced images. Incorporating the CISA module for refined feature extraction, it achieves high-precision tooth segmentation from CBCT scans. \\
\midrule

SMTLNet \cite{zhao2025smtlnet} & \makecell{1305 CBCT \ding{55}\\98 CBCT \ding{51}}  & I & \makecell{BCE\\Dice\\※Intraclass} & Adam & \makecell{ RTX Titan\\RTX 3090 Ti} & \makecell{DSC: 91.84±1.67\% \\Precision: 94.47±0.02\%\\HD: 1.41±0.82mm} & SMTLNet leverages self-supervised pretraining to learn 3D features, followed by coarse and fine segmentation to refine boundaries and overall results. \\
\midrule

\cite{ma2025high} & 1020 PAN  \ding{55}  & I & \makecell{Dice} &/& RTX 3090& \makecell{DSC: 92.0\% \\Precision: 92.3\%\\Recall: 92.5\%} & The method consists of tooth and alveolar bone segmentation, both based on nnU-Net. Optimization combines pixel aggregation and CVAE to improve completeness and anatomical plausibility. \\
\midrule

\cite{zhou2025enhancing} & 280 IOS \ding{51}  & I & \makecell{CE\\GD} & Adam & RTX 4090 & \makecell{Accuracy: 95.68±0.12\% \\mIoU: 90.32±0.41\%\\mAccuracy: 91.41±0.59\%} & The network extracts tooth features via coordinate and normal vector branches and fuses them to produce final semantic labels. \\
\midrule

GCNet \cite{zhong2025automatic} & 3187 PAN \ding{51} & S & \makecell{BCE\\IOU} & Adam & RTX 4090 & \makecell{DSC: 93.38\%\\Recall: 94.26\%\\MAE: 0.0259mm } & Enhanced UNet, GGA+CLF fuses multi-level features, DOD decodes edges and segmentation \\
\midrule

PXseg \cite{wang2025pxseg} & \makecell{420 PAN \ding{55}\\258 CBCT \ding{55}}& I & \makecell{CE\\BCE\\GIOU} & Adam &  RTX 3090 & \makecell{DSC: 88.2\%\\Accuracy: 95.7\%\\F1-score: 90.2\%} & PXseg based on BlendMask, mask branch and blender module perform tooth instance segmentation and abnormal detection \\
\midrule

TSegLab \cite{rekik2025tseglab} & 1800 IOS \ding{55}  & I & / & SGD & Titan V & \makecell{DSC: 95.10\% \\ Accuracy: 94.30\%\\F1-score: 98\%} & Coarse-to-fine pipeline: Mask R-CNN for detection/segmentation, GNN for tooth recognition and labeling. \\
\midrule

Seg-CSAB \cite{bajpai2025advanced} & \makecell{523 PAN \ding{51}\\802 PAN \ding{55}} & I & / & SGD & RTX 3090 & \makecell{DSC: 92.39\% \\ Accuracy: 99.90\%\\Specificity: 99.95\%} & ResNet50+FPN two-stage segmentation, CSAB enhances features, Deeper Mask Head improves tooth mask precision. \\
\midrule

ToothInstanceNet \cite{van2025automated} & 716 IOS \ding{51}  & I & / & / & / & \makecell{3DTeethSeg22:\\DSC: 98.24\%\\F1-score: 99.64\%} & ToothInstanceNet performs two-stage point cloud segmentation with Gaussian-based instance generation and MLP labeling for precise tooth segmentation. \\
\midrule

\cite{chen2025tooth} & 39 CBCT  \ding{55}  & I & \makecell{GD\\Focal} & / & RTX 3060 & \makecell{DSC: 96.33±0.89\% \\HD: 2.04±0.87\\ASD: 0.24±0.06} & Multi-model Attention U-Net and V-Net segmentation is fused to correct tooth positions and measure working length and canal angles. \\
\midrule
&  &  &  &  &   &  \multicolumn{2}{c}{\textbf{\small(End of Tabel. \ref{tab2}.)}}\\
\bottomrule
\end{tabular}
\renewcommand{\arraystretch}{1}
\end{table*} 

\clearpage
\begin{table*}[h]
\centering
\tiny
\setlength{\tabcolsep}{2pt}
\caption{A summary of tooth-level dense prediction studies based on ViT and hybrid architectures, arranged in ascending order of publication year. Asterisks (*) indicate that the study involves a detection task, while the symbol (※) denotes that the loss function was proposed by the authors. In the table, \ding{55} represents the use of a private dataset, and \ding{51} represents the use of a public dataset. S/I denotes the segmentation type: semantic segmentation (S) or instance segmentation (I). “/” indicates information not reported in the study; in the S/I column, it denotes that the study did not include a segmentation task. For studies employing more than three evaluation metrics, the three most representative metrics are selected. AUC (Area Under the Curve); IA (Identification Accuracy); GG (Geometry-Guided); CBL (Contrastive Boundary Learning) \cite{tang2022contrastive}; AR (Annotator-Robust) \cite{liu2017richer}.}
\label{tab3}
\renewcommand{\arraystretch}{1.5}
\begin{tabular}{%
c |
m{1.5cm} 
c 
c 
c 
c 
c 
c 
m{5.2cm} 
}

\toprule
Type & Paper & Dataset & S/I & Loss & Optimizer & GPU & Evaluation &\multicolumn{1}{c}{Highlights}\\
\hline

\textbf{\multirow{8}{*}{\makecell{Pure\\ \\ViT}}} &TSegFormer \cite{xiong2023tsegformer} & 16000 IOS \ding{55} & I & \makecell{CE\\※GG} & / & / & \makecell{DSC: 96.01\% \\ mIoU: 94.34\%\\Accuracy: 97.97\%} & TSegFormer encodes point cloud geometry, employing multi-layer self-attention and auxiliary heads for fine tooth–gum boundary modeling. \\
\cline{2-9}

&\cite{sheng2023transformer} & 100 PAN\ding{55}  & S & \makecell{BCE} & Adam & / & \makecell{IoU:\\PLAGH-BH: 46.89\%\\PDX: 69.56\%} & Benchmarked U-Net, LinkNet, FPN, and Swin-Unet for tooth segmentation. \\
\cline{2-9}

&\cite{dai2024sparse} & 158 CBCT \ding{55}  & S & \makecell{Tversky} & Adam & RTX A6000 & \makecell{DSC: 83.84±4.07\% \\Precision: 86.95±3.08\%\\HD: 8.73±1.78mm} & Self-supervised MAE with graph-attention sparse boundary prompts enables CBCT tooth segmentation, improving performance with limited annotations. \\
\cline{2-9}

&\cite{almalki2024self} & 1800 IOS \ding{51}  & S & \makecell{L2\\CD\\MSE} & AdamW & / & \makecell{Accuracy: 98.3\% \\DSC: 97.0\%\\PPV: 98.9\%} & DentalMAE employs a self-supervised Transformer to reconstruct masked mesh embeddings, learning transferable representations for enhanced 3D tooth segmentation. \\
\cline{2-9}

&Tooth-ASAM \cite{wang2025tooth} & \makecell{4938 CBCT \ding{51}\\200 CBCT \ding{55}\\2500 PAN \ding{51}\\30029 IP \ding{51}} & S & \makecell{BCE\\Dice} & Adam & RTX 4090 & \makecell{NC: \\DSC: 95.5\%\\HD95: 0.234mm\\ASSD: 0.023mm} & Tooth-ASAM uses an image encoder to extract feature embeddings and combines them with the prompt encoder to produce masks. \\
\hline
\hline

\textbf{\multirow{45}{*}{\makecell{Hybrid\\ \\Model}}}&\cite{ghafoor2023multiclass} & 540 PAN \ding{55} & I & \makecell{Square Dice} & / & RTX 3090 & \makecell{DSC: 91.02\% \\ Accuracy: 97.26\%\\Specificity 97.30\%} & UNet with Swin-Transformer integrates TAB, using self-attention to enhance tooth boundary features. \\
\cline{2-9}

&\cite{zhang2023multi} & 100 CBCT \ding{55}  & I & \makecell{L1\\CE\\Dice} & Adam & RTX 3090 & \makecell{DSC: 95.1±0.3\% \\ ASD: 0.13±0.02mm\\HD: 1.39±0.24mm} & UNETR++ with EPA modules enables 3D tooth ROI segmentation and centroid prediction. \\
\cline{2-9}

&\cite{almalki2023self} * & 543 PAN \ding{55}  & I & \makecell{L1\\MSE} & AdamW & Tesla V100 & \makecell{$\text{AP}_{box}$: 86.1\% \\ $\text{AP}_{mask}$: 84.6\%} & SimMIM self-supervised pretraining of Swin Transformer improves dental detection and instance segmentation. \\
\cline{2-9}

&\cite{mohan2023deep} * & 1500 PAN \ding{51}  & S & / & / & / & \makecell{DSC: \\1st.: 91.37±0.38\% \\2nd.: 89.69±1.07\%\\3rd.: 89.04±0.68} & Enhanced U-Net integrates Swin Transformer bottleneck, Teeth Attention Block, and multiscale supervision \\
\cline{2-9}

&WCTN \cite{ahn2024weighted} & 1587 IOS \ding{55}  & S & \makecell{CE} & Adam & RTX 3090 & \makecell{mIoU: 88.71\% \\ Accuracy: 94.01\% \\mAccuracy: 93.72\%} & WCTN integrates weighted sparse CNN and Transformer aggregation, employing group attention for efficient local–global feature fusion. \\
\cline{2-9}

&MPRP-LAGC \cite{wang2024multi} & 2692 CBCT  \ding{51}  & S & \makecell{MSE\\BCE\\Dice} & AdamW& / & \makecell{DSC: 91.58\% \\AUC: 92.62\%\\HD: 7.386mm} & MPRP-LAGC integrates dual-branch pretraining and graph convolution for precise tooth segmentation. \\
\cline{2-9}

&Trans-VNet \cite{wang2024trans}  & \makecell{150 CBCT \ding{51} \\ 42 CBCT \ding{55}}& S & / & / & RTX 3090 & \makecell{DSC: \\ MAD: 88.67\%\\Cui: 96.44\%} & Trans-VNet integrates a Transformer into V-Net’s bottleneck, fusing global context with local features under deep supervision. \\
\cline{2-9}

&FlowgateUNet \cite{cao2024flowgateunet} & 15748 CBCT \ding{55}  & S & \makecell{CE\\Dice} & SGD & Tesla P40 & \makecell{microCT dataset:\\DSC: 83.86\% \\ HD95: 20.33mm} & FlowgateUNet integrates FlowFormer and gated attention within U-Net for multi-scale long-range spatial dependency modeling. \\
\cline{2-9}

&\cite{li2024spatial} * & 5634 PAN \ding{51}  & I & \makecell{L1\\CE\\Focal} & Adam & RTX 3090 TI & \makecell{$O^2$PR: \\mAP: 82.4\%\\Accuracy: 99.9\&\\F1-score: 94.8\%} & U-Net dual-path: Transformer encodes local, CNN extracts global, fused via cross-attention for precise tooth instance segmentation. \\
\cline{2-9}

&\cite{guo2024efficient} & 2380 PAN \ding{51}  & I & \makecell{CE\\Dice} & Adam &Tesla V100-SXM2 & \makecell{DSC: 45.58\% \\ NSD: 52.27\%\\IA: 37.73\%} & ResUnet50 with SAM-Med2D leverages semi-supervised learning and data augmentation for enhanced tooth segmentation accuracy. \\
\cline{2-9}

&\cite{wang2024semi} & 2403 PAN \ding{51}  & I & \makecell{CE\\Dice} & SGD & RTX 4090 & \makecell{DSC: 87.07\% \\ NSD: 41.08\%\\IA: 4.78\%} & A semi-supervised teeth segmentation method combines U-Net and Transformer with cross-teaching, jointly training on labeled and unlabeled data. \\
\cline{2-9}

&FDNet \cite{feng2024fdnet} & 9000 CBCT \ding{55}  & S & /& /& / & \makecell{DSC: 85.28±9.78\% \\Precision: 82.67±12.47\%\\Recall: 89.77±10.99\%} & FDNet integrates LF-Wavelet and SAM encoder to enhance low-frequency semantics and refine boundaries, achieving accurate tooth segmentation. \\
\cline{2-9}

&DICL \cite{cai2024deformable} & 2409 PAN \ding{55} & S & \makecell{CE\\MSE\\Dice} & SGD & RTX 3060 & \makecell{DSC: 23.85\% \\ NSD: 66.53\%\\IA: 60.47\%} & Two-phase tooth segmentation: semi-supervised pseudo-labels, fully supervised refinement with deformable convolutions and consistency loss. \\
\cline{2-9}

&CrossTooth \cite{xi20253d} & 1800 IOS \ding{51}  & S & \makecell{CE\\CBL} & Adam & RTX 3090 & \makecell{mIoU: 95.86\% } & CrossTooth fuses point cloud and multi-view image features with selective downsampling and boundary-aware learning. \\
\cline{2-9}

&EASAM \cite{zhang2025easam} & 5273 PAN \ding{51}  & S & \makecell{IOU\\BCE} & Adam & RTX 3090 & \makecell{DSC: 89.03±4.96\% \\ Specificity: 98.93±0.16\%\\HD: 40.84±13.27mm} & EASAM integrates ViT and edge branches, using edge-aware and cross-branch interaction for medical imaging. \\
\cline{2-9}

&LSAFormer \cite{wang2025lsaformer} & 1800 IOS \ding{51} & I & \makecell{CE} & Adam & / & \makecell{mIoU: 90.5\% \\F1-score: 93.7\%} & LSAFormer integrates local surface reconstruction and global feature perception for multi-scale point cloud segmentation. \\
\cline{2-9}

&\cite{wathore2025bilateral} & 3909 PAN \ding{51} & S & \makecell{CE\\Dice} & / & / & \makecell{DSC:\\UNet: 66.96\% \\ SE-UNet: 69.89\%\\TransUNet: 76.69\%} & Leverages left-right symmetry of panoramic X-rays to create flipped images, augmenting data with consistent tooth labeling. \\
\cline{2-9}

&DenUnet \cite{manzari2024denunet} & 543 PAN \ding{51}  & I & \makecell{AR\\BCE\\Dice} & SGD & RTX Titan & \makecell{mAP: 71.32\% \\AP50: 97.81\%\\AP75: 88.18\%} & DenUnet is a dual-encoder network (PDC Edge + Swin Transformer) with Local Cross-Attention Fusion for dental segmentation. \\
\cline{2-9}

&UCL-Net \cite{xu2025tooth} * & 6204 PAN \ding{51} & I & \makecell{CE} & AdamW & RTX 3090 & \makecell{Recall:\\TDD: 58.7\% \\ DentalX: 61.9\% \\ CDD: 66.3\% } & It leverages an encoder–decoder framework with cross-scale attention to perform tooth instance segmentation \\
\cline{2-9}

&TSNEN \cite{liu2025two} & 200 IOS \ding{51}  & I & \makecell{Dice} & Adam & RTX 3090 & \makecell{Accuracy: 96.83\% \\ mIoU: 92.06\%} & TSNEN dual-stream network fuses coordinate and geometric features, using attention and pooling for 3D tooth mesh segmentation. \\
\cline{2-9}

&MDGFormer \cite{pan2025mdgformer} & 805 IP \ding{55} & S & \makecell{BCE\\IOU\\MSE} & Adam & RTX 4090 & \makecell{F-measure(BS): 97.7 \%} & MDGFormer achieves tooth segmentation via multi-domain feature extraction and interaction, while incorporating a super-resolution task to recover high-frequency details. \\
\bottomrule

\end{tabular}
\renewcommand{\arraystretch}{1}
\end{table*} 

\clearpage
\begin{table*}[h]
\centering
\tiny
\setlength{\tabcolsep}{2.5pt}
\caption{A summary of tooth anatomical-level segmentation studies, arranged in ascending order of publication year. Asterisks (*) indicate that the study involves a detection task, while the symbol (※) denotes that the loss function was proposed by the authors. In the table, \ding{55} represents the use of a private dataset, and \ding{51} represents the use of a public dataset. For studies employing more than three evaluation metrics, the three most representative metrics are selected. Shrinkage \cite{lu2018deep}; Lovász-hinge \cite{berman2018lovasz}; FTM (Focal Tree-Min) \cite{li2022deep}; FD (Fourier Description).}
\label{tab4}
\renewcommand{\arraystretch}{0.8}
\begin{tabular}{%
m{1.5cm} 
c 
c 
c
c 
c 
c 
c 
m{5cm} 
}
\toprule
Paper & Dataset & Targets & Methods & Loss & Optimizer & GPU & Evaluation &\multicolumn{1}{c}{Highlights}\\
\midrule

\cite{shi2021panoramic} & 798 PAN \ding{55}&  \makecell{Root} & CNN &/& / & / & \makecell{Precision: \\Original: 88.83\% \\ Sobel: 77.85\%\\Canny: 72.76\%} & Patch-based edge segmentation using LeNet-5, classifying pixel centers and connecting them to form tooth contours. \\
\midrule

GT U-Net \cite{li2021gt} & 248 PR \ding{55} &\makecell{Root}  & Hybrid & \makecell{※FD} & Adam &\makecell{RTX- \\ 2080 Ti} & \makecell{DSC: 92.54± 0.25\% \\Accuracy: 93.98± 0.11\%\\Specificity: 94.79± 0.14\%} & GT U-Net integrates Group Transformer and MHSA, progressively modeling features from local to global. \\
\midrule

RCS-Net \cite{zhang2021root} & 78 CBCT \ding{55}  & \makecell{Root Canal}  & CNN & \makecell{Dice} & Adam & \makecell{Titan- \\ Xp} & \makecell{DSC: 95.2\% \\VOE: 0.077mm\\ASD : 0.125mm} & RCS-Net extends 3D U-Net with dual-stage segmentation and multi-task loss for precise root canal detection. \\
\midrule

\cite{oztekin2022automatic} & 250 PAN \ding{55}  & \makecell{Dental\\ Prosthesis} & CNN & \makecell{Focal} & Adam & / & \makecell{mIoU: 76.7\% \\Accuracy: 99.81\%} & U-Net segments amalgam and composite resin fillings in panoramic dental radiographs automatically. \\
\midrule

\cite{rousseau2023pre} & 2500 BR \ding{55}  & \makecell{Dentin\\Enamel\\Pulp}& CNN & / & Adam & \makecell{Tesla-\\T4} & \makecell{All: \\mIoU: 76.96\%} & DDPM-pretrained Unet with fine-tuning enables few-shot multi-class anatomical-level semantic segmentation. \\
\midrule

\cite{wang2023root} & 215 CBCT \ding{55}  & \makecell{Tooth \\ Pulp} & CNN & \makecell{CE,L2\\Focal,Smo\\Shrinkage\\Lovász-hinge} & Adam & \makecell{Tesla- \\ V100} & \makecell{Pulp:\\DSC: 87.5±3.1\% \\ HD: 0.956±0.511mm} & DentalNet instance segmentation with 3D U-Net ROI refinement using spatial embedding clustering for tooth recognition. \\
\midrule

\cite{kim2023automatic} & 70 CBCT \ding{55}  & \makecell{Enamel\\Dentin\\Pulp} & CNN & \makecell{CE,BCE\\DICE\\FTM} & / & / & \makecell{Enamel: \\DSC: 85.65±0.29\%\\ HD95: 1.37±0.26mm} & Three-stage CBCT segmentation with DHU-Net and HFF module for multi-level tooth structure recognition. \\
\midrule

\cite{tan2023dental} & 200 CBCT \ding{55}  &\makecell{Enamel\\Dentin\\Pulp}& CNN & \makecell{BCE\\DICE} & Adam & \makecell{Tesla- \\ V100} & \makecell{Enamel: \\DSC: 97.8\% \\ HD95: 0.132mm} & HMG-Net instance segmentation with tooth skeleton and multi-scale Frangi filter improves CBCT pulp and dentin segmentation. \\
\midrule

DSIS-DPR \cite{wang2024dsis} & 29112 PAN  \ding{55}  & \makecell{Tooth\\Enamel\\Dentin\\Pulp}& CNN & \makecell{CE\\BCE} & Adam & / & \makecell{Enamel:\\ DSC: 79.14\% \\ HD95: 14.39mm} & Three-stage tooth structure segmentation: DSIS fine segmentation, ADM anomaly detection, DPR diffusion-based refinement. \\
\midrule

\cite{tan2024progressive}* & 314 CBCT \ding{55}  & \makecell{Tooth\\Enamel\\ Dentin\\Pulp} & CNN & \makecell{CE\\SL1\\DICE} & / & / & \makecell{Enamel: \\DSC: 92.85±0.04\% \\HD95: 0.297±0.22\%} & Progressive three-stage tooth and substructure segmentation: centroid detection, instance segmentation, enamel/dentin/pulp segmentation. \\
\midrule

\cite{li2024accurate} & 8892 CBCT  \ding{51}  & \makecell{Enamel\\Dentin\\Pulp} & CNN & \makecell{BCE\\SSIM} & Adam & \makecell{RTX- \\ 2070} & \makecell{Enamel \& Dentin: \\DSC: 95.79±0.58\% \\ASSD: 0.24±0.08mm} & A hybrid pipeline with a multi-scale attention U-Net for pulp segmentation and dual level set models for enamel-dentin segmentation, ensuring boundary accuracy and inter-slice consistency. \\
\midrule

D3UNet \cite{liang2024dual} & 300 CBCT \ding{51}  & \makecell{Tooth\\Root Canal} & CNN &\makecell{GD\\CE\\Focal} & Adam & \makecell{RTX \\4090} & \makecell{Root: \\DSC: 94.15\%\\ASSD: 0.1357mm} & Dual-view framework localizes ROI globally and refines locally; 2.5D fusion and dual-boundary loss improve accuracy. \\
\midrule

\cite{yu2025clinically} & 30 CBCT  \ding{55}& \makecell{Enamel}  & CNN & / & / & \makecell{RTX- \\ 5000} & \makecell{DSC: 96.6\%} & 2.5D Attention U-Net segments CBCT enamel, using early stopping, external validation, and 3D thickness computation. \\
\midrule

\cite{zhou2025prad} & 10k PR \ding{51}& \makecell{Anatomical- \\ structures\& \\ instruments}  & CNN & \makecell{CE \\ DICE} & \makecell{Adam} & \makecell{RTX- \\ 3090} & \makecell{DSC: 96.6\%} & PRAD-10K dataset and PRNet provide high-quality annotations and benchmarks for periapical radiograph segmentation. \\
\bottomrule

\end{tabular}
\renewcommand{\arraystretch}{1}
\end{table*} 
\clearpage

\noindent \cite{rousseau2023pre} proposed a pretraining strategy based on the denoising diffusion probabilistic model (DDPM), where a U-Net is pretrained on a large corpus of unlabeled data and subsequently fine-tuned on a limited set of annotated samples to achieve segmentation of dentin, enamel, and pulp in BR. As shown in Fig. \ref{fig15}, Kim et al. \cite{kim2023automatic} developed DHU-Net, which incorporates a hierarchical feature fusion module and a spatial attention mechanism to extract multi-scale features from individual tooth patches and differentiate between dentin, enamel, and pulp, achieving strong performance on CBCT images. Wang et al. \cite{wang2024dsis} introduced a three-stage model that first employs a GAN-based architecture to generate anatomical-level segmentation masks of teeth, followed by iterative refinement through an anomaly detection module and a diffusion prior restoration module. Their method achieved superior results on PAN images from 996 patients, outperforming previous models.

\subsubsection{Dental Diseases}
\begin{figure}[t]
	\centerline{\includegraphics[width=\linewidth]{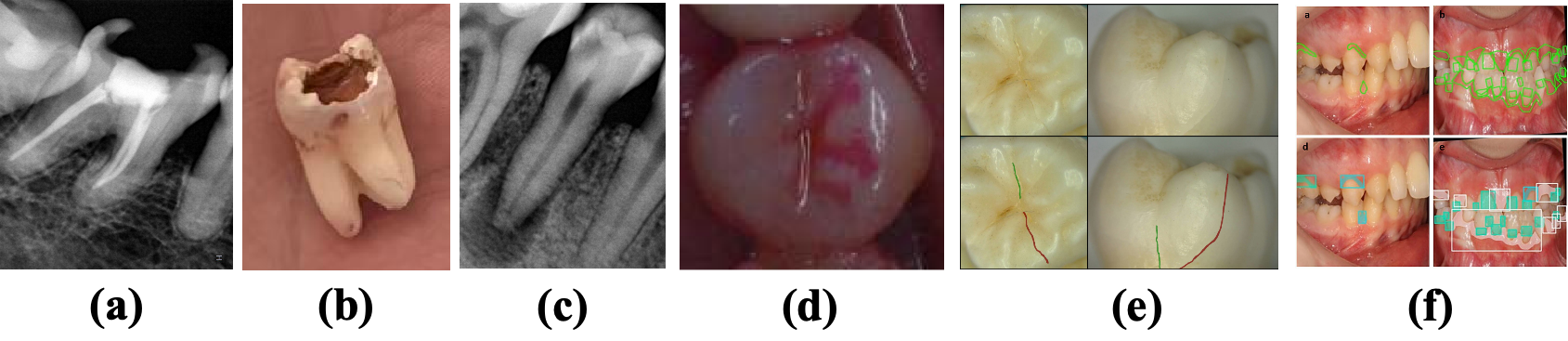}}
	\caption{Illustrations of common dental diseases. Panels (a) and (c) are sourced from \cite{zhou2025prad}, showing dental caries and periapical periodontitis, respectively. Panel (b) depicts an impacted third molar with caries, obtained from an extracted tooth by the author. Panel (d) shows dental plaque \cite{shi2022semantic}; panel (e) illustrates a tooth crack \cite{huang2025automatic}; and panel (f) displays dental calculus, gingival hyperplasia, and inflammation \cite{yurdakurban2025automated}.}
	\label{fig16}
\end{figure}

\begin{figure}[h]
	\centerline{\includegraphics[width=\linewidth]{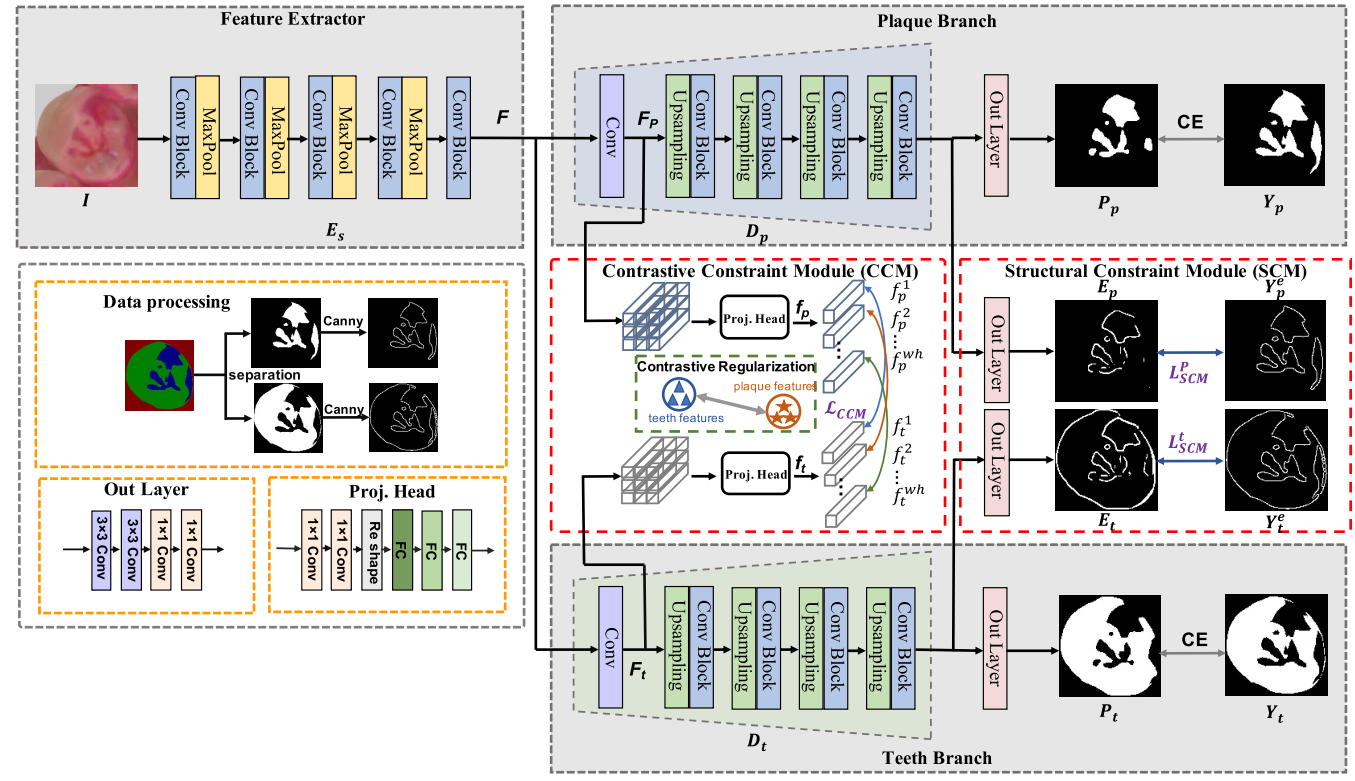}}
	\caption{SDNet for tooth plaque segmentation utilizes a shared encoder and dual decoders for semantic disentanglement. The Class Conditional Module enhances class separability, while the Spatial Conditioning Module refines spatial details through boundary supervision, thereby improving accuracy and robustness. Image sourced from \cite{shi2022semantic}.}
	\label{fig17}
\end{figure}

Common dental diseases encountered in clinical practice are illustrated in Fig. \ref{fig16}. As shown in Fig. \ref{fig19}, by accurately identifying pathological features, such models can provide clinicians with objective and quantitative references, thereby improving diagnostic efficiency and treatment precision. Dental plaque represents one of the most characteristic dental diseases. Unlike most pathologies that rely on radiographic image analysis, plaque detection is typically performed using stained IP. Li et al. \cite{li2020low, li2022automatic} proposed a two-stage architecture that integrates CNNs with heat kernel signatures and local binary patterns to extract local features, thereby enhancing the network’s ability to learn shapes and boundaries and achieving efficient plaque segmentation. Similarly, as shown in Fig. \ref{fig17}, Shi et al. \cite{shi2022semantic} developed a contrastive learning-based dual-decoder model that employs a shared encoder to extract image features, with two independent decoders to predict plaque and tooth masks, respectively. The model leverages contrastive learning to enhance the separation between the two decoders in latent space and employs auxiliary supervision to produce sharper segmentation boundaries.
 
\begin{figure}[t]
	\centerline{\includegraphics[width=\linewidth]{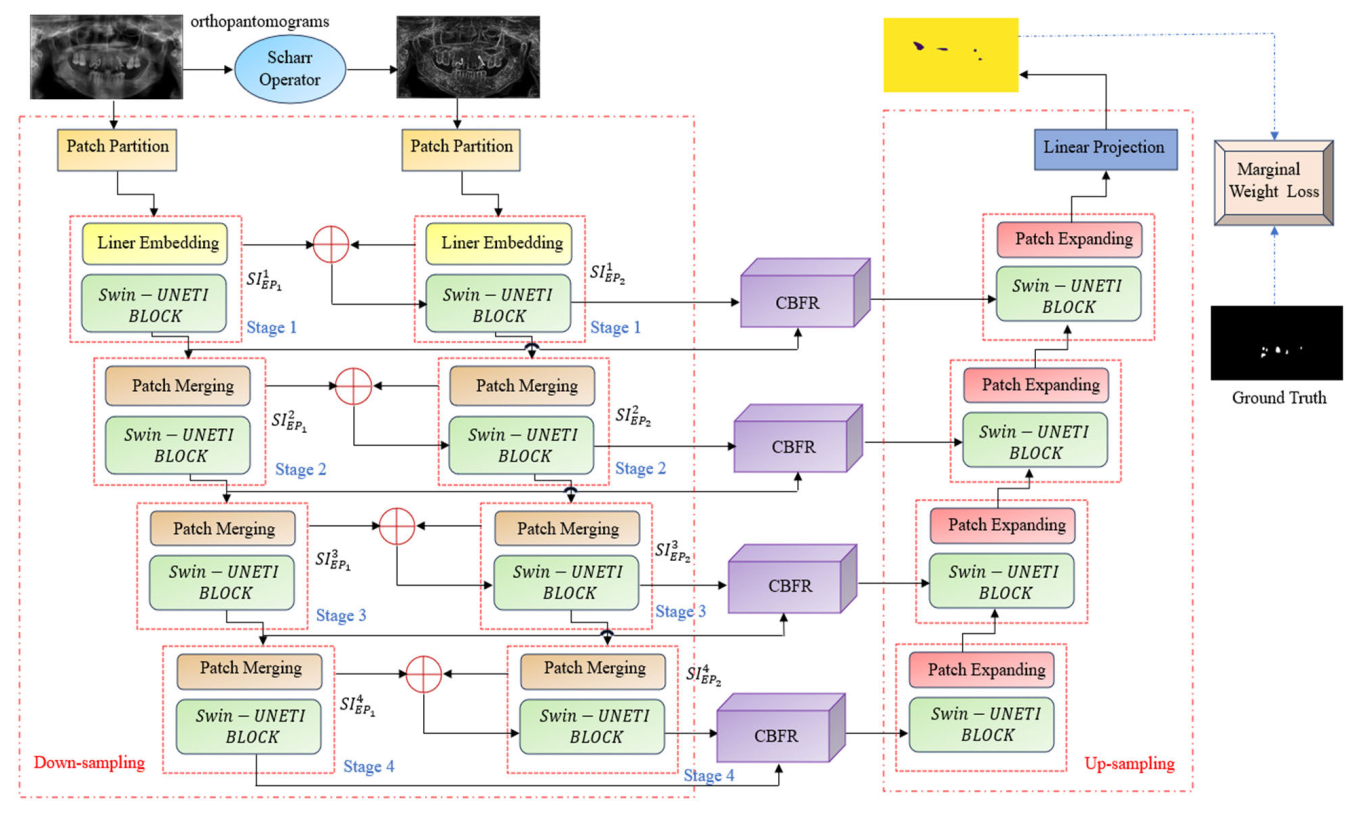}}
	\caption{DentC\_BSFLNet for caries segmentation employs a multi-path encoder-decoder architecture with Swin\-UNET blocks and a CBFRM module to fuse multi-scale features. A boundary-weighted loss function is utilized to enhance boundary accuracy and sensitivity. Image sourced from \cite{neeraja2025dentc_bsfl}.}
	\label{fig18}
\end{figure}

 The prediction of caries, periapical lesions, and other abnormalities has become a major research focus in the field of dental image analysis. For caries prediction, Wang et al. \cite{wang2023multi_} proposed a semi-supervised framework based on uncertainty estimation to segment carious regions in PAN images. This framework employs deterministic masks derived from Monte Carlo aggregation to ensure consistency between predictions of the teacher and student models. Zhu et al. \cite{zhu2023cariesnet} introduced CariesNet, a U-shaped network with deep supervision that incorporates a full-scale axial attention module to simultaneously capture boundary cues and extract fine-grained feature maps that integrate high-level semantics and low-level details, achieving multi-stage caries segmentation in PAN images. Liu et al. \cite{liu2025automated} further developed a cascaded learning architecture that integrates YOLOX \cite{ge2021yolox} and FCN \cite{long2015fully} for accurate caries detection and segmentation in PR images. For the detection of periapical diseases and other oral  

\clearpage
\begin{table*}[h]
\centering
\tiny
\setlength{\tabcolsep}{3.5pt}
\caption{A summary of dental disease dense prediction studies, arranged in ascending order of publication year. Asterisks (*) indicate that the study involves a detection task, while the symbol (※) denotes that the loss function was proposed by the authors. In the table, \ding{55} represents the use of a private dataset, and \ding{51} represents the use of a public dataset. When multiple dental lesions are involved, “Multiple Abnormal Conditions” is used to describe the targets. For studies employing more than three evaluation metrics, the three most representative metrics are selected. WT (Wisdom Teeth); RR (Residual Roots); WS (Wear Surface); DC (Dental Calculus); Combo \cite{taghanaki2019combo}; FT (Focal Tversky) \cite{abraham2019novel}; MW (Marginal Weight) \cite{tran2025litenext}; BD (Boundary Dou) \cite{sun2023boundary}; B-CE (Bootstrapped-CE) \cite{chao2019hardnet}; PA (Pixel Accuracy); FWIoU (Frequency Weighted IoU).}
\label{tab5}
\renewcommand{\arraystretch}{0.6}
\begin{tabular}{%
m{1.5cm} 
c 
c 
c
c 
c 
c 
c 
m{6cm} 
}
\toprule
Paper & Dataset & Targets & Methods & Loss & Optimizer & GPU & Evaluation &\multicolumn{1}{c}{Highlights}\\
\midrule

Dense U-NET \cite{zheng2020anatomically} & 20 CBCT \ding{55}  & \makecell{Periapical \\ Lesions}  & CNN & \makecell{CE\\Dice\\Focal} & Adam & / & \makecell{Precision: 90.0\%\\Recall: 84.0\%} & Dense U-Net from modified FC-DenseNet with anatomical constraints for multi-class dental CBCT segmentation. \\
\midrule

\cite{li2020low} & 607 IP \ding{55}  & \makecell{Dental \\ Plaque} & CNN & /& / & \makecell{GTX-\\1080} & \makecell{IOU: \\Children: 85.9\% \\Adults: 64.0\%} & DeepLabV3+ with SLIC superpixels and multi-feature fusion, combined with random forest for plaque segmentation. \\
\midrule

\cite{li2022automatic} & 2884 IP \ding{51} &\makecell{Dental \\ Plaque} & CNN & CE & / & \makecell{GTX-\\1080} & \makecell{PA: 84.59\% \\ mIoU: 73.64\%} & Self-attention CNN with local-to-global feature fusion for automatic dental plaque segmentation. \\
\midrule

SDNet \cite{shi2022semantic} & 1869 IP \ding{55}  & \makecell{Dental \\ Plaque} & CNN & \makecell{CE\\BCE\\DICE} & Adam & \makecell{Tesla-\\V100} & \makecell{DSC: \\ Seg-S: 95.58\% \\Seg-C: 94.94\%} & SDNet with shared encoder and dual decoders, enhanced by contrastive and structural constraints for plaque segmentation. \\
\midrule

TripleNet \cite{chen2022tooth} & 2884 IOS \ding{55}  & \makecell{Tooth-\\ Defect} & CNN &\makecell{Focal} & Adam & \makecell{RTX-\\3090}& \makecell{IoU: 77.57\% \\ Precision: 83.75\%\\F1-score: 86.25\%} & A triple-stream multi-scale EdgeConv network integrates 18D geometric features for point-wise defect detection. \\
\midrule

MLUA \cite{wang2023multi_} & 1000 PAN \ding{51}  & \makecell{Caries}& CNN & \makecell{CE\\Dice} & AdamW & \makecell{RTX-\\3090} & \makecell{DSC: 61.40\% \\Sensitivity: 58.77\%\\Precision: 70.04\%} & Teacher-student framework with EMA and Monte Carlo sampling generates uncertainty masks for consistency learning. \\
\midrule

CGA-UNet \cite{wang2023cga} & 591 IP \ding{55}& \makecell{Multiple\\Abnormal\\Conditions} & Hybrid & \makecell{B-CE}& Adam& \makecell{GTX-\\3090 Ti} & \makecell{mIoU: 63.21±1.47\%\\FWIoU: 94.35±0.84\%\\Accuracy: 96.97±0.51\%} & CGA-UNet decomposes dental images into low- and high-frequency features for lesion enhancement. \\
\midrule

\cite{hadzic2023teeth} & 144 CBCT \ding{55}  & \makecell{Periapical \\ Lesions} & CNN & \makecell{FT\\Combo\\Focal} & Adam &Titan X & \makecell{DSC:\\ FL: 67±3\%\\FTL: 70±4\%\\CL: 70±4\%} & SCN predicts 3D tooth coordinates and crops regions; U-Net segments apical lesions. \\
\midrule

\cite{felsch2023detection} & 18,179 IP \ding{55}  & \makecell{Caries\\Hypominer-\\alization}& VIT &/& / & \makecell{RTX-\\A6000} & \makecell{IoU: 95.9\% \\ Accuracy: 97.8\%\\F1-score : 97.7\%} & SegFormer-B5 visual Transformer trained with DeepSpeed multi-server setup and augmented images simulating diverse tooth imaging conditions. \\
\midrule

CariesNet  \cite{zhu2023cariesnet}& 1159 PAN \ding{55} & \makecell{Caries}& CNN & \makecell{BCE\\Dice} & Adam & \makecell{RTX- \\2080 Ti} & \makecell{DSC: 93.64\% \\Accuracy: 93.61\%\\F1-score: 92.87\%} & CariesNet employs U-Net with Res2Net backbone and full-scale axial attention to fuse multi-level features for caries saliency prediction. \\
\midrule

LABANet \cite{chen2023labanet} * & 1000 PAN \ding{55} & \makecell{Caries\\RR\\WT}& Hybrid & /& Adam & \makecell{Tesla-\\A40} & \makecell{DSC: \\ WT:86.0\% \\RR: 90.3\%\\Caries: 83.6\%} & LABANet employs dual Swin Transformer backbones with enhanced SE attention head for multi-lesion oral segmentation. \\
\midrule

\cite{ma2023semi} & 200 IP \ding{55} & \makecell{WS\\DC\\Gingivitis}& CNN & \makecell{CE\\Dice\\$L_{adv}$} & Adam & \makecell{GTX-\\3090} & \makecell{mIoU: 72.79\% \\Accuracy: 83.62\%\\F1-score: 81.31\%} & DGAN-based dental lesion segmentation combines supervised and semi-supervised learning, enhanced DeepLabV3+ for multi-scale lesions. \\
\midrule

BiseFormer \cite{ma2024lightweight}& 400 IP \ding{55}  & \makecell{Gingivitis\\DC\\WS}& Hybrid & \makecell{CE} & / & \makecell{RTX-\\3090} & \makecell{mIoU: 79.76\% \\mAcc: 87.7\%} & BiseFormer enables precise dental lesion segmentation via ConvFormer blocks, edge attention, and a hybrid CNN–ViT architecture. \\
\midrule

CenterFormer \cite{song2024centerformer}& 4884 IP\ding{51}  & \makecell{Dental\\Plaque}& Hybrid & / & / & / & \makecell{mIoU: 75.86\% \\mPA: 85.94\%\\Accuracy: 86.51\%} & CenterFormer uses a Cluster Center Encoder, multiple granularity perceptions, and a pyramid hybrid encoder for precise dental image segmentation. \\
\midrule

\cite{jiang2024frequency} & 2884 IP \ding{51}  & \makecell{Dental\\Plaque}&CNN & \makecell{CE\\BCE} & / & / & \makecell{mIoU: 74.02\% \\mAcc: 84.04\%} & FGN uses a high-to-low frequency dual-task pipeline with frequency-guided decoupling and a frequency-driven refinement module for plaque segmentation. \\
\midrule

\cite{koshy2024dental}& 560 PAN \ding{51} & \makecell{Caries} & CNN & \makecell{BCE} & Adam & / & \makecell{Accuracy: 89\%} & This study uses U-Net with encoder-decoder and skip connections to detect dental caries. \\
\midrule

\cite{liu2025automated} * & 5569 PR \ding{55}  & \makecell{Caries} & CNN & \makecell{Dice} & SGD & \makecell{RTX-\\3090} & \makecell{DSC: 84.6\%} & Cascaded YOLOX detection and FCN-8s segmentation to accurately detect and segment dental caries on periapical radiographs. \\
\midrule

\cite{yurdakurban2025automated} * & 1000 IP \ding{55}& \makecell{Multiple\\Abnormal\\Conditions} & CNN & \makecell{Tversky}& /&  \makecell{Tesla-\\T4} & \makecell{F1-score: \\ Y: 63.0±0.7\% \\U+R: 64.0±4.6\%} & YOLOv8 for fast multi-object detection, and U-Net with ResNet50 for precise high-resolution lesion segmentation. \\
\midrule

DentC-BSFLNet \cite{neeraja2025dentc_bsfl} & 153 PAN \ding{55}  & \makecell{Caries}& Hybrid & \makecell{MW} & AdamW &\makecell{ RTX-\\3050} & \makecell{DSC: 95.93\% \\Accuracy: 98.53\%\\Precision: 96.41\%} & DentC-BSFLNet uses a dual-path Swin-UNETI encoder, boundary refinement, and marginal weight loss for accurate caries boundary segmentation. \\
\midrule

AEDD-Net \cite{huang2025precise} & 389 PR \ding{55}  & \makecell{Caries}& Hybrid & \makecell{BD\\Dice\\※Edge} & SGD & / & \makecell{DSC: 76.8\% \\Precision: 82.7\%\\Sensitivity: 74.9\%} & AEDD-Net with a CNN-Transformer hybrid encoder and parallel CA-Decoder, Edge-Decoder for precise caries segmentation, and a multi-loss function with boundary optimization for enhanced edge accuracy. \\
\midrule

\cite{jones2025dental} & 4396 IP \ding{55}& \makecell{Caries} & CNN & \makecell{Focal}& AdamW&  \makecell{RTX-\\4090} & \makecell{Precision:\\MB: 99\%\\VASC: 99\%} & Attention U-Net with a frequency-guided dual-task framework enables accurate boundary segmentation of children's dental caries. \\
\midrule
 
\cite{huang2025automatic}& 2210 IP \ding{51}& \makecell{Cracked-\\Tooth}&CNN & \makecell{BCE\\SL1}&SGD&\makecell{Tesla-\\V100} & \makecell{AP: 79.51\%\\AP50: 95.75\%\\AP75: 86.78\%} & An improved Mask R-CNN with ResNeXt backbone, crack feature enhancement, and dynamic snake convolution enables boundary-aware tooth crack segmentation. \\
\midrule

\cite{tez2025deep}& 506 IP \ding{55}& \makecell{Dental\\Plaque}& \makecell{CNN\\Hybrid} &/&\makecell{Adam\\SGD\\RMSProp}&/& \makecell{Precision: \\ 1st.: 87.82\%\\2nd.: 82.03\%\\3rd.: 80.06\%} & Six SOTA DL models with U-Net variants for automated dental plaque segmentation in children, and a statistical comparison with dentists for clinical validation.. \\
\midrule

\cite{hussein2025deep} * & 9573 PAN \ding{55}  & \makecell{Multiple\\Abnormal\\Conditions} & \makecell{CNN\\VIT}  &/& AdamW & / & \makecell{mAP50: \\YOLO: 61\%\\RT-DETR: 60\%} & RT-DETR: Transformer-based, detects overlapping dental X-rays via global attention; YOLO: CNN-based, fast single-stage detection. \\
\midrule

\cite{ari2025evaluation} & 300 CBCT \ding{55}  & \makecell{Multiple\\Abnormal\\Conditions}& CNN & /& SGD &\makecell{Tesla-\\V100} & \makecell{Caries: \\ Sensitivity: 91.66\% \\Precision: 97.84\%\\AUC: 86.79\%} & Mask R-CNN with a ResNet101 backbone detects and segments 15 dental conditions in CBCT slices via region-based CNNs with mask prediction. \\
\bottomrule
\end{tabular}
\renewcommand{\arraystretch}{1}
\end{table*} 
\clearpage

\begin{figure*}[!t]
	\centerline{\includegraphics[width=\linewidth]{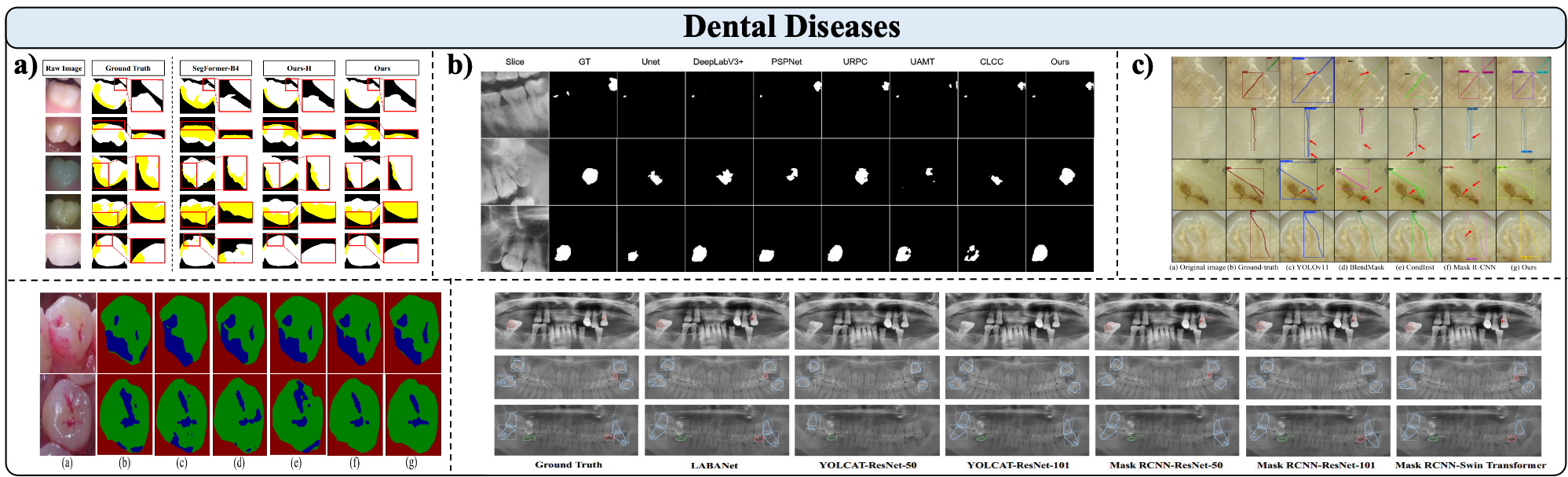}}
	\caption{Visualization of dense detection results for dental lesions. Panels (a) and (b) display dental plaque, sourced from \cite{jiang2024frequency} and \cite{shi2022semantic}, respectively; panel (c) depicts tooth cracks from \cite{huang2025automatic}; panel (d) illustrates impacted third molars, sourced from \cite{chen2023labanet}; and panel (f) presents trends from \cite{wang2023multi_}.}
	\label{fig19}
\end{figure*}
\FloatBarrier

\noindent abnormalities and infections, CNN-based architectures remain the mainstream research paradigm. Zheng et al. \cite{zheng2020anatomically} proposed an enhanced Dense U-Net that incorporates anatomical constraints during training, embedding oral structural knowledge into the network to achieve accurate segmentation of periapical lesions, tooth structures, and fillings in CBCT images. Chen et al. \cite{chen2022tooth} designed an edge convolution-based network that fuses outputs from multi-scale convolutional branches, enhancing the model's adaptability to diverse tooth surface morphologies and successfully predicting tooth defect regions in IOS data.

Wang et al. \cite{wang2023cga} developed CGA-UNet, which decomposes the input into high- and low-frequency components and applies a class-guided attention mechanism to each frequency branch for feature reweighting. The fused representations are integrated into the encoder-decoder structure of an Attention-UNet, achieving accurate segmentation of multiple abnormal structures in IP images. Chen et al. \cite{chen2023labanet} proposed LABANet for multi-lesion segmentation in oral radiographs. It employs a Swin Transformer-based backbone and an auxiliary network. Their outputs are fused through a region proposal network and refined by an attention-based FCN, yielding precise identification of caries, residual roots, and wisdom teeth on the PAN dataset.

\subsubsection{Maxillofacial Structure}
\begin{figure*}[!t]
	\centerline{\includegraphics[width=\linewidth]{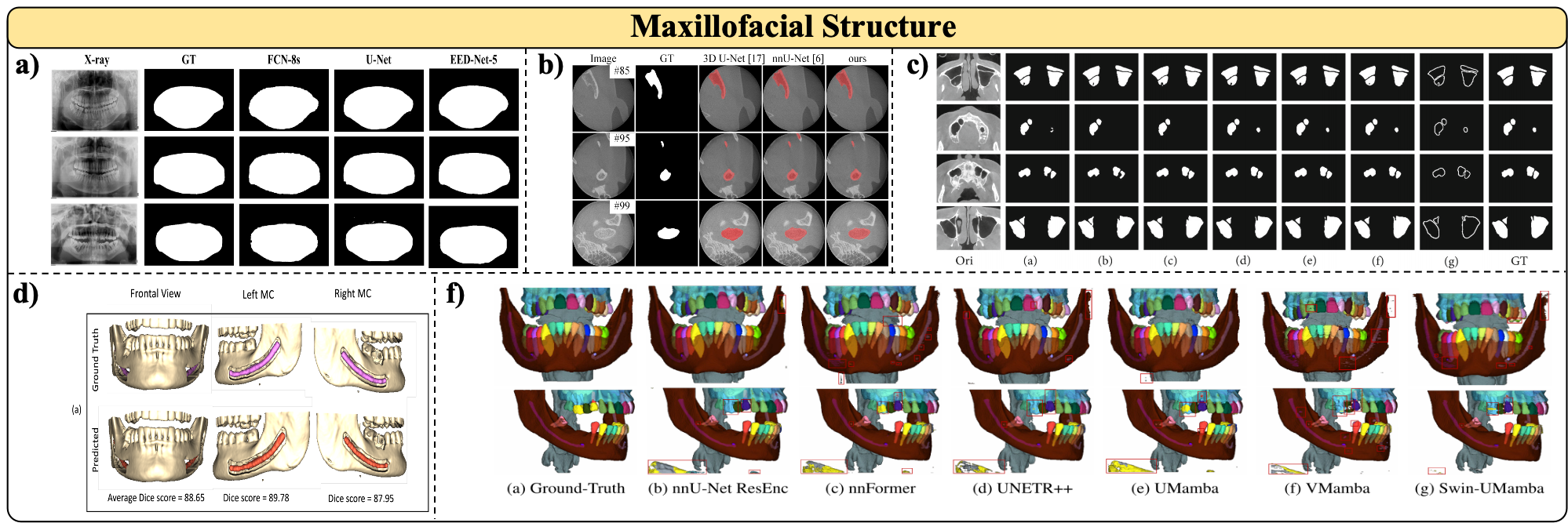}}
	\caption{Visualization of segmentation results for maxillofacial structures. Panel (a) shows the maxillofacial region from \cite{kong2020automated}; panel (b) shows the temporomandibular joint from \cite{zhang20213d}; panel (c) presents the maxillary sinus from \cite{zhang2023dba}; panel (d) illustrates the segmentation of the mandibular canal from \cite{usman2022dual}; and panel (f) depicts multi-structure segmentation results in the ToothFairy2 dataset, from \cite{bolelli2025segmenting}.}
	\label{fig20}
\end{figure*}

As shown in Fig. \ref{fig20}, the maxillofacial structure comprises multiple critical anatomical structures, including the maxilla, mandible, mandibular canal, inferior alveolar nerve, mental foramen, temporomandibular joint, maxillary sinus, and hard palate. These structures play vital roles in mastication, speech, and sensory function. Accurate prediction of these regions is essential for safe surgical planning, dental implant placement, and the prevention of nerve injury.

According to existing studies, predictions of maxillofacial structures are typically based on relatively fixed imaging modalities such as CBCT or PAN, as radiographic imaging inherently excels in visualizing internal anatomical structures. In the early stages of research, Dasanayaka et al. \cite{dasanayaka2019segmentation} proposed a semi-automatic segmentation framework for the mental foramen based on U-Net, achieving accurate prediction through customized image augmentation strategies. Kong et al. \cite{kong2020automated} developed EED-Net, which introduced a multi-path feature extraction module into the bottleneck of the U-Net, thus improving the breadth of encoder features and maintaining high-resolution representations for efficient segmentation of maxillofacial ROIs. Zhang et al. \cite{zhang20213d} proposed a multi-directional resampling mechanism that generated new input images from three orthogonal directions, enabling the model to capture spatial information from multiple perspectives and successfully segment the temporomandibular joint in CBCT. Lang et al. \cite{lang2022localization} designed a multi-region proposal network for landmark localization in CBCT images, generating multiple landmark candidates, and an FCN-based recognition network predicted bone masks from these proposals, effectively capturing the structural information surrounding the maxillofacial landmarks. With the expansion of research efforts and the increasing availability of large-scale datasets, more maxillofacial structures have been incorporated into DL studies, and hybrid architectures have gradually emerged. Jarnstedt et al. \cite{jarnstedt2022comparison} constructed a private large-scale CBCT dataset containing 1,103 patient scans and conducted a comprehensive performance analysis of various CNN models for mandibular canal segmentation. Amorim et al. \cite{amorim2023mandible} employed a 3D U-Net with residual modules, achieving efficient and stable segmentation of mandibular regions across 476 CBCT scans. Zhang et al. \cite{zhang2023dba} proposed a two-stage model named DBA-UNet, where the first stage, a boundary cross-fusion network, consists of two parallel U-Nets of identical size to refine boundary representations,  

\clearpage
\begin{table*}[h]
\centering
\tiny
\setlength{\tabcolsep}{3.5pt}
\caption{A summary of maxillofacial structure dense prediction studies, arranged in ascending order of publication year. Asterisks (*) indicate that the study involves a detection task, while the symbol (※) denotes that the loss function was proposed by the authors. In the table, \ding{55} represents the use of a private dataset, and \ding{51} represents the use of a public dataset. For studies employing more than three evaluation metrics, the three most representative metrics are selected. MF (Mental Foramen); LLC (Landmark Localization Craniomaxillofacial); TMJ (Temporomandibular Joint); MS (Maxillary Sinus); AB (Alveolar Bone); MC (Mandibular Canal); IAN (Inferior Alveolar Nerve); SMCD (Symmetric Mean Curve Distance); MSE (Mean Squared Error); FD (Focal Dice); SD (Soft Dice).}
\label{tab6}
\renewcommand{\arraystretch}{0.8}
\begin{tabular}{%
m{1.5cm} 
c 
c 
c
c 
c 
c 
c 
m{5cm} 
}

\toprule
Paper & Dataset & Targets & Methods & Loss & Optimizer & GPU & Evaluation &\multicolumn{1}{c}{Highlights}\\
\midrule

\cite{dasanayaka2019segmentation} & 1000 PAN \ding{55}  & \makecell{MF}& CNN & / & / & / & \makecell{DSC: 98.7\%} & U-Net-based semi-automatic DPT mental foramen segmentation with CLAHE, ROI cropping, and multi-augmentation. \\
\midrule

EED-Net \cite{kong2020automated} & 2602 PAN \ding{55} & \makecell{Maxillofacial}& CNN & \makecell{BCE\\Dice} & Adam & \makecell{RTX- \\ 2080 Ti} & \makecell{IoU: 98.29\% \\Accuracy: 99.28\% \\ HD: 8.32mm} & EED-Net combines U-Net with residual encoder, multipath Inception-ResNet, and lightweight decoder for efficient maxilla segmentation. \\
\midrule

\cite{zhang20213d} & 89 CBCT \ding{55} & \makecell{TMJ}& CNN & \makecell{BCE} & SGD & \makecell{RTX- \\ 2080 Ti} & \makecell{DSC: 98.14±0.54\% \\ASD: 0.0555±0.0198mm\\HD: 1.5711±1.0252mm} & Two-stage multi-view 3D U-Net fuses transverse, sagittal, coronal, and original features for final segmentation. \\
\midrule

\cite{huang2021combining} & 40 CBCT \ding{55}&  \makecell{Maxillofacial}& CNN & / & / & / & \makecell{Accuracy\\ Lesion: 63\% \\Bone: 89\%\\Teeth: 80.1\%} & Semi-supervised 3D U-Net integrates anatomical knowledge constraints to enhance multi-label CBCT segmentation. \\
\midrule

\cite{karacan2022deep} & 1000 PAN \ding{51}  & \makecell{Tooth\\Maxillofacial} & \makecell{CNN\\VIT} &/ & Adam & \makecell{Tesla- \\ P100} & \makecell{DSC(Maxillofacial): \\ Segmenter: 98.12\% \\ConvNeXt: 95.83\%\\U-Net: 91.50\%\\ViT: 87.33\%} & Deep learning models segment teeth and mandible using ViT, Segmenter, ConvNeXt with mask supervision. \\
\midrule

\cite{lang2022localization} * & 50 CBCT \ding{55}  & \makecell{LLC}& CNN & \makecell{CE\\L1} & SGD & \makecell{GTX- \\ 1080 Ti} & \makecell{MSE:\\ 1.38 ± 0.95 mm} & Introduces a Multi-RPN module combined with Mask R-CNN to achieve accurate craniofacial landmark localization. \\
\midrule

\cite{du2022mandibular} & 20 CBCT \ding{55}  & \makecell{MC}& CNN & \makecell{Weighted-\\ BCE} & Adam & \makecell{RTX- \\ 2080} & \makecell{DSC: 85.91\% \\HD95: 0.5371mm} & Attention-3D U-Net segments mandibular canal using scSE channel/spatial attention, fixed-point ROI cropping, and weighted BCE. \\
\midrule

\cite{usman2022dual} & 1357 CBCT \ding{51}  & \makecell{MC}& CNN & \makecell{Dice} & Adam & \makecell{RTX- \\ Titan } & \makecell{DSC: 74.8\% \\Specificity: 99\%\\F1-score: 74.9\%} & Two-stage deep learning framework segments mandibular canal using VOI cropping, 3D attention UNet, and multi-scale residual UNet. \\
\midrule

\cite{jarnstedt2022comparison} & 1103 CBCT \ding{55}  & \makecell{MC} & CNN & / & / & / & \makecell{SMCD: 0.81mm} & 3D U-Net FCN segments mandibular canal, refined via skeletonization and heuristic path selection. \\
\midrule

\cite{zou2022interactive} & 30 CBCT \ding{55}  & \makecell{Hard Palate\\ in the \\Oral Cavity}& CNN & \makecell{BCE\\Dice} & Adam & \makecell{RTX- \\ 2080 Ti} & \makecell{mDSC: 76.44\% \\ mHD: 10.63mm} & Dual-branch Res2Net with channel interaction and PSA enhances palatal edges, multi-step decoder refines segmentation. \\
\midrule

\cite{amorim2023mandible} & 476 CBCT \ding{55}  & \makecell{Mandible} & CNN & \makecell{Dice} & Adam & \makecell{RTX- \\ 6000} & \makecell{DSC: 96±0.17\% \\ HD: 0.49±0.28mm} & 3D U-Net-based mandibular segmentation using internal CT/CBCT dataset. \\
\midrule

MPUNet \cite{xu2023deep} & 13 CBCT \ding{55}  & \makecell{Maxilla\\Mandible\\Tooth} & CNN & \makecell{CE} & Adam & \makecell{RTX- \\ 3090} & \makecell{All:\\DSC: 95.14±1.21\% \\HD: 7.098±1.48mm} & MPUNet-based 2D-view segmentation with pretraining, weighted distance loss, and watershed for tooth separation. \\
\midrule

DBA-UNet \cite{zhang2023dba} & 538 CBCT \ding{55}  & \makecell{MS}& CNN & \makecell{BCE\\Dice} & RMSprop & / & \makecell{DSC: 97.48\% \\Specificity: 99.87\%\\HD95: 1.929mm} & DBA-UNet: dual-stage network with boundary cross-fusion and axial-attention boundary segmentation. \\
\midrule

\cite{widiasri2023alveolar} & 563 CBCT \ding{55}  & \makecell{AB\\MC}& CNN & \makecell{BCE} & Adam & / & \makecell{mIoU:\\ AB: 98\% \\MC: 81\%} & U-Net for 2D CBCT AB and MC segmentation with encoder-decoder and skip connections. \\
\midrule

X-Match \cite{weng2024x} & 13,200 CBCT \ding{55}  & \makecell{Maxilla\\Mandible} & CNN & / & /& / & \makecell{mIoU(All):\\ ResNet-50: 86.2\% \\ResNet-101
: 87.1\%} & X-Match semi-supervised segmentation leverages wavelet transform to generate multi-scale augmentations and fuses labeled with unlabeled data via a label fusion module. \\
\midrule

\cite{han2024weakly} & 443 CBCT \ding{51}  & \makecell{IAN} & CNN & \makecell{IOU} & SGD & / & \makecell{DSC: 73.54±10.69\% \\HD95: 7.85±8.22} & Three-stage IAN segmentation: weakly supervised label generation, manual refinement, followed by AttentionUnet3D supervised training. \\
\midrule

\cite{mai2024multi} & / CBCT \ding{51} & \makecell{Maxillofacial}& Hybrid  & \makecell{CE\\Dice} & / & \makecell{RTX- \\ 4090D} &/ & Two-stage CBCT segmentation: multi-view 2D SwinUnet extracts features, improved 3D Unet classifies fused results. \\
\midrule

DMVEL-Net \cite{hu2025temporomandibular} & 88 CBCT \ding{55}  & \makecell{TMJ} & CNN & \makecell{BCE} & SGD & \makecell{RTX- \\ 2080 Ti} & \makecell{DSC: 98.18±0.51\% \\ ASD: 0.0554±0.0182mm\\HD95: 0.1743±0.0550mm} & DMVEL-Net integrates coronal and sagittal CBCT features and enforces symmetry for TMJ segmentation. \\
\midrule

\cite{bolelli2024segmenting} & 493 CBCT \ding{51}  & \makecell{MC} & CNN & \makecell{CE,Dice\\FD,SD\\Focal\\cl-Dice \\Modifild HD} & \makecell{Adam\\AdamW} & \makecell{Tesla- \\ T4} & \makecell{1st.: \\ DSC: 79.6\% \\ HD95: 4.49mm} & The results of a medical image segmentation challenge.These methods primarily leveraged the nnU-Net,ResUNet framework and enhanced performance through model ensembling,advanced data augmentation, and specialized training strategies for precise segmentation. \\
\midrule

\cite{chen2025convolutional} & 300 CBCT \ding{55}  & \makecell{MS} & CNN & / & RMSprop & \makecell{RTX-\\ 3060} & \makecell{IoU\\ Axial plane :94.2\% \\ Sagittal plane: 93.7\%\\Coronal plane: 91.6\%} & U-Net performs automatic segmentation of maxillary sinus on 2D slices, with encoder feature extraction and decoder upsampling and concatenation. \\

\bottomrule
\end{tabular}
\renewcommand{\arraystretch}{1}
\end{table*} 
\clearpage
\noindent while the second stage, a boundary attention network with a dual-input single-output structure, enables precise segmentation of the maxillary sinus region in CBCT scans. Wen et al. \cite{weng2024x} introduced X-Match, a semi-supervised framework for maxillofacial bone segmentation in CBCT images, which applies wavelet transform to decompose images into low- and high-frequency components and leverages consistency training to enhance multi-scale feature extraction. Hu et al. \cite{hu2025temporomandibular} developed DMVEL-Net, a multi-view ensemble learning network composed of an encoder-decoder backbone and multi-view feature fusion branches, which takes coronal and sagittal CBCT slices as inputs and enhances spatial representations through feature fusion, achieving accurate segmentation of the temporomandibular joint.

\subsection{Classification}
Dental image classification tasks can generally be divided into two major directions: the identification of tooth types and conditions, and the recognition of dental diseases. DL models, through end-to-end feature extraction, enable pixel- or region-level dense predictions that provide additional structural features and well-defined ROIs, thereby enhancing the accuracy and robustness of classification performance.
\FloatBarrier

\begin{figure}[t]
	\centerline{\includegraphics[width=\linewidth]{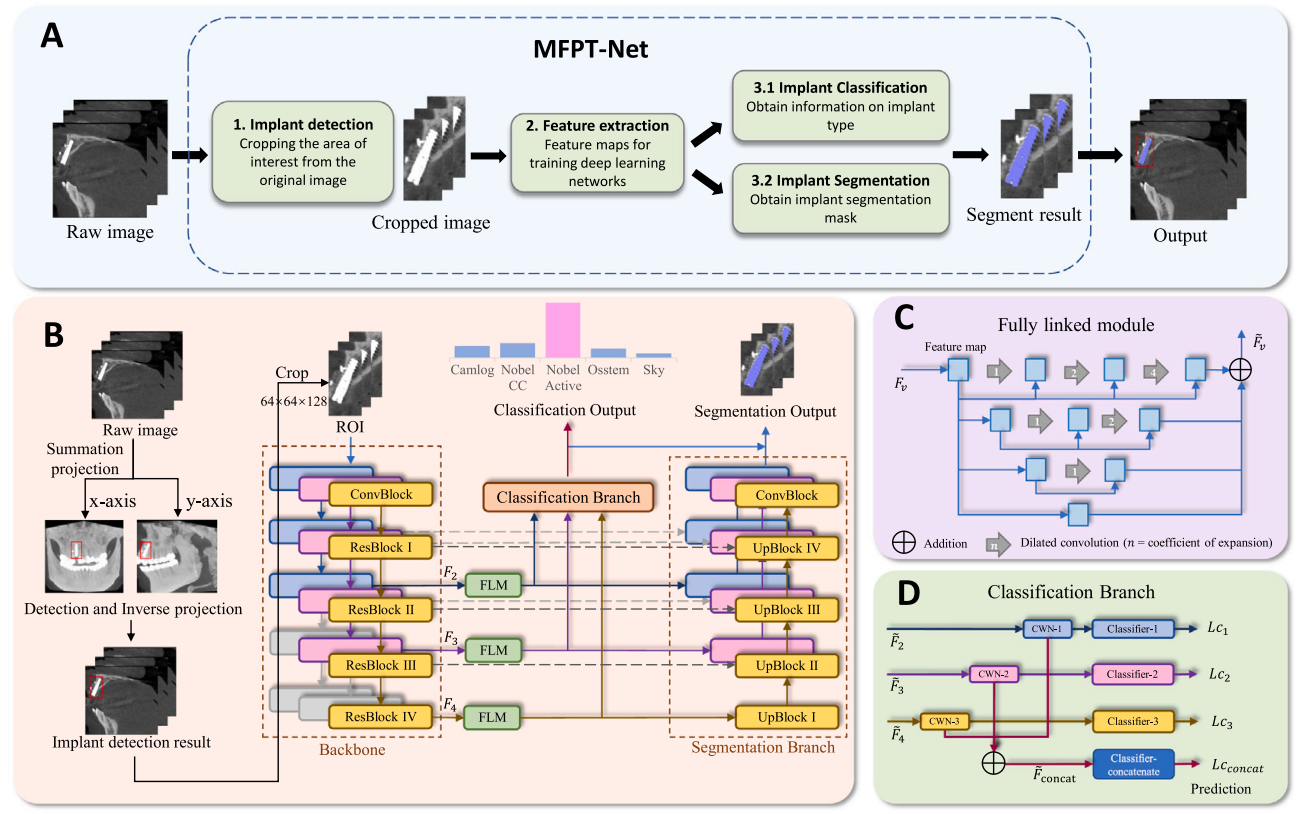}}
	\caption{Overall architecture of MFPT-Net for implant classification and segmentation in CBCT images. Built on ResNet10, it employs multi-scale progressive training with Fully Linked Module and Channel-Wise Normalization modules to enhance global and channel features for robust multi-task performance. Image sourced from \cite{zhao2025progressive}.}
	\label{fig21}
\end{figure}

\subsubsection{Tooth}
For tooth classification tasks, dense prediction results are often employed either as foundational features or as intermediate steps to construct more intelligent and accurate automated diagnostic systems. Awari et al. \cite{awari2024three} proposed a tooth image segmentation and classification framework, where a CNN was first utilized for preliminary feature extraction to obtain tooth ROIs, followed by an enhanced CNN for refining the segmentation results. Subsequently, a multilayer convolutional network was applied for tooth classification, achieving precise identification on IOS data. Alam et al. \cite{alam2025using} aimed to segment impacted molars from PAN images by integrating classification models based on CNNs, ViT, CLIP \cite{radfordLearningTransferableVisual}, and hybrid architectures at the input stage to determine the presence of impacted wisdom teeth. If the classifier yielded a positive result, the image was subsequently passed to a UNet for segmentation. As shown in Fig. \ref{fig21}, Zhao et al. \cite{zhao2025progressive} introduced MFPT-Net for implant classification and segmentation in CBCT images. After feature extraction by the backbone, a Fully Linked Module was employed to globally enhance multi-scale representations, while channel-wise normalization in the classification branch improved feature expressiveness, enabling efficient classification of implant types. Meine et

\clearpage
\begin{table*}[h]
\centering
\tiny
\setlength{\tabcolsep}{5.5pt}
\caption{A summary of tooth and dental disease studies, arranged in ascending order of publication year. Asterisks (*) indicate that the study involves a detection task, while the symbol (※) denotes that the loss function was proposed by the authors. In the table, \ding{55} represents the use of a private dataset, and \ding{51} represents the use of a public dataset. When multiple dental lesions are involved, “Multiple Abnormal Conditions” is used to describe the targets. For studies employing more than three evaluation metrics, the three most representative metrics are selected. OC (Odontogenic Cysts); FI (Furcation Involvement); MCC (Matthews Correlation Coefficient); MA (Macro-Accuracy); CR (Classification Rate); AUROC (Area Under the Receiver Operating Characteristic Curve).}
\label{tab7}
\renewcommand{\arraystretch}{0.6}
\begin{tabular}{%
m{1.5cm} 
c 
c 
c 
c 
c 
m{6.3cm} 
}

\toprule
Paper & Dataset & Targets & Methods & \multicolumn{1}{c}{\makecell{Loss \& \\ Optimizer}} & Evaluation &\multicolumn{1}{c}{Highlights}\\
\midrule

\cite{tian2019automatic} & 600 IOS \ding{55} &Tooth & CNN & \makecell{CE \\ \& SGD} & \makecell{MA: \\ Dataset1: 91.44\%\\Dataset2: 95.43\%\\Dataset3: 95.96\%} & A 3D CNN-based hierarchical classification and segmentation framework integrates sparse octree preprocessing and CRF refinement for automatic tooth analysis. \\
\midrule

\cite{lakshmi2020classification} & 1900 PAN \ding{51}  & \makecell{Caries} & CNN & / & \makecell{Accuracy: 96.08\% \\ Precision: 78.50\%\\Recall: 84.20\%} & Employ histogram equalization and Sobel edge detection, followed by an AlexNet-based deep CNN for early caries detection in dental X-rays. \\
\midrule

\cite{zhang2021image} & 1000 IP \ding{55}  & \makecell{Caries} & CNN & / & \makecell{Accuracy: 97\%} & Transfer learning via ImageNet pre-training and dental dataset fine-tuning enhances feature extraction. \\
\midrule

\cite{sivasundaram2021performance} & 1171 PAN \ding{55}  & \makecell{OC} & CNN & / & \makecell{CR: 96.51\% \\ Precision: 96.01\%\\F-score: 96\%} & The modified LeNet-CNN performs four-class oral cyst classification and morphology-based segmentation of abnormal regions. \\
\midrule

\cite{kumari2022design} & 120 PR \ding{51}  & \makecell{Caries}  & \makecell{CNN \\ RNN} & / & \makecell{Accuracy: 93.67\% \\ Specificity: 92.73\%\\F1-score :93.54\%} & The HSLnSSO algorithm optimizes FOC-KKC segmentation and ResNeXt-RNN classification parameters, improving model adaptability and convergence. \\
\midrule

\cite{li2023transformer} & 151 CBCT \ding{55}& \makecell{Pulp\\Calcification} & ViT & \makecell{CE + BCE\\Dice\\※Triple\\ \& Adam}  & \makecell{Accuracy: 78.47±0.25\% \\Precision: 77.31±0.37\%\\Recall: 82.08±0.23\%} & A coarse-to-fine CBCT tooth segmentation method is proposed, using a Swin Transformer and triple loss to improve segmentation, identification, and calcification detection. \\
\midrule

PaXNet \cite{haghanifar2023paxnet} & \makecell{350 PAN \ding{51}\\120 PAN\ding{55}}& \makecell{Caries} & CNN & / & \makecell{Accuracy: 86.05\%  \\ Precision: 89.41\% \\Recall: 50.67\%} & About 400 X-ray images were used for tooth segmentation, ROI extraction, and caries classification. \\
\midrule

\cite{sarvesan2024enhancing} & PR,BR,PAN \ding{55}  & \makecell{Periapical \\ Lesions} & CNN & / & {Accuracy: 97\% } & A dual-branch model with feature fusion, transfer learning, and partial fine-tuning is used for periodontal classification. \\
\midrule

3DDISC-DLTSA \cite{awari2024three} & 1400 CBCT \ding{55} &Tooth& CNN & / & \makecell{Accuracy: \\ D1: 96.47\%\\D2: 97.55\%\\D3: 97.80\%} & The method uses U-Net for segmentation, DenseNet-169 for feature extraction, and a Tunicate Swarm-optimized Extreme Learning Machine for classification.\\
\midrule

\cite{hameed2024novel} & 100 PAN \ding{51} &Tooth & CNN &/ & \makecell{Accuracy: 97.14\%\\Precision: 97.00\%\\Recall: 97.20\%}& First use CM-CNN for 3D tooth segmentation, then AEG for classification, with FPN, ROI Align, and Swish/DropBlock to enhance performance. \\
\midrule

MFPT-Net \cite{zhao2025progressive} & \makecell{537 CBCT\ding{55}} &\makecell{Dental-\\Implant}& CNN & \makecell{CE\\Dice} & \makecell{Accuracy: 92.98\%\\mPrecision: 93.15\%\\mRecall: 93.31\%} & MFPT-Net uses ResNet10 for implant classification and segmentation with multi-scale features and progressive training, while FLM enhances features and CWN normalizes channels. \\
\midrule

\cite{fang2025enhancing} &3005 PAN\ding{55} &\makecell{Wisdom\\Teeth} & CNN & / & \makecell{Accuracy: 82.46\%\\F1-score: 77.49\%\\AUROC: 88.66\%} & Left/right CNNs for image halves and a Full Image CNN for multi-scale features, then fuses them for multi-label wisdom tooth classification. \\
\midrule

TEANet \cite{li2025teanet} & 304 IOS \ding{55} &Tooth & CNN & \makecell{L1 \\ \& Adam} & \makecell{Accuracy: 98.41\% } & PointNet extracts features, a node-erasable graph models tooth adjacency, and Graph STN performs 3D alignment. \\
\midrule

\cite{alam2025using} & 693 PAN \ding{55} &Tooth & \makecell{CNN\\VIT\\CLIP\\Hybrid}& \makecell{CE\\Dice \\ \&Adam}  & \makecell{F1-score: \\1st.:51\% \\2nd: 43\%\\3rd: 38\%} & The study uses CNN, ViT, and CLIP for tooth classification, U-Net for segmentation, with feature fusion, data augmentation, and regularization to improve performance. \\
\midrule

\cite{sehar2025automatic} & 105 CBCT\ding{51} &Tooth& CNN & \makecell{CE \\ \& Adam }  & \makecell{Accuracy: 96.40\%\\Precision: 96.75\%\\Recall: 94.92\%} & The method segments teeth using thresholding and 3D U-Net, reconstructs point clouds, extracts features via PointNet and HamNet, and classifies with a prototype network. \\
\midrule

\cite{meine2025determination} & 209 PAN \ding{55} &Tooth& CNN & \makecell{CE\\SL1\\BCE\\ \& SGD}& \makecell{Precision: 98.1\%\\Recall: 97.1\%\\Specificity: 96.2 \% } & Mask R-CNN segments teeth with bounding boxes, masks, and FDI labels, filtering predictions by confidence and determining jaw type via Kennedy’s classification. \\
\midrule

\cite{kibcak2025deep} & 7696 PAN\ding{51}  & \makecell{Peri- \\ implantitis} & CNN & Adam & \makecell{Precision: 77.7\%\\Recall: 90.3\%\\DSC: 98.6\%\\} & Implants are segmented with U-Net using skip connections, then classified with AlexNet to detect peri-implantitis. \\
\midrule

\cite{zhang2025enhancing} &\makecell{1089 PAN\ding{55}\\506 CBCT\ding{55}}& \makecell{FI} & \makecell{CNN \\ ViT} & \makecell{Adam} & \makecell{F1-score(Both): \\MLP: 84±2.0\%\\ GoogLeNet: 91±2.4\% \\VGGNet: 93±2.5\% \\ ViT: 95±2.3\%} & The method uses six-fold stratified cross-validation; VGGNet extracts fine-grained features, GoogleNet captures multi-scale patterns, and ViT models global dependencies with multi-head self-attention. \\
\midrule

\cite{silva2025holistic} * &\makecell{16824 PAN\ding{55}}& \makecell{Multiple\\Abnormal\\Conditions} & \makecell{CNN \\ ViT} & \makecell{BCE\\MCC\\ \& AdamW}  & \makecell{MCC: 47.5\%} & HTC generates tooth pseudo-labels, ViT with MAE classifies lesions, GPT-4 extracts lesion names, and MCC evaluates performance. \\
\midrule

\cite{arslan2025tooth} * & 2332 PAN \ding{51}&\makecell{Caries\\Periapical Lesions\\Impacted teeth} & \makecell{CNN\\VIT\\Hybrid} & / & \makecell{F1-score: \\ DINOV2: 79.00\%\\CNN: 66.19\%\\EfficientNet: 69.33\%} & The method integrates TransUNet, YOLO, SAM, GAN, and DINO for tooth segmentation, detection, and classification. \\
\midrule

\cite{li2025dense} & 358 CBCT \ding{55} & \makecell{Multiple \\Abnormal\\Conditions} & Hybrid & \makecell{CE \\ Dice \\ Focal \\ \& SGD}  & \makecell{ Accuracy: 89.23\% \\ Precision: 90.13\%\\ Recall: 88.18\%} & A unified network combines 3D U-Net segmentation and Transformer classification, using attention to fuse mask and image features. \\
\midrule

HC-Net \cite{mei2025clinical} & 481 PAN \ding{55}  & \makecell{Periapical \\ Lesion} & CNN & \makecell{Focal,L1\\ CE,Dice\\ BCE,MSE \\ \& Adam}  & \makecell{Accuracy: 92.7±1.3\% \\ Precision: 90.2±1.4\%\\ Recall: 95.8±2.1\%} & Hybrid framework combines tooth-level segmentation and patient-level prediction with clinical-guided fusion for periodontal diagnosis. \\
\midrule

TWGS \cite{pang2025establishment} & 388 IP \ding{55}  & \makecell{Tooth\\Wear\\Degree} & \makecell{CNN\\VIT} & \makecell{/} & \makecell{Precision: 91\% \\ Recall: 88\%\\mPA : 95\%} & TWGS: a Mask R-CNN-based model for single-tooth segmentation, followed by a ViT for wear severity classification on individual tooth surfaces. \\
\midrule

\cite{nasra2025deep} & 11,653 IP \ding{55} & \makecell{Multiple\\Abnormal\\Conditions} & CNN & Adam & \makecell{Accuracy: 93\%} & InceptionResNetV2 with customized linear head extracts multi-scale features for accurate oral disease classification. \\

\bottomrule

\end{tabular}
\renewcommand{\arraystretch}{1}
\end{table*} 
\clearpage
\noindent al. \cite{meine2025determination} utilized Mask R-CNN for tooth instance segmentation in PAN images as a prerequisite step, and subsequently implemented an automatic five-class classification of partially edentulous jaws based on confusion matrix analysis.

\subsubsection{Dental Disease}
\begin{figure}[t]
	\centerline{\includegraphics[width=\linewidth]{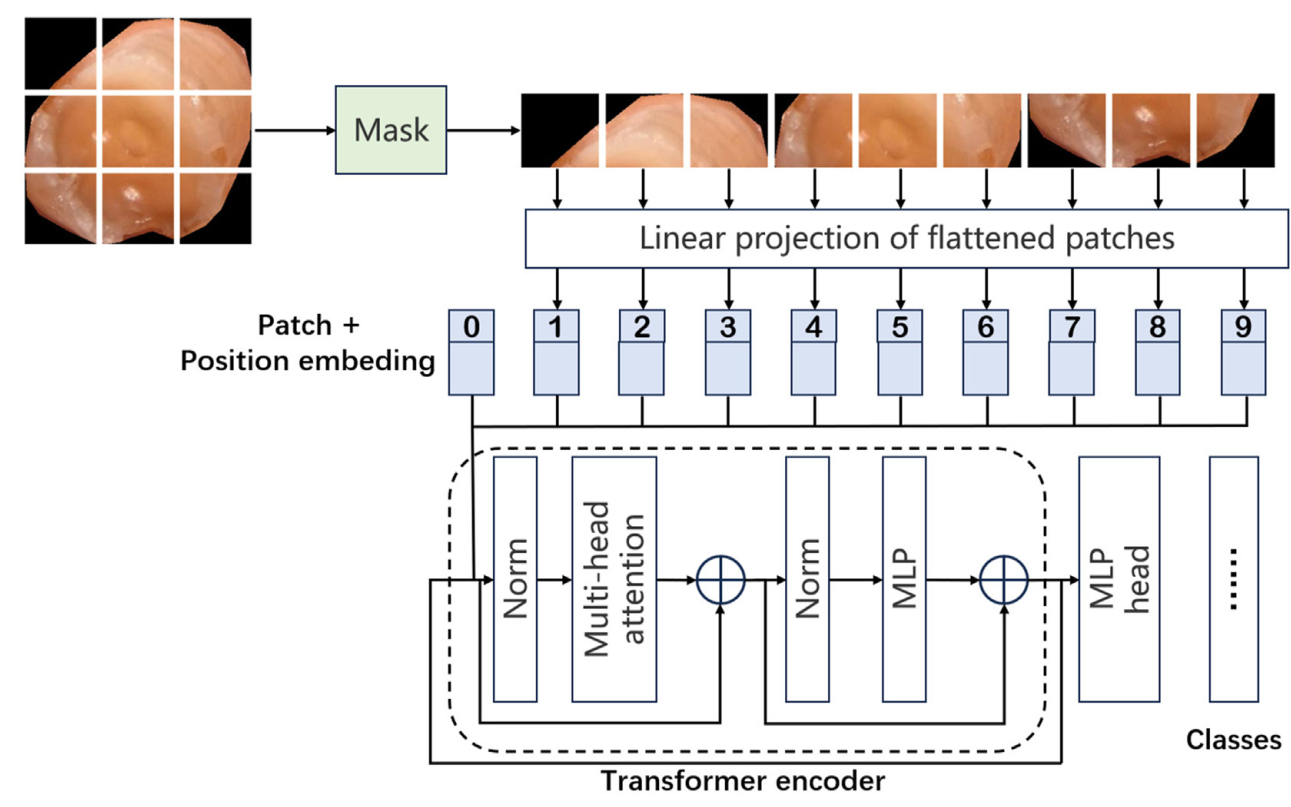}}
	\caption{The architecture based on ViT is designed to classify the degree of tooth wear on different surfaces (occlusal/incisal vs. non-occlusal/non-incisal) of individual teeth in IP images. Image sourced from \cite{pang2025establishment}.}
	\label{fig22}
\end{figure}

Classification of dental lesions is a crucial component of computer-aided diagnosis in dentistry. Such systems typically rely on dense detection results to identify pathological or infectious regions in the teeth and maxillofacial area, providing clinicians and patients with intuitive risk alerts. Li et al. \cite{li2023transformer} developed a ViT-based multi-task learning framework with a contrastive triple-loss mechanism to enhance the weak distinction between normal and calcified teeth, enabling both tooth segmentation and pulp calcification classification. Haghanifar et al. \cite{haghanifar2023paxnet} proposed a caries classification framework that first extracts ROIs for each tooth from PAN images, followed by a CNN-based binary classification to determine caries presence. Sarvesan et al. \cite{sarvesan2024enhancing} designed a dual-branch hybrid model combining pretrained VGG16 and MobileNet backbones to classify periodontal diseases from dental X-ray images. Mei et al. \cite{mei2025clinical} employed a backbone network to perform instance segmentation for detecting tooth centroids, bounding boxes, and masks, then used a pretrained encoder and classification head to achieve periodontal disease grading. As shown in Fig. \ref{fig22}, Pang et al. \cite{pang2025establishment}
\noexpand developed the TWGS system for tooth wear grading based on IP images, using Mask R-CNN for tooth instance segmentation and ViT for classifying abrasion levels across different tooth surfaces. Silva et al. \cite{silva2025holistic} proposed a multimodal dental lesion classification framework integrating PAN images and textual reports, where GPT-4 automatically extracts tooth-related lesion noun phrases from reports, and multiple ViT-based binary classifiers are trained to identify specific dental pathologies.

\subsection{Others}
In addition to the aforementioned tasks, this study also identified several derivative tasks that build upon dense prediction and classification as fundamental components, extending their applications to fields such as biometric identification, registration, planning, and reconstruction. For example, in the field of dental biometric identification, Atacs et al. \cite{atacs2022biometric} analyzed PAN images from 372 patients collected between 2018 and 2020, with each patient having at least two scans taken at different times. They proposed PDR-Net, a model combining convolutional and fully connected layers for feature extraction and similarity measurement to match PAN images from the same patient. Similarly, Lai et al. \cite{lai2020lcanet} utilized clinical PAN images captured before and after orthodontic treatments over the past two decades. Their model employed a shallow CNN backbone integrated with a channel attention module and a learnable connected module to enhance target tooth region representations and automatically learn cross-layer connection weights, achieving identity recognition via cosine similarity.

Beyond identification, DL has also been leveraged for registration, surgical planning, and 3D reconstruction in dental imaging. Jang et al. \cite{jang2024fully} proposed TSIM, an automated registration and fusion framework for IOS and CBCT. TSIM-IOS and TSIM-CBCT respectively perform instance segmentation and labeling for each modality, followed by a stitching error correction module that incrementally aligns IOS data with CBCT references for accurate multimodal registration. Xing et al. \cite{xing2024dental} developed an end-to-end framework for tooth segmentation and implant planning directly from PAN images. Using ResUNet for segmentation, they applied edge detection for contour extraction and utilized principal component analysis and linear regression to predict optimal implant locations. Tan et al. \cite{tan2023coupling} introduced BSegNet, a graph-based network for bracket segmentation on 3D dental models, which further supports surface reconstruction. Their results demonstrate superior accuracy and efficiency compared with interactive reconstruction workflows, showing strong potential for clinical adoption.

\begin{figure}[b]
	\centerline{\includegraphics[width=\linewidth]{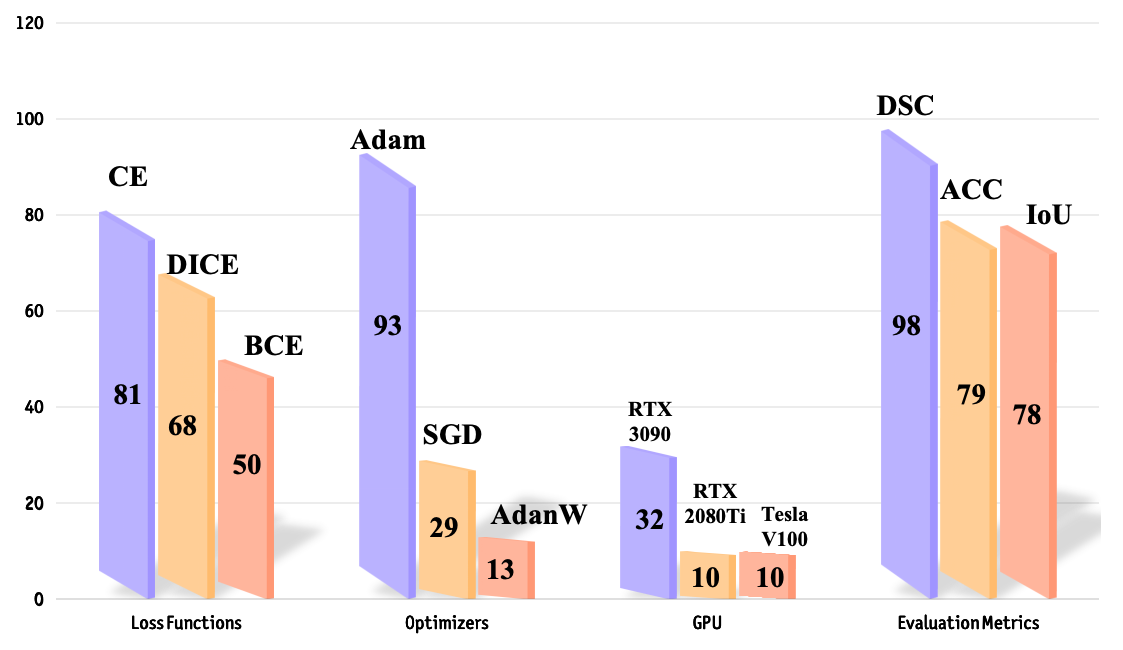}}
	\caption{The figure presents statistics on the top three most frequently used loss functions, optimizers, GPUs, and evaluation metrics reported in the reviewed studies. The numbers indicate the frequency of occurrence for each item.}
	\label{fig23}
\end{figure}

\section{Training and Evaluation}
\label{sec4}
This section summarizes the key components involved in model training from three perspectives: loss functions, optimizers, and GPU hardware configurations. Furthermore, it categorizes and introduces the commonly used evaluation metrics in the DIA field from two dimensions: confusion-matrix-based and set-distance-based measures.

\subsection{Loss Functions}
In the training phase, the loss function serves as a crucial supervision strategy to measure the discrepancy between the model’s output and the ground truth (GT). By minimizing the loss value, the model is encouraged to align its predictions with the true labels, thereby improving accuracy. In the field of DIA, as shown in Fig. \ref{fig23}, Cross-Entropy (CE) loss and Dice loss \cite{milletariVNetFullyConvolutional2016} are the two most commonly used loss functions.

\subsubsection{\textbf{L2 Loss}}
L2 loss, also known as Mean Squared Error (MSE), serves to quantify the discrepancy between the model’s predictions and the GT. It calculates the mean of the squared differences, thus penalizing larger prediction errors more heavily, as defined in Equation \ref{eq15}.
\begin{equation}
\label{eq15}
\mathcal{L}_{\mathrm{MSE}} = \frac{1}{N} \sum_{i=1}^{N} \left( y_i - \hat{y}_i \right)^2
\end{equation}
Here, $N$ denotes the number of training samples, $y_i$ represents the GT of the $i$-th sample, and $\hat{y}_i$ denotes the model’s prediction for the $i$-th sample. DDue to the squaring operation, MSE is a continuous and differentiable function, which facilitates gradient-based optimization. However, this squaring property can amplify errors when $|y_i - \hat{y}_i|>1$. Consequently, outlier samples can exert a disproportionate influence on model parameter tuning, causing the model to overfit these anomalies while underfitting the majority of normal samples.

\subsubsection{\textbf{L1 Loss}} 
L1 Loss is the sum of the absolute differences between the GT and the model’s predictions, mathematically defined as in Equation. \ref{eq16}:
\begin{equation}
\label{eq16}
\mathcal{L}_{\mathrm{L1}} = \sum_{i=1}^{N} \left| y_i - \hat{y}_i \right|
\end{equation}
The gradient of L1 loss is constant and non-differentiable at zero, which results in smaller model updates when errors are large and may require numerical approximations near zero, slowing down convergence. Unlike squared error loss, L1 loss does not amplify the discrepancies caused by outliers, making it more robust and suitable for regression problems with anomalous data.

\subsubsection{\textbf{Smooth L1 Loss}} 
Smooth L1 Loss combines the advantages of L1 and L2 losses, providing a balance between sensitivity to outliers and convergence stability. Its mathematical definition is shown in Equation. \ref{eq17} and \ref{eq18}.

\begin{equation}
\label{eq17}
\mathcal{L}_{\mathrm{SL1}} = \sum_{i=1}^{N} \ell_{\mathrm{SL1}}\left( y_i - \hat{y}_i \right)
\end{equation}

\begin{equation}
\label{eq18}
\ell_{\mathrm{SL1}}(y_i - \hat{y}_i) =
\begin{cases} 
0.5 ( y_i - \hat{y}_i )^2, & \text{if } |y_i - \hat{y}_i| < 1 \\
|y_i - \hat{y}_i| - 0.5, & \text{otherwise}
\end{cases}
\end{equation}
It can be observed that when the error is small, the Smooth L1 Loss offers a stable gradient in a quadratic form. Conversely, when the error is large, it transitions to a linear form, enhancing the model's robustness and mitigating the influence of outliers on fitting. This property ensures precise boundary fitting for small targets while preventing gradient explosion caused by large prediction errors.
\subsubsection{\textbf{Cross Entropy Loss}}
The Cross-Entropy (CE) Loss quantitatively measures the discrepancy between the model's predicted probability distribution and the true label distribution. It is the most widely used loss function in DIA. Its mathematical definition is given in Equation. \ref{eq19}:

\begin{equation}
\label{eq19}
\mathcal{L}_{\mathrm{CE}} = - \frac{1}{N} \sum_{j=1}^{N} \sum_{i=1}^{C} y_{j,i} \log \hat{y}_{j,i}
\end{equation}
Here, $N$ denotes the total number of samples, $C$ denotes the total number of classes, $y_{j,i}$ represents the true distribution for the $i$-th class of the $j$-th sample, and $\hat{y}_{j,i}$ denotes the predicted probability for the $i$-th class of the $j$-th sample,The softmax function is typically used to map mutually exclusive classes into a probability distribution summing to 1. A smaller cross-entropy value indicates that the predicted distribution $\hat{y}_{j,i}$ is closer to the true distribution $y_{j,i}$.

For multi-target dental image detection, each sample may contain various types of teeth, such as incisors, molars, and wisdom teeth. The CE loss measures the discrepancy between the predicted probability $\hat{y}_{j,i}$ and the GT $y_{j,i}$ for each tooth class. Through gradient optimization, CE loss enables the model to distinguish between different classes, thereby achieving precise tooth detection.

The Binary Cross Entropy (BCE) loss is a specialized form of the CE loss function, tailored for binary classification tasks. Its mathematical definition is as the follow Equation. \ref{eq20}:
\begin{equation}
\label{eq20}
\mathcal{L}_{\mathrm{BCE}} = - \frac{1}{N} \sum_{i=1}^{N} \left[ y_i \log \hat{y}_i + (1 - y_i) \log (1 - \hat{y}_i) \right]
\end{equation}
Here, $N$ denotes the total number of samples, $y_i$ represents the GT for the $i$-th sample, and $\hat{y}_i$ represents the model’s predicted probability of the positive class. The BCE loss considers both positive and negative predictions, effectively measuring the model’s overall performance in binary classification tasks.

\subsubsection{\textbf{Dice Loss}}
The Dice Loss \cite{milletariVNetFullyConvolutional2016} is a commonly used loss function in segmentation tasks, derived from the Dice Similarity Coefficient (DSC), which measures the overlap between the predicted and GT segmentation regions, with values ranging from 
$[0,1]$. By incorporating the loss minimization concept into this metric, the Dice Loss is defined in Equation. \ref{eq21}:
\begin{equation}
\label{eq21}
\mathcal{L}_{\mathrm{Dice}} = 1 - \frac{2 \sum_{i=1}^{N} p_i g_i + \epsilon}{\sum_{i=1}^{N} p_i + \sum_{i=1}^{N} g_i + \epsilon}
\end{equation}
Here, $p_i$ denotes the predicted probability of the $i$-th pixel, $g_i$ represents the corresponding GT, $N$ is the total number of pixels, and $\varepsilon$ is a small constant added for numerical stability. Dice Loss directly optimizes the overlap between prediction and GT, effectively addressing class imbalance — a particularly important issue in medical image segmentation.

\subsubsection{\textbf{IoU Loss}}
Similar to the design principle of the Dice Loss, the IOU Loss is derived from a commonly used evaluation metric in segmentation tasks—Intersection over Union (IoU). IoU measures the overlap between the predicted segmentation and the ground truth, defined as the ratio of the intersection area to the union area of the predicted and true regions. The IOU Loss is formulated to directly optimize this overlap, thereby improving segmentation accuracy. Its mathematical definition is given in Equation. \ref{eq22}:

\begin{equation}
\label{eq22}
\mathcal{L}_{\mathrm{IOU}} = 1 - \frac{\sum_{i=1}^{N} p_i g_i + \epsilon}{\sum_{i=1}^{N} \left(p_i + g_i - p_i g_i\right) + \epsilon}
\end{equation}
The IOU Loss, by penalizing the non-overlapping regions between the predicted segmentation and the GT, enhances the model's segmentation accuracy, especially along the object boundaries. This focus on boundary accuracy makes it particularly effective in improving the precision of segmentation tasks.

\subsection{Optimizer and GPU}
Optimizers and GPUs are two indispensable core components in the training of deep learning models. As shown in Fig. \ref{fig23}, we summarized the optimizers and GPU configurations used in all reviewed studies. The results indicate that Adam \cite{kingma2015adam} remains the most widely adopted optimizer in the DIA field. Its popularity stems from its combination of the advantages of momentum and adaptive learning rate, enabling faster convergence and greater stability in complex medical imaging tasks. Compared with the traditional SGD \cite{robbins1951stochastic} optimizer, Adam demonstrates superior robustness and generalization when dealing with challenges commonly found in dental imaging, such as small sample sizes, high noise levels, and class imbalance. Furthermore, since DIA tasks often involve computationally intensive structures such as 3D convolutions and attention mechanisms, the adaptive nature of Adam in parameter updates helps alleviate issues of gradient explosion or vanishing.

In terms of GPU usage, the NVIDIA RTX 3090 is currently the most commonly adopted graphics card model in the DIA field. This preference is primarily attributed to its large memory capacity (24 GB) and favorable cost-performance ratio, which effectively support batch training of high-resolution dental images and the computational demands of 3D convolutional models. In contrast, high-end GPUs such as the A100 are less frequently used in DIA research, a trend that differs somewhat from other branches of medical imaging. The fundamental reason lies in the fact that DIA still lacks sufficiently large and standardized benchmark datasets; consequently, the data volume and parameter scale of model training remain relatively modest, and the demand for ultra-high computing power is not yet pressing. Moreover, many research teams conduct experiments within clinical institutions, where hardware budgets and deployment conditions are often constrained, leading them to prefer mid-to-high-end consumer-grade GPUs as their primary computing platforms.

\subsection{Evaluation Metric‌}
\subsubsection{Metrics Based on the Confusion Matrix‌}
\paragraph{\textbf{Intersection over Union (IoU)}} also known as the Jaccard coefficient, is a commonly used evaluation metric in image segmentation tasks. It measures the degree of overlap between the predicted result and the GT. IoU is defined as the ratio of the intersection area to the union area of the predicted and true segmentation regions, providing a quantitative assessment of segmentation accuracy. Its definition is shown in Equation. \ref{eq1}. 

\begin{equation}
\label{eq1}
\text{IoU} = \frac{TP}{TP + FP + FN}
\end{equation}
Here, $TP$ (True Positive) denotes the number of pixels correctly predicted as positive, $FP$ (False Positive) denotes the number of pixels incorrectly predicted as positive, and $FN$ (False Negative) denotes the number of pixels incorrectly predicted as negative. The IoU value ranges from $[0, 1]$, with higher values indicating a greater overlap between the results and the GT. The Mean Intersection over Union (mIoU) is particularly suitable for multi-class segmentation tasks. The basic idea is to compute the IoU for each class individually and then take the average across all classes, as shown in Equation. \ref{eq2}:

\begin{equation}
\label{eq2}
\text{mIoU} = \frac{1}{N} \sum_{i=1}^{N} \frac{TP_i}{TP_i + FP_i + FN_i}
\end{equation}
Here, $N$ represents the number of classes. In DIA, IoU is widely used as an evaluation metric for multi-tooth or multi-structure segmentation tasks. Due to its intuitive geometric interpretation in object detection, IoU is also commonly applied in dental lesion detection and tooth localization tasks.

\paragraph{\textbf{Dice Similarity Coefficient (DSC)}} is one of the most commonly used evaluation metrics in medical image segmentation and is widely applied in DIA. Its definition is given by Equation. \ref{eq3}: 

\begin{equation}
\label{eq3}
\text{DSC} = \frac{2TP}{2TP + FP + FN},
\end{equation}
The Mean Dice Similarity Coefficient (mDSC) extends the DSC to multi-class segmentation, similar to mIoU, and is defined as follow:

\begin{equation}
\label{eq4}
\text{mDice} = \frac{1}{N} \sum_{i=1}^{N} \frac{2TP_i}{2TP_i + FP_i + FN_i}
\end{equation}
Compared with IoU, the DSC is more sensitive to extreme class imbalance, which is common in medical imaging. In dental images, target regions such as teeth or dental pulp usually occupy only a small portion of CBCT or PAN scans; thus, using DSC provides a more reasonable reflection of segmentation performance.

\paragraph{\textbf{Recall (Sensitivity)}} is a commonly used performance metric in classification and segmentation tasks, aimed at evaluating a model’s ability to correctly identify positive samples. Its value ranges from $[0,1]$, with higher values indicating better coverage of target regions. Recall is defined as follow Equation. \ref{eq5}:

\begin{equation}
\label{eq5}
\text{Recall} = \frac{TP}{TP + FN}
\end{equation}
In the field of DIA, Recall has been widely employed in disease detection tasks. For instance, in the identification of periodontal disease or caries, a higher Recall can effectively improve lesion detection rates and reduce the risk of missed diagnoses.

\paragraph{\textbf{Precision}} Evaluating model performance solely based on Recall is not comprehensive because excessively high Recall may result in increased false positives. Therefore, it is essential to balance Recall with Precision. Precision measures the proportion of true positive samples among all positive predictions made by the model. The value of Precision ranges from 
$[0,1]$, with higher values indicating greater accuracy in positive predictions. It is defined as follow:

\begin{equation}
\label{eq6}
\text{Precision} = \frac{TP}{TP + FP}
\end{equation}
In the field of DIA, a higher Precision is crucial for reducing the incidence of misdiagnoses. For instance, in caries detection on PAN images, a higher Precision indicates that most of the predicted lesions are truly present, minimizing the excessive inclusion of background or non-pathological areas. This ensures that clinicians receive more reliable information, aiding in more accurate clinical decision-making and improving patient outcomes.

\paragraph{\textbf{Specificity}} is an important metric that measures the model’s ability to correctly identify negative samples, i.e., background or non-target regions. It is defined in Equation. \ref{eq7}:

\begin{equation}
\label{eq7}
\text{Specificity} = \frac{TN}{TN + FP}
\end{equation}
Here, $TN$ (True Negative) denotes the number of pixels that are truly negative and correctly predicted as negative. The value of Specificity ranges from 
$[0,1]$, where a higher value indicates the model’s superior ability to avoid false positives. In medical image analysis, Specificity is often used in conjunction with Recall to provide a comprehensive assessment of diagnostic performance. In the field of DIA, Specificity is critically important. For instance, in CBCT-based segmentation of teeth or alveolar bone, a higher Specificity indicates that the model can effectively prevent background or noise from being misclassified as target structures, thereby enhancing the reliability of segmentation outcomes.

\paragraph{\textbf{F1-score}} is the harmonic mean of Precision and Recall, designed to provide a balanced evaluation between these two metrics. It ranges from $[0,1]$, where a higher F1-score indicates an improved ability of the model to balance Precision and Recall. In medical image analysis, the F1-score is frequently utilized to evaluate the overall accuracy and stability of models in detection and segmentation tasks. It is defined in Equation. \ref{eq8}.

\begin{equation}
\label{eq8}
\text{F1-score} = \frac{2 \cdot \text{Precision} \cdot \text{Recall}}{\text{Precision} + \text{Recall}}
= \frac{2TP}{2TP + FP + FN}
\end{equation}
In DIA, the F1-score is a valuable and intuitive evaluation metric for dental clinical applications. For example, in dental plaque detection tasks, clinicians prioritize not only accurately identifying plaque regions but also maintaining precision in distinguishing background tooth areas. The F1-score effectively balances Precision and Recall, offering a comprehensive reflection of the model’s overall performance in real-world clinical scenarios.

\paragraph{\textbf{Accuracy}} is a fundamental and widely used evaluation metric that measures the proportion of correctly classified pixels in the overall prediction. As defined in Equation. \ref{eq9}, Accuracy ranges from $[0,1]$, with higher values indicating a greater proportion of correctly predicted pixels, thereby reflecting the model’s overall discriminative capability.

\begin{equation}
\label{eq9}
\text{Accuracy} = \frac{TP + TN}{TP + TN + FP + FN}
\end{equation}
In DIA, Accuracy is frequently used to evaluate the overall performance of classification and segmentation tasks. However, because background pixels often dominate dental images, relying solely on Accuracy can be misleading in scenarios with class imbalance. Therefore, Accuracy is typically used alongside the aforementioned metrics to provide a more comprehensive and reliable performance assessment.

\subsubsection{Metrics based on the Set and Geometry}
\paragraph{\textbf{Hausdorff Distance (HD)}} is an important metric for evaluating the discrepancy between the predicted and true boundaries in segmentation tasks. It quantifies the maximum distance between the predicted boundary and the ground truth (GT) boundary. A smaller HD value indicates closer alignment between the predicted and true boundaries, reflecting higher precision in boundary localization. The mathematical definition is presented in Equation. \ref{eq9}.

\begin{equation}
\label{eq10}
\text{HD}(A,B) = \max \left\{ \sup_{a \in A} \inf_{b \in B} \|a-b\| , \sup_{b \in B} \inf_{a \in A} \|b-a\| \right\}
\end{equation}
Here, $A$ and $B$ denote the sets of predicted and GT boundary points, respectively, while $a$ and $b$ represent arbitrary points within each set. $\inf$ (infimum) refers to the greatest lower bound, taking the minimum among all distances, and $\sup$ (supremum) refers to the least upper bound, taking the maximum among all values. $|| \cdot ||$ denotes the Euclidean distance, which measures the spatial distance between two points.   In DIA, particularly in 3D CBCT analysis, HD is a commonly used metric for evaluating boundary accuracy and is widely applied in the segmentation assessment of teeth, root canals, and jawbones. Since DSC or IoU alone cannot accurately capture boundary precision, HD's sensitivity to boundary deviations makes it an important complement to overlap-based metrics.

However, the standard HD is sensitive to noise, outliers, and anomalous points. To address this issue, the 95th percentile Hausdorff Distance (HD95) is commonly used in practical studies. HD95 considers the 95th percentile of the boundary distance distribution instead of the maximum distance, thereby reducing the impact of extreme points on the evaluation. The mathematical definition is presented in Equation. \ref{eq11}.

\begin{equation}
\label{eq11}
\text{HD95}(A,B) = \text{percentile}_{95} \left( \{ d(a,B), d(b,A) \} \right)
\end{equation}
HD95 eliminates the influence of extreme outliers, providing a more stable and objective reflection of the model’s actual boundary performance.

\paragraph{\textbf{Average Surface Distance (ASD)}} quantifies the average distance between the predicted segmentation boundary points and the ground-truth boundary points. This metric provides insight into the general alignment of the predicted boundaries with the true boundaries, as defined in Equation. \ref{eq12}.

\begin{equation}
\label{eq12}
\text{ASD}(A,B) = \frac{1}{|A|} \sum_{a \in A} d(a,B),
\end{equation}
Here, $A$ denotes the set of predicted boundary points, and $B$ represents the set of ground-truth boundary points. $|A|$ is the cardinality of set $A$, and $d(a, B)$ denotes the minimum distance from point $a \in A$ to set $B$. A smaller ASD indicates that the predicted boundary is, on average, closer to the GT boundary. In DIA, ASD is commonly used to assess the boundary consistency of target structures such as tooth surfaces. It is frequently employed alongside HD95 to provide a more comprehensive evaluation of boundary accuracy.

\paragraph{\textbf{Average Symmetric Surface Distance (ASSD)}} is an extension of ASD. While ASD considers only the average deviation from the predicted boundary to the GT boundary, it may underestimate errors when the GT boundary exhibits complex structures. To address this limitation, ASSD calculates the average distance from the predicted boundary to the GT and from the GT to the predicted boundary, then takes the mean of both distances. This symmetric approach provides a more comprehensive measure of boundary accuracy. Its definition is given in Equation. \ref{eq13}.

\begin{equation}
\label{eq13}
\text{ASSD}(A,B) = \frac{1}{|A| + |B|} \left( \sum_{a \in A} d(a,B) + \sum_{b \in B} d(b,A) \right),
\end{equation}
In DIA, Average Symmetric Surface Distance (ASSD) is frequently used as a key metric for assessing boundary accuracy, offering a more intuitive and precise geometric error measurement for evaluating model performance in clinical applications.

\section{Discussion, Challenges and Future Directions}
\label{sec5}
In the preceding sections, we systematically reviewed DL-based studies in DIA from three perspectives: data, models, and tasks with applications. The findings indicate that DL and AI technologies have become the most widely adopted core methodologies in this field. Building upon this foundation, this chapter further summarizes research progress across these aspects, analyzes the major challenges currently facing DIA studies, and discusses potential future directions and trends.

\subsection{Discussion and Challenges}
\subsubsection{Public Datasets}
Public datasets are the cornerstone and essential resource driving advancements in AI and DL research. They provide standardized benchmarks for algorithm training and performance evaluation, enabling reproducibility and fair comparison of research outcomes. However, constructing medical public datasets poses significantly greater challenges compared to natural image datasets. On one hand, medical image annotation requires the involvement of clinicians with specialized expertise, as it often involves delineating complex anatomical structures or pathological regions, making the process time-consuming and costly. On the other hand, medical imaging data are diverse in modality and complex in acquisition, necessitating professional radiological procedures and strict adherence to patient privacy and ethical regulations. Consequently, the establishment and sharing of medical public datasets face substantial technical, temporal, and regulatory obstacles compared to their natural image counterparts.

This issue is particularly pronounced in the field of DIA. Most researchers still rely heavily on private datasets for experimentation and validation. Although considerable progress has been made, the lack of publicly available and reusable datasets hinders the full verification of algorithm reproducibility and transferability. As shown in Table \ref{tab1}, while a number of public DIA datasets have emerged, they remain significantly smaller in scale and quantity compared to those in other medical imaging domains, such as fundus images \cite{wuMMRetinalKnowledgeEnhancedFoundational2024, li2019diagnostic} or chest X-ray analysis \cite{johnsonMIMICCXRJPGLargePublicly2019, zhaoTopicwiseSeparableSentence2024}. Even when the sample size exceeds 5K \cite{li2024multi, aminhj_tooth_decay_2023, seifossadat_dental_radiography_segmentation_2023, engineeringubu_root_disease_xray_2024}, many datasets only provide detection- or classification-level annotations, lacking high-quality, dense segmentation masks—thereby constraining advancements in fine-grained segmentation research.

From a modality perspective, most existing public datasets are focused on CBCT and PAN images, while modalities such as IOS, PR, and IP have received relatively limited attention. However, these data types hold substantial clinical value in orthodontics, endodontics, and oral health, respectively, and should be given greater emphasis in future research. Notably, with the increasing application and proven potential of vision-language models (VLMs) in medical image analysis \cite{guRadAlignAdvancingRadiology2025, wuUniBiomedUniversalFoundation2025, nathVILAM3EnhancingVisionLanguage2025}, several DIA studies have begun constructing multimodal dental imaging datasets to explore the feasibility and benefits of multimodal learning in this field \cite{wang2025mmdental, silva2025holistic}. Encouragingly, since 2023, multiple international challenges dedicated to DIA have been launched \cite{ricoleehduu_STS-Challenge_2023, sts_challenge_2024_website, sts2025_organizers, bolelli2025segmenting}, providing standardized platforms for fair algorithm evaluation, enhancing research engagement, and fostering the dissemination of high-quality scientific outcomes.

However, the construction of high-quality public datasets for DIA still faces several pressing challenges. First, existing datasets are limited in both scale and quality, with incomplete coverage of certain imaging modalities that require further supplementation and refinement. Second, most available datasets are derived from economically developed regions, while publicly accessible DIA data from less developed areas remain scarce. According to the World Health Organization, Africa accounts for approximately 23\% of the global disease burden but possesses only 4\% of the world’s healthcare workforce \cite{world2025workforce}. As of 2022, there were only about 34,405 dentists across the entire continent, averaging 0.29 dentists per 10,000 people. Under such an imbalanced ratio of medical professionals to patients, the need for intelligent diagnostic and treatment assistance systems becomes particularly urgent, and the availability of open datasets serves as the fundamental basis for their development. Furthermore, due to disparities in imaging equipment, environmental conditions, and operator expertise, models trained on data from developed regions often fail to generalize effectively to less developed areas, resulting in a significant decline in performance. Addressing this cross-domain distribution discrepancy has thus become one of the core scientific challenges in current DIA research. Finally, with the growing adoption of VLMs, the construction of multimodal datasets for DIA has become increasingly imperative. Exploring the integration of textual data into dental imaging tasks has emerged as a key research direction. Although several studies have attempted to develop such multimodal datasets, most remain single-modality and limited in scale, making them insufficient to support the training of large-scale, comprehensive VLM baseline models for dental applications.

\begin{figure*}[h]
	\centerline{\includegraphics[width=\linewidth]{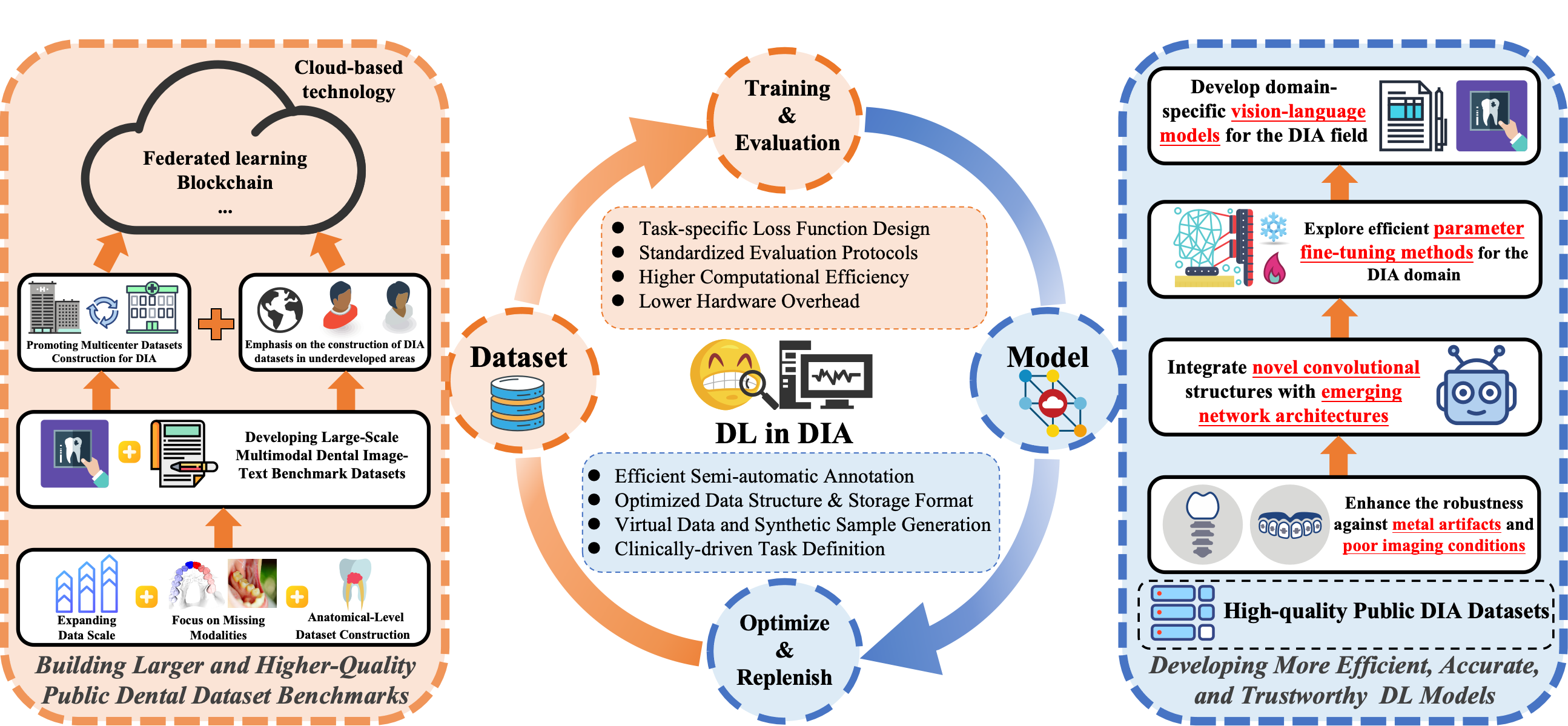}}
	\caption{Overview of future research directions of DL in DIA. The framework is divided into two main parts: the left side focuses on datasets, and the right side focuses on models. These two aspects are complementary and mutually reinforcing, jointly advancing the intelligence and clinical applicability of DIA.}
	\label{fig24}
\end{figure*}

\subsubsection{Model and Performance}
Compared to traditional algorithms, the introduction of DL has significantly enhanced the accuracy and efficiency of DIA tasks. Due to the advantages of deep models in large-scale data feature representation learning, DL methods demonstrate superior performance over traditional algorithms in handling complex dental images and diverse tasks. From a model architecture perspective, most current studies focus on CNN architectures or hybrid models combining CNN with ViT, whereas the application of pure ViT models in DIA is relatively rare. This is mainly because tooth structures are complex and variable, and the boundaries between different tissues and bones are often indistinct. The strong ability of CNNs in local feature extraction is crucial for capturing these detailed features. Combining the global modeling capability of ViT with the local feature learning ability of CNN allows for better feature representation and complementary performance; however, pure ViT models, lacking local sensitivity, exhibit relatively limited performance in DIA tasks.

In model designs, incorporating various attention mechanisms to enhance feature representation has become a common optimization strategy. Some researchers have also attempted to integrate expert prior knowledge into the training process, explicitly improving the model’s ability to recognize and differentiate specific dental target structures. Overall, deep learning has achieved remarkable progress in core tasks such as tooth segmentation. For instance, in CBCT-based tooth identification, the average DSC of existing methods has exceeded 94\%; in IOS-based tooth segmentation, the average DSC can reach over 95\%; and even in PAN images, which are more susceptible to artifacts and low resolution, the instance-level tooth segmentation performance generally achieves an average DSC above 90\%. These results clearly demonstrate that DL-based methods have reached a high level of performance in DIA; however, further improvements will depend on exploring more generalizable model architectures with stronger multimodal fusion capabilities.

Dentin, enamel, and dental pulp are the three primary internal structures of the tooth and represent core targets in anatomy-level prediction tasks. For the segmentation and recognition of such fine-grained structures, most studies continue to adopt CNN-based architectures, often combined with multi-scale feature extraction modules to improve prediction accuracy. Due to the lack of high-quality publicly available anatomy-level datasets, most current research relies on private data for experimentation, resulting in considerable variability among studies, with average DSC values ranging from 85\% to 95\%. In dental lesion analysis, the IP modality has seen a notable rise in application in recent years due to its ability to visually present the oral condition. Among various lesions, dental plaque and caries are the most intensively studied. To more effectively differentiate normal tissues from lesion regions, some studies, in addition to conventional attention mechanisms, have introduced contrastive learning approaches to enhance inter-class discriminability from the perspective of feature space.

In maxillofacial structure prediction tasks, CBCT is the most widely used imaging modality, and most related studies continue to rely on CNN-based architectures. Overall, high segmentation accuracy has been achieved for structures such as the temporomandibular joint, maxillary sinus, and alveolar bone, with average DSC values typically exceeding 97\%. However, for finer anatomical structures such as the mandibular canal and inferior alveolar nerve, segmentation accuracy remains relatively low due to their subtle morphology and indistinct boundaries, with average DSC values ranging from 75\% to 85\%, indicating room for further improvement.

For classification or other non-dense network tasks, existing studies can generally be categorized into two types. The first type leverages dense prediction results as prior information to facilitate tasks such as multi-lesion or tissue classification, implant planning, and tooth surface reconstruction. The second type incorporates tooth-number classification as an auxiliary branch to enhance the precision of instance-level dense detection. Overall, studies focusing solely on classification tasks are relatively scarce, as most works tend to integrate classification and segmentation, thereby exploiting the advantages of multi-task learning in feature sharing and structural constraint.

In terms of model development and performance optimization, DIA still faces multiple challenges. Firstly, dental imaging is often affected by suboptimal acquisition angles, limited radiation doses, and the presence of dental fillings or implants, which can introduce noise or metal artifacts and interfere with image interpretation. Consequently, existing models require enhanced robustness when handling noisy inputs. Secondly, model diversity in the DIA field remains insufficient. According to recent surveys, deep learning applications in DIA are still primarily based on conventional CNN architectures, while the potential of emerging CNN paradigms \cite{liuMoreConvNets2020s2023, xuPreCMPaddingbasedRotation2025} and innovative architectures comparable to ViT (e.g., Mamba \cite{liuVMambaVisualState2024} or RWKV \cite{duanVisionRWKVEfficientScalable2025}) has yet to be fully explored in dental scenarios. Efficiently leveraging these new models, or integrating different architectures to exploit their complementary strengths, remains a crucial direction for future research. Furthermore, some foundation models trained on ultra-large-scale medical imaging datasets (e.g., MedSAM \cite{ma2024segment}) perform suboptimally on dental tasks, primarily due to the scarcity of dental data during training. Therefore, developing parameter-efficient fine-tuning strategies or semi-supervised transfer approaches under limited data conditions is of significant importance for improving the predictive accuracy of foundation models in dental applications. This approach not only reduces dependence on costly annotated data but also substantially enhances the practicality and generalizability of the models.

\subsection{Future Directions}
As shown in Fig. \ref{fig24}, we discuss the potential future research directions of DL in DIA from two perspectives: dataset construction and model development.

\textbf{From the Perspective of Public Datasets Construction:}

\begin{itemize}
\item Develop larger-scale, high-quality public DIA benchmark datasets, promote cross-hospital and cross-regional collaboration, and leverage key technologies such as federated learning or blockchain to expand data sources while ensuring patient privacy and ethical compliance, thereby enhancing data diversity and coverage;

\item Increase focus on establishing DIA databases in underdeveloped regions by leveraging cloud-based annotation and federated sharing systems, collaborating with local universities, medical institutions, and research centers, and involving clinicians from other regions to assist with data annotation and processing, thereby enabling efficient dataset construction;

\item Efforts should be intensified to construct datasets for underrepresented modalities, such as IOS and PR. Although CBCT provides more comprehensive scanning information, its high cost and radiation dose prevent it from fully replacing other imaging modalities. Therefore, developing datasets for these relatively scarce modalities remains crucial;

\item The construction of multimodal dental datasets should be given high priority. Currently, publicly available multimodal DIA datasets are scarce, and existing datasets are often limited to specific modalities and tasks, lacking diversity in both image modalities and text categories. Therefore, establishing a comprehensive large-scale multimodal benchmark dataset is of great significance and serves as a crucial foundation for developing large-scale dental VLMs.
\end{itemize}

\textbf{From the Perspective of Model Developments:}
\begin{itemize}
\item More targeted attention mechanisms or loss functions should be developed, in combination with generative training strategies such as diffusion models, to improve the model’s ability to recognize images with poor shooting angles or metal artifacts. In addition, the model’s robustness across different regions, institutions, and imaging devices should be further enhanced to ensure stable and generalizable performance in diverse and complex clinical environments;

\item The potential of novel architectures in DIA tasks should be thoroughly explored, focusing on how to effectively leverage new convolutional paradigms and organically combine them with other architectures to exploit complementary strengths. Under the constraints of efficient training and low deployment overhead, such efforts should aim to further improve model performance and generalization in DIA applications;

\item Efficient parameter-efficient tuning models or adapters should be designed to fully leverage foundation models pretrained on large-scale medical datasets. By integrating lightweight fine-tuning techniques such as LoRA, domain-specific dental imaging knowledge can be effectively injected into the model, enabling efficient transfer and reuse of large foundation medical image analysis models for DIA tasks;

\item Expert-level VLM architectures and training frameworks tailored for the DIA domain should be developed. Unlike other medical imaging fields, dental targets are often more complex, requiring models to possess finer-grained understanding of both visual and textual information. Moreover, the model should be capable of interpreting tooth numbering and integrating positional and structural features from different tooth locations to achieve comprehensive understanding of dental images.
\end{itemize}

\section{Conclusion}
\label{sec6}
Efficient, accurate, and reproducible DIA technology serves as the foundation and key enabler for computer-aided dental diagnosis. In this paper, we provide a comprehensive review of recent advances and research trends in the application of DL to DIA. Centered on two fundamental pillars of DL research—public datasets and model architectures—we systematically summarize 49 publicly available DIA datasets and 213 related studies. We begin by introducing the research background and major task types in DIA, followed by a comparative analysis of existing review literature and a detailed description of our literature retrieval and screening methodology. Section 2 presents the main dental imaging modalities, illustrates real-world clinical imaging equipment, and systematically reviews the current landscape of public DIA datasets, including detailed information and access methods to facilitate dataset reuse by researchers. In Section 3, we categorize existing research based on task type and model architecture, covering dense prediction, classification, and other tasks, as well as CNN-based, ViT-based, and hybrid models. For dense prediction tasks, we further divide studies into four subcategories targeting tooth-level, anatomical-level, pathological, and maxillofacial structure analysis. Summarizing the key features and innovations of each work, we also analyze representative models in terms of architecture and performance, discussing the relationships and potential transferability between different task types. Furthermore, we review commonly used loss functions, optimizers, GPU configurations, and performance metrics in DIA model training to provide practical experimental guidance. Finally, we discuss the current state, limitations, and future directions of DIA research from two perspectives: data and models. We hope this review will serve as a valuable reference for the DIA research community—offering systematic background knowledge to newcomers and fresh insights to experienced researchers. Looking ahead, with the establishment of more high-quality open datasets and the emergence of innovative algorithms, DIA technologies are expected to achieve higher levels of intelligence and greater clinical applicability.

\section*{Acknowledgments}
This work is partially supported by the National Natural Science Foundation (62272248), the Natural Science Foundation of Tianjin (23JCZDJC01010).

\bibliographystyle{./elsarticle-harv.bst}
\bibliography{./ref.bib}

\end{document}